%% file: neurips_2026.tex
\documentclass{article}

\usepackage[dblblindworkshop, final, nonatbib]{neurips_2026}
\usepackage{pgfplots} \pgfplotsset{compat=1.18}

\usepackage[utf8]{inputenc}
\usepackage[T1]{fontenc}
\usepackage{url}
\usepackage{booktabs}
\usepackage{amsfonts}
\usepackage{amsmath}
\usepackage{nicefrac}
\usepackage{microtype}
\usepackage{xcolor}

\usepackage{multirow}
\usepackage{graphicx}
\usepackage{tabularx}
\usepackage{float}

\usepackage{algorithm}
\usepackage{algpseudocode}

\usepackage{hyperref}
\usepackage{array}
\usepackage{needspace}
\usepackage{enumitem}
\usepackage{caption}
\usepackage[
     backend=biber,
]{biblatex}
\AtEveryBibitem{
    \clearfield{urlyear}
    \clearfield{urlmonth}
    \clearfield{doi}
    \clearlist{language}
}
\DeclareSourcemap{\maps[datatype=bibtex]{
  \map{\step[fieldsource=journal, match=\regexp{^arXiv\s+preprint}, final]\step[fieldset=eprint, null]}
  \map{\step[fieldsource=booktitle, match=\regexp{^arXiv\s+preprint}, final]\step[fieldset=eprint, null]}}}
\newenvironment{promptbox}[1]{%
  \par\medskip\noindent\textbf{\small #1}\par\smallskip
  \begingroup\small\ttfamily\raggedright\sloppy\leftskip=1.2em\relax}%
  {\par\endgroup\medskip}

\title{What Does Multiple Choice Actually Measure? Disentangling Video Understanding from Option Exploitation}

\workshoptitle{TAE (Trust-AI-Eval): Can We Trust AI Evaluation?}

\author{%
  Elliott Piercy\thanks{Equal Contribution}\qquad
  Justin Chan\qquad
  Sepand Dyanatkar\textsuperscript{\normalfont$\ast$,}\thanks{Corresponding author:
    \href{mailto:sepand@ondeckai.com}{\texttt{sepand@ondeckai.com}}.} \\[4pt]
  OnDeck AI \\
}

\input{preamble}
\begin{document}

\maketitle

\begin{abstract}
    Video understanding evaluations are intended to measure a model's proficiency in video perception and task solving. 
    However, current VideoQA benchmarks struggle to isolate these capabilities due to their reliance on multiple-choice (MC) answer guidance, language priors, and fragile scoring pipelines.
    Disentangling these confounding factors reveals how much of a model's score is supplied by the evaluation context rather than earned through faithful perception.
    Across six VideoQA benchmarks of varying length, we find that accuracy is significantly above chance without any visual input. 
    This is caused by shortcutting and option-prior reliance.
    We express these effects in the language of faithfulness: option-exploitation and answer-conditioned reasoning traces are unfaithful even when they are correct. 
    Together, our results show that a VideoQA score is jointly determined by language priors, option timing, and how outputs and labels are converted into correctness.
    We produce a novel Option-Delayed Replay (ODR)  mechanism that discloses options at controlled stages of an open-ended (OE) trajectory, to obtain objective MC scores from faithful option-free inference.
    Our work demonstrates that agentic methods with ODR and faithful inference can match option-exploiting performance.
    Finally, we formalize our diagnostic collection and ODR as a reproducible evaluation protocol, and we release a human-relabelled edition of LVBench (correcting $20.5\%$ of keys) to remove annotation noise as a confounding variable.
\end{abstract}

\section{Introduction}

Vision Language Models (VLMs) have achieved rapid progress on Video Question Answering (VideoQA) tasks, with benchmark accuracy commonly interpreted as a measure of visual understanding. However, two issues increasingly undermine the reliability of current VideoQA benchmarks. First, multiple-choice (MC) evaluation provides strong contextual cues through a fixed answer set, enabling elimination strategies and surface matching that inflate performance relative to open-ended generation. Second, the widespread use of public benchmarks raises concerns that frontier closed-source models may have been trained on, or otherwise shortcut, benchmark test data, making it difficult to distinguish genuine video understanding from data exploitation. Together, these factors can substantially overestimate a model's perceptual reasoning ability. For real-world deployment this is a faithfulness problem: a high benchmark number can reflect option-exploitation and answer-conditioned reasoning rather than true perception, so it does not, on its own, license trusting the model to watch and reason faithfully in the wild. Ultimately, for video understanding research, this makes it hard to diagnose and improve systems. 

While the most reputable video benchmarks rely on multiple-choice (MC) VideoQA, providing answer options upfront conflates their effect on evidence acquisition and reasoning with their effect on final answer generation and scoring.
Existing work identifies the numerous challenges to measurement validity \cite{chen_are_2024,accuracywithoutgrounding2026,vbenchcomp,videoevalpro2026,neptune2024,huang_what_2018,ucf101ad2026}.

Rather than simply comparing MC and open-ended (OE) performance, we ask a different question: how much does each piece of additional context actually help and when? 
We answer these by systematically withholding elements, starting from the question alone (no video) and progressively adding elements or perturbing them, such as the video, retrieved evidence, and finally the answer options~\cite{loginova_addressing_2025}.
By analysing the impact of answer extraction, context-dependent evaluations, and long-video performance with respect to dataset quality, we show that benchmark scores also depend on the evaluation pipeline itself.

Disentangling when options matter is especially critical for agentic VideoQA systems, where candidate answers can remain in context across multi-step tool use. 
Existing work does not show how the MC advantage enters an agentic vision pipeline: if an agent scores higher with access to the options, is it because it grounds its visual reasoning better, or because it uses the options to guide its search and selection? 

Specifically, an agent may use answer choices throughout its trajectory to decide what to inspect, when to stop, and how to interpret retrieved observations. Alternatively, it may gather sufficient evidence without bias from choices, and only need choices to express its final answer. 
We introduce \emph{Option-Delayed Replay} (ODR) to separate these cases by resuming checkpointed OE trajectories with the candidate options disclosed either mid-search or only at final selection.

\subsection{Contributions}
We make three core contributions:
(i) Building on prior VQA input ablations, we execute a unified diagnostic decomposition across six diverse VideoQA benchmarks, pairing direct native-video QA with three heterogeneous agentic vision systems. We quantify non-visual solvability (including a model-free heuristic floor), answer-extraction and judge sensitivity, and the gap between free generation, post-hoc option matching, and native MC scoring.
(ii)~We introduce \emph{Option-Delayed Replay} (ODR) for agentic vision, an agentic protocol that hides the answer options during search and reveals them only at a the desired step. ODR ensures an agent's evaluation measures evidence acquisition and reasoning (which we consider `faithful'), rather than option-exploitation.
(iii)~We relabel a widely-used dataset of long-video understanding, LVBench, correcting $20.5\%$ of ground-truth keys and changing question, answer or distractor content in over $30\%$ of items. We release the labels to ensure question, answer, and distractor quality are removed as a confounding factor in vision evaluation.

We first outline pre-existing work that builds a diagnostic and evaluation collection for VideoQA, as well as motivation for our contributions. Then we describe the procedures for executing our analyses with direct QA and vision agents, including the diagnostics' procedures. We finally show the results, and we provide recommendations for faithful video understanding evaluations and future research.

\section{Related Work}

\textbf{Visual and Temporal Necessity.}
Language priors and unimodal shortcuts are longstanding concerns in visual question answering~\cite{vqav2_2017,vqacp_2018}.
Audits show many video benchmark queries remain answerable without the video, with a single frame, or after temporal perturbation~\cite{chen_are_2024, accuracywithoutgrounding2026, buch_revisiting_2022}. 
VBenchComp uses no-video and shuffled-frame behaviour to categorize questions as language-answerable, semantic, temporal, or residual~\cite{vbenchcomp}, while recent work more directly decomposes whether benchmark performance depends on visual content or temporal order~\cite{temporalbenchmark2026}. 
Going beyond these necessity findings, our work connects visual necessity to answer format, scoring/evaluation, and agentic inference (including evidence acquisition) across key video benchmarks. 

\textbf{Option-dependence during evaluation.}
Evaluation is confounded further when using MC options alongside queries. 
Options expose regularities, show the correct answer between similar distractors, and provide candidate hypotheses for reasoning. 
As a result, strong language priors are even more capable of shortcutting queries with options~\cite{dovideollmswatch2026} and there is a substantial MC$-$OE gap across diverse video benchmarks~\cite{videoevalpro2026}. 
Selection and distractor bias remain a persistent problem, regardless of the isolation mechanism~\cite{mcvqa_bias2023, loginova_addressing_2025, zhang_mitigating_2025}. 
\textcite{loginova_addressing_2025} in particular analyse selection bias through eleven variants to the video, question, and answer options. 
They also propose calibration to reduce option-position bias.
The persistent problem of the MC$-$OE gap motivates our decomposition across four stages: free-form OE generation, candidate recovery with our OE$\to$MC matching, answer extraction and scoring, and option-conditioned search.

\textbf{Benchmark validity and scoring.}
Removing options avoids exposure during inference, but depends on semantic evaluation, which is challenging to do well~\cite{autoconverter2025}---though reference-aware answer matching can nearly recover human grading in text QA~\cite{chandak_answer_matching_2025}. 
Benchmark quality and construction pose additional problems due to poor or duplicate distractors, ambiguous questions or options, and objective mistakes in content. This adds noise and compresses the range of useful benchmark scores (see examples in Appendix~\ref{sec:appendix-a3}). 
Structural drawbacks of MC VQA have been known since 2021~\cite{nextqa2021}, yet the format remains standard for video understanding. 
Recent work such as \textcite{autoconverter2025} even automates OE$\rightarrow$MC conversion based on image-VQA heuristics~\cite{datologyai_datbench_2026}. 
Because complex non-MC video datasets are expensive to build, the field has leaned heavily on MC VQA with AI-generated distractors. 
As a result, publicized video benchmarks often measure option-exploitation within an AI-generated answer space rather than grounded video understanding. 
We investigate scoring reliability by combining two LLM judges with post-hoc option matching to triangulate scoring sensitivity on fixed outputs. Furthermore, our human re-annotation of LVBench minimizes construction failures to better isolate the factors affecting video evaluation~\cite{lvbench2025} (Appendix~\ref{app:lvbench-relabel}).

\textbf{Option-timing analysis for video understanding.}
Option timing affects evidence acquisition, not just final selection. 
Agentic video understanding is structured to acquire evidence and reason to resolve queries~\cite{dvd2025, avp2026, morevqa2024}, suggesting a more faithful approach. 
However, \textcite{qiu_videoseal_2026} find that correct answers can still be unsupported by retrieved evidence, similar to shortcutting in direct QA. 
We introduce Option-Delayed Replay (ODR) for agentic vision, disclosing candidate options at controlled checkpoints along an OE trajectory to measure how option exposure changes subsequent tool calls, requested video segments, stopping behaviour, and final answer selection.
The closest prior work is \textcite{raman_reasoning_2025}, who test whether text QA models see options before or after chain-of-thought reasoning across $15$ benchmarks and $27$ models. They find that reasoning \emph{after} seeing the options inflates performance over free-text QA, whereas reasoning \emph{before} the options appear keeps MC evaluation honest. 
We extend this concept to autonomous decision-making environments in vision rather than single-pass text decoding. 
In agentic VideoQA, options can change what visual evidence the agent chooses to gather in the first place. 
Because ODR resumes from a saved trajectory checkpoint rather than comparing independent generations, we can inject options mid-trajectory or at the final step to test \emph{when} options begin to affect search and accuracy, while also repurposing existing MC datasets to reduce option-induced bias.

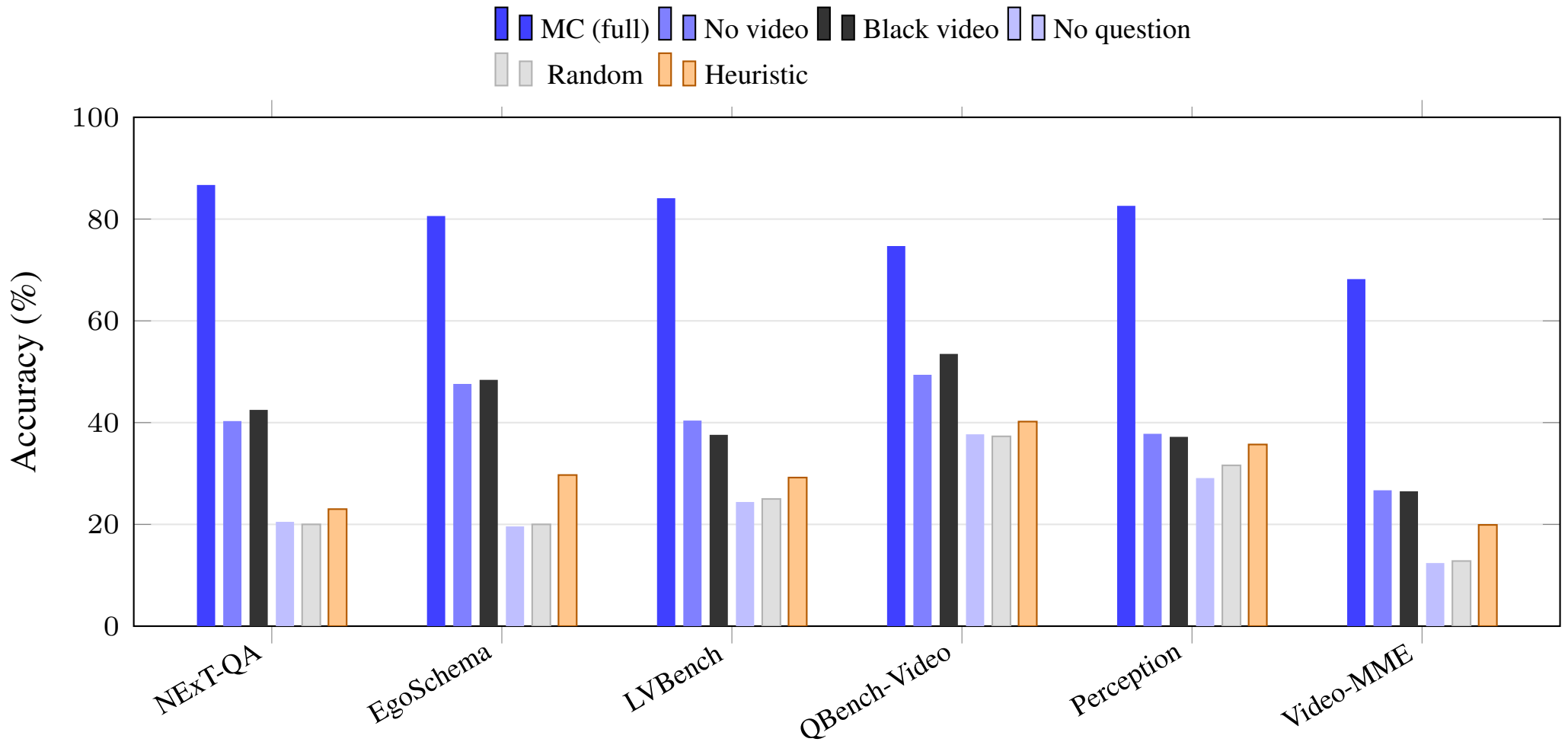
\begin{figure}[t]
  \centering
  \begin{tikzpicture}
\begin{axis}[
    ybar,
    width=\linewidth,
    height=6cm,
    bar width=4.5pt,
    ylabel={Accuracy (\%)},
    ymin=0,
    ymax=100,
    symbolic x coords={
        NExT-QA,
        EgoSchema,
        LVBench,
        QBench-Video,
        Perception,
        Video-MME
    },
    xtick=data,
    x tick label style={
        rotate=30,
        anchor=east,
        font=\scriptsize
    },
    tick label style={font=\scriptsize},
    label style={font=\small},
    ymajorgrids=true,
    grid style={gray!20},
    enlarge x limits=0.12,
    legend style={
        at={(0.5,1.02)},
        anchor=south,
        legend columns=4,
        draw=none,
        font=\scriptsize
    },
]

\addplot[fill=blue!75, draw=none] coordinates {
    (NExT-QA,86.7)
    (EgoSchema,80.6)
    (LVBench,84.1)
    (QBench-Video,74.7)
    (Perception,82.6)
    (Video-MME,68.2)
};

\addplot[fill=blue!50, draw=none] coordinates {
    (NExT-QA,40.3)
    (EgoSchema,47.6)
    (LVBench,40.4)
    (QBench-Video,49.4)
    (Perception,37.8)
    (Video-MME,26.7)
};

\addplot[fill=black!80, draw=none] coordinates {
    (NExT-QA,42.5)
    (EgoSchema,48.4)
    (LVBench,37.6)
    (QBench-Video,53.5)
    (Perception,37.2)
    (Video-MME,26.5)
};

\addplot[fill=blue!25, draw=none] coordinates {
    (NExT-QA,20.5)
    (EgoSchema,19.6)
    (LVBench,24.4)
    (QBench-Video,37.7)
    (Perception,29.1)
    (Video-MME,12.4)
};

\addplot[
    fill=gray!25,
    draw=gray!60
] coordinates {
    (NExT-QA,20.0)
    (EgoSchema,20.0)
    (LVBench,25.0)
    (QBench-Video,37.3)
    (Perception,31.6)
    (Video-MME,12.8)
};

\addplot[
    fill=orange!45,
    draw=orange!70!black
] coordinates {
    (NExT-QA,23.0)
    (EgoSchema,29.7)
    (LVBench,29.2)
    (QBench-Video,40.2)
    (Perception,35.7)
    (Video-MME,19.9)
};

\legend{
    MC (full),
    No video,
    Black video,
    No question,
    Random,
    Heuristic
}

\end{axis}
  \end{tikzpicture}
  \caption{The core context variations against direct QA, drawn against each benchmark's chance baseline and its model-free heuristic. The gap between the blind bars and chance is the accuracy the
  question and options carry on their own; the \emph{No question} bars withhold the options along with the question. See more in Table~\ref{tab:core}.} 
  \label{fig:context-ablation}
\end{figure}

\section{Methods and Evaluation}
\label{sec:methods}

We investigate whether the apparent advantage of MC VQA over OE VQA reflects improved video understanding across six multiple-choice benchmarks: NExT-QA, Perception Test, EgoSchema, LVBench, QBench-Video, and Video-MME\,v2~\cite{lvbench2025, egoschema2023, qbench, videommev2_2026, perceptiontest2025challenge, nextqa2021}. We chose the intersection of high quality, reputable, and commonly used video understanding datasets, yet they still reflect common problems for both evaluation and dataset construction.
Appendix~\ref{app:benchmarks}, Table~\ref{tab:benchmarks} states the exact evaluated split, $n$, option count, modality, and scorer for each of the six.
Our evaluation separates four potential performance metrics attributed to visual grounding, question structure, answer space and shortcutting, where models exploit benchmark examples or answer patterns encountered during training.

Above-chance performance without video establishes non-visual solvability, but does not by itself distinguish between factors such as language priors, benchmark artifacts, or training-data contamination. We therefore reserve claims of memorization or leakage for analysis with conclusive proof.

\subsection{Models and inference regimes}
We evaluate frontier methods using two inference regimes: direct question answering (QA) and three agentic vision approaches. 
For direct QA, we use Gemini 3.6 Flash, a native-video model with a 1,048,576-token context window and leading performance on video evaluations~\cite{gemini3.6}.

Video is sampled at 1\,fps by default and at 0.5\,fps for LVBench videos exceeding the context limit. 
For any given item, the sampled video is identical across its paired MC, OE, and partial-input conditions.
Deep Video Discovery (DVD)~\cite{dvd2025,dvd_code}, Active Video Perception (AVP)~\cite{avp2026,avp_code}, and Gaia are our three agentic approaches that enable our investigation over richer grounding and multi-step inference.
DVD is caption-based, while AVP is caption-free, but both perform agentic search by iteratively gathering evidence and reasoning. We use the authors' released implementations and configurations.
Gaia is a proprietary black-box agent loop.
We ensure Gaia's completion protocol is equivalent to DVD's \texttt{finish} for ODR equivalence.
We restrict the upper bound of Gaia's token-usage \textit{and} tool calling across trajectories to be equivalent to DVD's (which is much larger than AVP's across all datasets).
We otherwise withhold Gaia's orchestration and evaluate it as a black box. 
All three systems are held fixed across evaluation conditions and benchmarks. 
Because the agents use different backbones and video representations, we use them as heterogeneous replications of evaluation-format effects rather than to rank agent architectures.

\subsection{Multiple-choice and open-ended evaluation}
\label{sec:mc-oe}

For each dataset and inference method we construct corresponding open-ended (OE) evaluation by withholding the answer options. 
MC predictions are scored by exact match against the ground-truth option, while OE responses are evaluated using an LLM-as-a-judge scorer. 
We use each dataset's official evaluator, augmented with an LLM fallback to recover the selected option when a model produces a verbose response rather than a bare option letter.
The authors' released QBench-Video and EgoSchema scorers accept only a bare option letter and mark a verbose but otherwise correct answer wrong. 
For Video-MME\,v2, the primary reported score is the flat per-question MC accuracy, with its native non-linear group score also calculated and included off to the side. 
See Appendix~\ref{app:benchmarks}, Table~\ref{tab:benchmarks} for more details about datasets and splits used.

MC options can provide substantial information allowing the model to constrain the search space and ground on suitable candidates. OE evaluation removes these biases, creating a harder problem where recovering the answer requires correct \textit{evidence accumulation and reasoning}.

We do not infer data memorization from the MC$-$OE performance gap alone. MC answering is inherently easier than OE with differences in answer formulation, evaluation type or task type impacting the final result.

\subsection{Trajectory Replay and Option-Delayed Replay}

We introduce \textbf{agentic vision Replay} as an additional diagnostic tool. Given a recorded trajectory of $N$ steps, Replay restores the serialised state of the agent at any step $k$, including its message history and previous tool observations. This mechanism allows us to explore the tree of executions for a trajectory at controlled checkpoints. 
Replay restores the agent's observable state rather than hidden model activations, and is independent of any particular agent architecture or video representation.

In this work we introduce a novel \textbf{Option-Delayed Replay} (ODR).

In agentic VQA, providing the answer choices may change more than the format of the final response. The choices constrain the space of plausible answers, and can therefore guide the evidence gathering process. Consequently, MC evaluation may make the underlying video-search problem easier, and investigating this counterfactual requires controlled re-execution~\cite{dover, causalflow} with ODR. 

ODR first executes an agentic trajectory of $N$ agentic turns (where each turn may execute 1 to many actions) in the OE setting. We then evaluate ODR from 3 intervention points, Start (equivalent to regular MC), Midpoint, and Last-step:

\begin{enumerate}
    \item \textbf{MC Options from Start.} The agent receives the answer choices at $k=0$ before beginning its trajectory and can use them throughout its lifetime.

    \item \textbf{Midpoint Option Disclosure Replay.} We preserve the OE trajectory up to step $k=\lfloor N/2 \rfloor$, disclose the answer choices, and allow the agent to generate a new continuation from this point.

    \item \textbf{Last-step Option Disclosure Replay} (selection-only replay). We preserve the complete OE trajectory, $k=N$, disclosing the answer choices only in a new final \texttt{finish} call, where the agent is forced to select among them. 
\end{enumerate}

Replaying at both $\lfloor N/2 \rfloor$ and $N$ steps allows us to examine when answer options begin to affect an agentic trajectory, separating the effect of options on search and evidence gathering from their effect on final answer selection. 
In Last-step Option Disclosure (selection-only) replay, the search trajectory remains entirely option-free and options are introduced only when answering. 
This isolates the evaluation from the variability of LLM-as-a-judge grading or post-hoc matching on free-form text, providing a direct multiple-choice readout of what the option-free trajectory can support at selection time. 
Midpoint replay additionally exposes the continuation policy to the answer choices, allowing us to measure whether option disclosure alters subsequent tool use, requested video segments, stopping behaviour, and accuracy (\S\ref{sec:replay-results}).
As a control, Replay is also run without option disclosure to verify equivalence with the original OE run.

\subsection{Interpreting Evidence of Non-Visual Shortcutting and Memorization}
Our experiments test whether a substantial fraction of MC benchmark performance stems from prior exposure, language priors, and option artifacts rather than grounded video understanding. Because several frontier models are accessible only through proprietary APIs with unobservable training data and internal representations, it is impossible to directly verify whether a specific benchmark instance appeared during training. Moreover, any single intervention controls only a subset of confounding variables. We therefore rely on converging evidence across complementary perturbations: video removal and blanking to test visual necessity~\cite{chen_are_2024, accuracywithoutgrounding2026}, question removal to test label biases~\cite{gururangan_annotation_2018, zhang_mitigating_2025}, and option shuffling and numeric relabeling to test positional and format sensitivity~\cite{mcvqa_bias2023, loginova_addressing_2025, gupta_changing_2024, elhady_wicked_2025} (Table~\ref{tab:core}; Appendix~\ref{app:additional-results}).

A further limitation of current benchmarks is that questions vary widely in their degree of visual necessity. Poorly constructed items may leave a single candidate far more probable from language priors alone, while others require only a single frame rather than temporal reasoning. Standard evaluations collapse these regimes into a single accuracy figure; future benchmark design should explicitly stratify items across language-only, single-frame, short-clip, and whole-video understanding.

\subsection{Matching Open-Ended Answers to Multiple-Choice Options}
\label{sec:oe_mc_matching}

Finally, we triangulate whether the apparent MC advantage persists when the model itself never sees answer choices while solving the VQA task. We first execute the complete agentic process in OE format. During scoring, a separate LLM evaluator selects the option that most closely matches the OE response. The resulting option is then scored using the standard MC evaluation procedure for each dataset.

The evaluator does not receive the video and is not asked to solve the original VQA problem. Its role is solely to map the semantic content of the model's OE answer onto the corresponding MC option. 

If the difference between OE and MC evaluation is primarily a matter of answer representation, then mapping OE responses back to the candidate options should recover direct MC performance:
\[
\mathrm{Acc}_{\mathrm{OE}\rightarrow\mathrm{MC}} \approx \mathrm{Acc}_{\mathrm{MC}}.
\]
Conversely, a substantial remaining gap indicates that access to the answer choices during the VQA process itself improves performance, rather than merely making otherwise equivalent answers easier to score.

\begin{table}[!tbp]
  \centering
  \caption{\textbf{Multiple choice, open-ended, and the mechanisms that convert between
  them.} \emph{OE} scores free-text answers with two judges, \texttt{claude-sonnet-4-5} and
  \texttt{gpt-5.4-mini}, against the correct option's text (Appendix~\ref{app:judge}). 
  \emph{OE$\to$match} maps the prediction back onto the options, with
  the matcher's abstention rate in brackets (\texttt{abst}). MC, OE$\to$match and both
  replay columns are option-label accuracy. The two OE columns report the mean judge rating ($0$--$100$) followed in parentheses by the binarised accuracy (the percentage of items rated $\ge 80$, corresponding to ``mostly correct'' in Box~3); all MC$-$OE differences in the paper use this binarised accuracy (see Table~\ref{tab:judge-threshold} for alternative thresholds). 
  \emph{Replay} rewinds the rollout and
  discloses the options at its last step or halfway through. Note: Direct QA has no replay condition
  because it is never denied the options. Video-MME\,v2 is reported using flat per-question accuracy with its non-linear group score in parentheses, except in the OE columns where only per-question judge ratings apply. 
  Paired confidence intervals for the MC$-$OE and MC$-$replay differences are in Appendix~\ref{app:paired-ci}. 
  Observed evaluator variance is low: over the 77 cells here whose scorer was run three times, the run-to-run standard deviation has median $0.03$ and maximum $0.65$ accuracy points.}
  \label{tab:general-mcoe}
  \label{tab:mechanisms}
  \label{tab:one-number-five-ways}
  \label{tab:replay}
  \scriptsize
  \setlength{\tabcolsep}{2pt}
  \begin{tabular}{llcccccc}
    \toprule
    & & \multicolumn{3}{c}{Native scoring} & \multicolumn{3}{c}{Option-recovery mechanisms} \\
    \cmidrule(lr){3-5}\cmidrule(lr){6-8}
    Benchmark & Method & MC & OE (Claude) & OE (OpenAI) & OE$\to$match & Replay (last) & Replay (halfway) \\
    \midrule
    \multirow{4}{*}{NExT-QA (2021-05)}
      & Direct QA & 86.7 & 77.6 {\scriptsize (63.2)} & 63.6 {\scriptsize (54.0)} & 77.1 {\tiny [abst 6.9\%]} & n/a & n/a \\
      & Gaia loop & 85.5 & 72.7 {\scriptsize (55.5)} & 66.7 {\scriptsize (55.2)} & 78.4 {\tiny [abst 4.4\%]} & 85.5 & 85.1 \\
      & Author DVD & 81.2 & 71.6 {\scriptsize (58.6)} & 60.7 {\scriptsize (48.0)} & 72.9 {\tiny [abst 8.7\%]} & 81.3 & 82.7 \\
      & Author AVP & 83.1 & 68.0 {\scriptsize (49.4)} & 53.3 {\scriptsize (37.4)} & 72.1 {\tiny [abst 10.7\%]} & 82.3 & 82.5 \\
    \midrule
    \multirow{4}{*}{EgoSchema (2023-08)}
      & Direct QA & 80.6 & 58.2 {\scriptsize (22.6)} & 58.0 {\scriptsize (30.2)} & 73.9 {\tiny [abst 5.5\%]} & n/a & n/a \\
      & Gaia loop & 81.0 & 55.1 {\scriptsize (18.2)} & 48.7 {\scriptsize (24.2)} & 72.8 {\tiny [abst 4.0\%]} & 80.8 & 80.2 \\
      & Author DVD & 73.8 & 54.1 {\scriptsize (25.4)} & 47.5 {\scriptsize (25.6)} & 67.3 {\tiny [abst 14.7\%]} & 75.0 & 75.8 \\
      & Author AVP & 71.8 & 47.6 {\scriptsize (9.2)} & 44.7 {\scriptsize (15.4)} & 72.1 {\tiny [abst 7.1\%]} & 69.6 & 71.0 \\
    \midrule
    \multirow{4}{*}{LVBench (2024-06)}
      & Direct QA & 84.1 & 74.6 {\scriptsize (63.8)} & 65.2 {\scriptsize (56.0)} & 67.7 {\tiny [abst 12.6\%]} & n/a & n/a \\
      & Gaia loop & 82.3 & 71.0 {\scriptsize (58.7)} & 62.6 {\scriptsize (53.3)} & 67.8 {\tiny [abst 10.0\%]} & 79.2 & 81.1 \\
      & Author DVD & 73.7 & 61.4 {\scriptsize (50.8)} & 56.0 {\scriptsize (46.0)} & 58.3 {\tiny [abst 18.8\%]} & 70.3 & 75.0 \\
      & Author AVP & 76.3 & 60.9 {\scriptsize (47.2)} & 51.0 {\scriptsize (38.9)} & 56.5 {\tiny [abst 19.7\%]} & 77.6 & 77.7 \\
    \midrule
    \multirow{4}{*}{QBench-Video (2024-09)}
      & Direct QA & 74.7 & 70.5 {\scriptsize (64.2)} & 66.7 {\scriptsize (60.3)} & 66.2 {\tiny [abst 5.4\%]} & n/a & n/a \\
      & Gaia loop & 68.3 & 63.6 {\scriptsize (51.6)} & 63.1 {\scriptsize (55.2)} & 60.8 {\tiny [abst 3.2\%]} & 65.6 & 68.1 \\
      & Author DVD & 60.9 & 58.5 {\scriptsize (51.0)} & 55.6 {\scriptsize (49.0)} & 52.1 {\tiny [abst 8.5\%]} & 57.9 & 57.9 \\
      & Author AVP & 77.7 & 67.4 {\scriptsize (56.3)} & 60.7 {\scriptsize (51.6)} & 62.6 {\tiny [abst 5.8\%]} & 76.4 & 75.9 \\
    \midrule
    \multirow{4}{*}{Perception Test (2025-07)}
      & Direct QA & 82.6 & 72.1 {\scriptsize (65.6)} & 72.3 {\scriptsize (64.5)} & 69.8 {\tiny [abst 9.2\%]} & n/a & n/a \\
      & Gaia loop & 80.6 & 71.7 {\scriptsize (64.1)} & 74.8 {\scriptsize (70.9)} & 73.6 {\tiny [abst 8.2\%]} & 78.9 & 81.1 \\
      & Author DVD & 69.8 & 56.8 {\scriptsize (48.2)} & 57.3 {\scriptsize (48.9)} & 56.4 {\tiny [abst 18.9\%]} & 68.9 & 70.5 \\
      & Author AVP & 78.2 & 64.8 {\scriptsize (56.8)} & 65.0 {\scriptsize (56.8)} & 67.7 {\tiny [abst 10.8\%]} & 77.1 & 78.2 \\
    \midrule
    \multirow{4}{*}{Video-MME\,v2 (2026-04)}
      & Direct QA & 68.2 {\scriptsize (51.5)} & 62.9 {\scriptsize (43.6)} & 55.4 {\scriptsize (38.0)} & 49.3 {\scriptsize (30.3)}\,{\tiny [abst 10.1\%]} & n/a & n/a \\
      & Gaia loop & 67.8 {\scriptsize (50.7)} & 61.2 {\scriptsize (41.6)} & 52.8 {\scriptsize (35.6)} & 47.2 {\scriptsize (28.4)}\,{\tiny [abst 8.9\%]} & 66.6 {\scriptsize (49.0)} & 66.6 {\scriptsize (49.1)} \\
      & Author DVD & 45.0 {\scriptsize (27.7)} & 47.7 {\scriptsize (31.2)} & 43.0 {\scriptsize (26.9)} & 34.1 {\scriptsize (16.9)}\,{\tiny [abst 20.6\%]} & 44.7 {\scriptsize (26.6)} & 47.5 {\scriptsize (29.9)} \\
      & Author AVP & 59.4 {\scriptsize (41.9)} & 51.2 {\scriptsize (27.0)} & 44.9 {\scriptsize (26.2)} & 42.1 {\scriptsize (23.6)}\,{\tiny [abst 14.7\%]} & 58.1 {\scriptsize (39.7)} & 58.8 {\scriptsize (40.9)} \\
    \bottomrule
  \end{tabular}
\end{table}

\section{Results and Discussion}
\label{sec:results}

\begin{table}[!tbp]
  \centering
  \caption{\textbf{How much multiple-choice accuracy survives without the evidence it
  should require.} Each column withholds or perturbs a single input while holding all other settings fixed relative to \emph{MC (full)}. 
  \emph{Chance} and \emph{Heuristic} use no model at all (Appendix~\ref{app:heuristic}). \emph{No question}
  hides the question \emph{and} the options, leaving the video and an instruction to emit a
  single letter, so it measures the label prior rather than a question-only baseline
  (\S\ref{app:additional-results}). As expected, it scores below \emph{Heuristic} on all six benchmarks. \emph{No video} hides the video, and \emph{Black} replaces the video with
  a blank one. \emph{Shuffle} permutes the option order. \emph{Numeric} relabels the
  options as digits. Splits and scorers are in Table~\ref{tab:benchmarks}, and
  Appendix~\ref{app:paired-ci} gives confidence intervals for the no-video, MC$-$OE and replay contrasts. Without video, models retain roughly half of their full MC accuracy on average ($51\%$ for direct QA and $50\%$ for the Gaia loop, using flat per-question accuracy throughout). After correcting for random chance, no-video performance retains $29\%$ of the available headroom above chance for direct QA ($26\%$ for the Gaia loop), ranging from $45\%$ on EgoSchema to $12\%$ on Perception Test. Shuffling or relabelling options has little effect.}
  \label{tab:core} 
  \label{tab:memorization}
  \scriptsize
  \setlength{\tabcolsep}{2.5pt}

  \begin{tabular}{lcccccccc}
    \toprule
    Benchmark & Chance & Heuristic & No question & No video & Black & MC (full) & Shuffle & Numeric \\
              &        & \emph{(no model)} & & & & & & \\
    \midrule
    \multicolumn{9}{l}{\emph{Direct QA (Gemini 3.6 Flash)}} \\
    NExT-QA (2021-05) & 20.0 & 23.0 & 20.5 & 40.3 & 42.5 & 86.7 & 87.1 & 87.0 \\
    EgoSchema (2023-08) & 20.0 & 29.7 & 19.6 & 47.6 & 48.4 & 80.6 & 80.2 & 81.2 \\
    LVBench (2024-06) & 25.0 & 29.2 & 24.4 & 40.4 & 37.6 & 84.1 & 84.2 & 83.7 \\
    QBench-Video (2024-09) & 37.3 & 40.2 & 37.7 & 49.4 & 53.5 & 74.7 & 74.3 & 74.8 \\
    Perception Test (2025-07) & 31.6 & 35.7 & 29.1 & 37.8 & 37.2 & 82.6 & 83.9 & 82.8 \\
    Video-MME\,v2 (2026-04) & 12.8 {\tiny (3.5)} & 19.9 {\tiny (7.2)} & 12.4 {\tiny (3.3)} & 26.7 {\tiny (10.2)} & 26.5 {\tiny (10.0)} & 68.2 {\tiny (51.5)} & 68.3 {\tiny (51.5)} & 68.2 {\tiny (52.1)} \\
    \midrule
    \multicolumn{9}{l}{\emph{Gaia agentic loop}} \\
    NExT-QA (2021-05) & 20.0 & 23.0 & --   & 42.2 & --   & 85.5 & --   & 85.3 \\
    EgoSchema (2023-08) & 20.0 & 29.7 & --   & 44.4 & --   & 81.0 & --   & 80.4 \\
    LVBench (2024-06) & 25.0 & 29.2 & --   & 38.3 & --   & 82.3 & --   & 82.0 \\
    QBench-Video (2024-09) & 37.3 & 40.2 & --   & 42.7 & --   & 68.3 & --   & 69.1 \\
    Perception Test (2025-07) & 31.6 & 35.7 & --   & 37.2 & --   & 80.6 & --   & 81.7 \\
    Video-MME\,v2 (2026-04) & 12.8 {\tiny (3.5)} & 19.9 {\tiny (7.2)} & --   & 27.7 {\tiny (10.9)} & --   & 67.8 {\tiny (50.7)} & --   & 67.5 {\tiny (50.6)} \\
    \bottomrule
  \end{tabular}
  \\[2pt]
  \scriptsize
\end{table}

\subsection{Multiple-choice and open-ended evaluation}

As expected, multiple-choice (MC) and open-ended (OE) evaluations produce substantially different scores. As MC provides candidate answers and therefore presents an easier evaluation setting, it scores higher on every dataset and for every method once open-ended answers are put on the same per-item correctness scale (Table~\ref{tab:general-mcoe}, bracketed figures).

We test whether this gap can be recovered by mapping OE responses to the best MC option match. Under direct QA this recovers only part of the difference and stays below native MC (Table~\ref{tab:replay}). Under the agentic methods it behaves differently: on NExT-QA and EgoSchema, matching scores above both judges for all three methods, so what the matcher recovers depends on the inference regime and not on the scoring step alone. The gap cannot be explained solely by answer formatting or poorer evaluation. 

We consider three sources of uncertainty in the results. Evaluator variance is low: across three LLM-as-a-judge based scoring runs, the median standard deviation is $0.03$ accuracy points and the maximum is $0.65$. We quantify uncertainty from the finite evaluation set using $95\%$ paired bootstrap confidence intervals for the MC$-$OE and MC$-$replay differences (Appendix~\ref{app:paired-ci}), resampling videos rather than individual questions to preserve dependence between questions from the same video. We report generation seed variability separately in \S\ref{app:seed-variance}. Finally, judge choice has a substantially larger effect, shifting scores for the same generations by up to $14.7$ rating points, or $11.9$ once binarised; we therefore report results from both judges explicitly rather than incorporating judge variation into the confidence intervals.

\subsection{Grounding and Answer-Format Diagnostics}

On all six benchmarks, accuracy remains above chance without visual input (Table~\ref{tab:core}, Fig.~\ref{fig:context-ablation}).
EgoSchema, NExT-QA, LVBench, Video-MME\,v2 and QBench-Video achieve blind accuracies $27.6$, $20.3$, $15.4$, $13.9$ and $12.1$ points above chance, respectively, showing that large non-visual baselines persist. Perception Test has the smallest margin, $+6.2$.
A training-free, model-free heuristic over the question stem and option texts alone scores $2.9$--$9.7$ points above chance (Appendix~\ref{app:heuristic}). This is a floor, not a ceiling, on what is available without the video: given the same stem and options, the no-video condition exceeds it on all six benchmarks, by $10.8$ points on average for direct QA and $9.1$ for the Gaia loop.
On the no-video variant, QBench-Video's evaluator implementation scores $17.3$ compared to $50.8$ on the same run with an answer extractor defined in \S\ref{sec:mc-oe} (five-seed mean $49.4$ in Table~\ref{tab:core}). The gap is format compliance, not accuracy: $562$ of the $868$ predictions are a parenthesised letter the released protocol does not accept. Because the effect is a property of output format rather than of selection, we report the permissive score throughout.

Shuffling answer choices or replacing letter labels with numeric labels has minimal effect on accuracy: neither moves any benchmark by more than $1.3$ points~\cite{gupta_changing_2024, elhady_wicked_2025}. Removing the question eliminates most of the blind advantage, reducing all six benchmarks to approximately random chance (QBench-Video $37.7$ against $37.3$).

Replacing each video with a length-matched black screen at $0.1$\,fps (against $1$\,fps elsewhere) produces nearly the same performance as withholding the video entirely, remaining far above chance on EgoSchema ($48.4$ vs.\ $20.0$), NExT-QA ($42.5$ vs.\ $20.0$), QBench-Video ($53.5$ vs.\ $37.3$), LVBench ($37.6$ vs.\ $25.0$), and Video-MME\,v2 ($26.5$ vs.\ $12.8$), with Perception Test the lone smaller margin ($37.2$ vs.\ $31.6$). 
This reproduces prior findings that non-visual signals support considerable VideoQA accuracy~\cite{ok_tempcore_2026}.

\subsection{Replay and Option-Delayed Evaluation}
\label{sec:replay-results}

Replay experiments test whether adding MC options gives the model extra context that can be used in its evidence collection process. 
Revealing choices halfway through, or at the final step of a trajectory, impacts the agentic methods differently (Table~\ref{tab:replay}). 
DVD appears more sensitive to when answer choices are introduced than AVP. 
Halfway replay moves DVD by more than a point on five of six benchmarks, Perception Test being the exception, and three of those intervals exclude zero: replay helps on NExT-QA and Video-MME\,v2 and hurts on QBench-Video (Table~\ref{tab:paired-ci}). AVP stays within a point of native performance on four of six. Last-step replay tends to hurt both approaches. This can be explained by how the methods work. DVD has access to global caption information allowing grounding to happen at any point. On the other hand, AVP acquires visual evidence sequentially, so halfway through the trajectory much of its perception budget has already been committed and introducing the options has limited ability to change the evidence gathered.

At the step level, disclosing options mid-search does not make the agent stop early or inspect different parts of the video ($0$\,s median start-time shift), but it changes which tool it calls: \texttt{clip\_search} rises from $4.0\%$ to $10.1\%$ of next actions while \texttt{frame\_inspect} falls from $76.2\%$ to $70.2\%$. Using the question categories of \textcite{vbenchcomp}, this increase in search effort is roughly twice as large on Temporal questions as on Text or Semantic ones.

Replay nearly closes the gap to native MC evaluation even when the options are introduced only at the final step. Across all $18$ agentic benchmark--method pairs, last-step replay falls within $3.4$ points of native MC performance, exceeds post-hoc $\text{OE}\rightarrow\text{match}$ on $17$ of the $18$ pairs, and on the Gaia loop recovers $79$--$100\%$ of the MC$-$OE gap (see Appendix~\ref{app:paired-ci} for replay controls and matcher abstention details).
Most of the MC advantage arises at answer time rather than from option-guided search, because last-step replay and $\text{OE}\rightarrow\text{match}$ share the same option-free evidence-gathering trajectory and differ only in whether the agent selects among the choices at the end.
One caveat is that the \texttt{finish} call sees the options with the trajectory still in context, so last-step replay separates search-time from answer-time use of the options, rather than pure label selection from answer-time reasoning. 
Furthermore, because the final step still selects among the MC options, it retains whatever non-visual option priors the no-video baseline exposes; ODR therefore ensures faithful, option-free \emph{search}, while operating within the standard MC option space at selection time.

\subsection{Interpreting the Evidence Together}

Visual evidence contributes substantially to performance, but so do the semantic relationship between questions and candidate answers, when those candidates are exposed, and how responses are evaluated. On several benchmarks, a large fraction of accuracy remains available without the video, while removing the question reduces this advantage to approximately chance. Replay experiments further show that access to answer choices during inference can change performance, with the effect depending on when choices are introduced and how the agent gathers evidence. Consequently, similar headline accuracies can reflect materially different combinations of visual grounding, non-visual benchmark signal, option-conditioned reasoning, and scoring behaviour. Reporting these components separately provides a more interpretable account of what improvements in VideoQA accuracy actually measure.
Appendix~\ref{app:limitations} sets out what these measurements do not establish.

\Needspace{22\baselineskip}
\section{Conclusion}

{\interlinepenalty=10000
Across six VideoQA benchmarks and four inference methods, our audit demonstrates that headline multiple-choice accuracy is a composite of visual perception, language priors, option-timing effects, dataset quality, and scoring mechanics.
Models retain roughly half of their full MC accuracy without seeing a single frame of video, while open-ended evaluation exposes severe sensitivity to judge selection and post-hoc answer matching.
For agentic systems, Option-Delayed Replay (ODR) reveals that nearly the entire multiple-choice advantage stems from generation-time selection rather than superior visual search. Withholding options throughout evidence acquisition and revealing them only at the final step recovers $79$--$100\%$ of the MC$-$OE gap on our primary loop and lands within $3.4$ points of native MC across all $18$ agentic evaluations.\par\nopagebreak

A single VideoQA leaderboard number therefore cannot, on its own, justify trusting a model to watch and reason faithfully in deployment.
Until the field transitions toward contamination-resistant, task-centric, non-MC evaluations, we recommend that VideoQA benchmarks and model evaluations report a minimal faithfulness audit: (1)~mandatory blind and model-free heuristic baselines to expose non-visual solvability, (2)~decoupled, permissive answer extraction so format compliance is never conflated with perception, (3)~Option-Delayed Replay for agentic systems to ensure evidence gathering remains unconditioned by candidate answers, and (4)~human-verified ground-truth keys, as exemplified by our released re-annotation of LVBench.\par}

\section*{Author Contributions}
{\small Elliott Piercy and Sepand Dyanatkar jointly led the research, contributing equally to its conceptualization, methodology, experimental design, implementation and execution, analysis, and manuscript preparation.
Sepand Dyanatkar additionally provided overall research direction, technical leadership, and supervision.
Justin Chan contributed to experimental implementation and execution, as well as manuscript writing.\par}

\small
\printbibliography

\normalsize
\clearpage

\appendix
\section{Benchmark Details}
\label{app:benchmarks}

Each of the six benchmarks is evaluated on one fixed split file that every condition in
this paper shares, so all conditions are paired on the identical question set.
Table~\ref{tab:benchmarks} states that split, its size, its per-question option
count, the modality actually supplied to the model, and the scorer whose number
the main text reports.

Video-MME\,v2's primary reported metric is the flat accuracy over individual multiple-choice questions, with its native non-linear group score given as the secondary figure in brackets. QBench-Video's $868$ MC questions are the \texttt{what/how} and \texttt{yes-or-no} types of the $1{,}192$-question sim+real merge. The remaining $324$ real-OE questions are excluded from every reported cell.

\begin{table}[!tbp]
  \centering
  \caption{\textbf{The evaluated split of each benchmark.} Every condition in this paper
  runs the same split file, so all conditions are paired on identical questions. \emph{Chance}
  is the exact analytic baseline over the split's per-question option counts, which are
  fixed on three benchmarks and mixed on the rest, and \emph{MC acc.}\ is the sighted
  direct-QA score of Table~\ref{tab:core} under the scorer beside it.}
  \label{tab:benchmarks}
  \scriptsize
  \setlength{\tabcolsep}{3pt}
  \begin{tabular}{llrrlll}
    \toprule
    Benchmark & Evaluated split & $n$ & Chance & Options & Modality & Scoring (MC acc.) \\
    \midrule
    NExT-QA         & validation ($570$ videos)        & 4{,}996 & 20.0 & 5            & video\,+\,audio & author MC exact match ($86.7$) \\
    EgoSchema       & public labelled subset           &  500    & 20.0 & 5            & video only      & MC match + extractor fallback ($80.6$) \\
    LVBench         & full released set                & 1{,}549 & 25.0 & 4            & video\,+\,audio & MC match + extractor fallback \texttt{acc} ($84.1$) \\
    QBench-Video    & MC subset of the $1{,}192$ merge &  868    & 37.3 & 2--4         & video\,+\,audio & MC match + extractor fallback ($74.7$) \\
    Perception Test & \texttt{mc\_vqa\_unified\_valid} &  454    & 31.6 & 3\,/\,5      & video\,+\,audio & author MC exact match ($82.6$) \\
    Video-MME\,v2   & $800$ four-question groups       & 3{,}200 & 12.8 & 2--8         & video\,+\,audio & flat $68.2$ (group $51.5$) \\
    \bottomrule
  \end{tabular}
\end{table}

\section{Limitations}
\label{app:limitations}

Our diagnostics observe model inputs, tool calls, and final answers rather than internal reasoning: a trajectory that selects the correct option for unfaithful reasons still scores as correct, meaning our interventions bound what information an answer could draw upon rather than verifying the reasoning trace itself. Additionally, direct QA is evaluated on a single backbone (Gemini 3.6 Flash), while Author-DVD and Author-AVP use the backbones specified by their respective releases (OpenAI o3 and Gemini 2.5 Pro). Consequently, agent architecture effects are not isolated from backbone differences.
Apart from the five-seed no-video runs of Appendix~\ref{app:seed-variance}, whose seed
spread is $0.16$ to $1.20$ points, every cell is a single run.
On the open-ended columns, scoring choices outweigh sampling: the two judges differ by up to
$11.9$ points on identical generations once binarised, and the binarisation threshold moves
the mean MC$-$OE gap from $+13.0$ to $+42.2$ points (Table~\ref{tab:judge-threshold}).
The systems and one judge are proprietary APIs, so a provider-side model update could move
any number here undetectably.

\section{LVBench Relabelled Dataset}
\label{app:lvbench-relabel}
We release a relabelled annotation set for the LVBench dataset~\cite{lvbench2025}. All 1,549 LVBench multiple-choice questions (103 videos) were independently re-annotated by a third-party vendor. Thirty-four annotators took part. Of the 1,549 items, 19 were excluded as unanswerable, leaving 1,530 for correction or keeping. Table~\ref{tab:lvbench-relabel}'s second column drops those 19 for a fairer comparison.
Each annotator wrote a free-text answer and cited an evidence window in the source video. Then the existing key was shown to them, and they selected an existing option matching their answer if possible.

Relative to the published LVBench keys, content changed on 466 of 1,530 items: the question stem on 137, the distractors on 356, and the correct answer on 313. The correct-answer count compares option \emph{text} rather than the option letter, so reordering the options is not counted as a change, and it normalises case and punctuation so that typography fixes are not counted either.
Of those 313, 94 were relabelled to a different original option and 219 to an option whose text the vendor rewrote.
Scoring against the corrected keys improves accuracy across every evaluation condition while leaving the gap between sighted and blind performance intact.
The agentic loop gains more than direct QA, overtaking it where comparable: baseline multiple choice (agentic $82.2 \to 86.3$ against direct QA $84.1 \to 84.9$) and OE$\to$match ($66.9 \to 70.5$ against $67.7 \to 70.2$), on the $1{,}530$-item column of Table~\ref{tab:lvbench-relabel}.
Note that we primarily report the relabel as a measurement of LVBench's key quality and not as evidence about system ranking.

\textbf{Data availability.}
Our corrected LVBench annotations, adjudication metadata, evidence windows, and exclusion notes are available at \url{https://github.com/OnDeckAI/LVBench-Relabel}.

\begin{table}[!tbp]
\centering
\caption{\textbf{LVBench accuracy under the published and the corrected keys.}
Every condition is scored three ways: against the published keys on the full set, against
the published keys once the $19$ unanswerable items are dropped, and against the corrected
keys on that same reduced set. Every cell uses the benchmark's released scorer, as in
Table~\ref{tab:general-mcoe}. The middle column isolates the effect of correcting the
keys from the effect of removing the items.}
\label{tab:lvbench-relabel}
\scriptsize
\begin{tabular}{lccc}
\toprule
 & Published keys & Published keys & Corrected keys \\
 & full set & answerable only & answerable only \\
 & ($n{=}1{,}549$) & ($n{=}1{,}530$) & ($n{=}1{,}530$) \\
\midrule
Direct QA --- no video    & 40.61 & 40.65 & 41.76 \\
Gaia loop --- no video    & 38.35 & 38.43 & 39.72 \\
Direct QA --- baseline    & 84.12 & 84.12 & 84.90 \\
Gaia loop --- baseline & 82.31 & 82.22 & 86.27 \\
Replay --- halfway        & 81.11 & 81.31 & 83.33 \\
Replay --- last step      & 79.17 & 79.15 & 80.94 \\
Direct QA --- OE$\to$match    & 67.66 & 67.73 & 70.24 \\
Gaia loop --- OE$\to$match & 67.76 & 66.87 & 70.46 \\
\bottomrule
\end{tabular}
\end{table}

\section{Details of Context and Option-Channel Ablations}
\label{app:additional-results}
Below we detail the experimental setup for each context and option-channel perturbation reported in Table~\ref{tab:core}.

Direct QA results for a wider range of proprietary and open-weight models are available on benchmarks \cite{videomme2025,videommev2_2026,lvbench2025,egoschema2023,longvideobench2024,mlvu2025,mmbenchvideo2024,videoevalpro2026}. Our goal is not to rank model families but to test whether evaluation-format effects recur under heterogeneous inference procedures. We therefore hold each system fixed within paired conditions and report effects separately for direct inference and three agentic implementations.

\subsection{Video removal and blanking}
For no-video eval, we omit the video entirely, only prompting each system with the question and answer options. For blank/black screen eval, video is turned entirely black and audio is silenced \cite{ok_tempcore_2026, accuracywithoutgrounding2026}. Any performance above random chance is indicative of benchmark shortcutting~\cite{chen_are_2024, buch_revisiting_2022, loginova_addressing_2025}.

\subsection{Question Removal}
We withhold the question and the answer choices alike, leaving the video and an instruction to reply with a single option letter. Question removal therefore measures label letter bias compared to chance, which happens when a benchmark's keys are unevenly distributed over positions or when a system favours a particular letter~\cite{gururangan_annotation_2018, niven_probing_2019, zhang_mitigating_2025}.

\subsection{Shuffle options}
We randomly permute the ordering of the answer choices while preserving their contents. A model that can identify the correct answer from its content instead of the related QA index should remain robust. Any degradation in performance indicates sensitivity to the original answer positioning consistent with answer shortcutting~\cite{mcvqa_bias2023, loginova_addressing_2025, zhang_mitigating_2025}.

\subsection{Numeric labels}
We replace the standard alphabetical answer labels (A/B/C/D) with numeric labels (1/2/3/4), while leaving the question, video, and answer contents unchanged. A large performance change indicates the predictions depend on language representation rather than the sole semantic relationship between question and answer~\cite{videoqa_llm_empirical2025, mcvqa_bias2023}.

\section{Paired Confidence Intervals on the Principal Contrasts}
\label{app:paired-ci}

\textbf{Statistical uncertainty.}
Every within-benchmark comparison is paired, since the conditions run the same items. For
the principal contrasts we report 95\% confidence intervals on the score difference from
$10{,}000$ paired bootstrap replicates, resampling videos with replacement, keeping all of
a sampled video's questions, and recomputing the benchmark's own metric.
Videos are the unit of resampling, because four of the six datasets put $>1$ question on one
video. Treating questions as independent would understate those intervals.

These bootstrap intervals capture finite-sample variability across videos, separate from evaluator variance and generation seed variance.
Scoring an individual item is itself uncertain, because an extractor or a judge decides many of them. 
Across the $77$ cells the pipeline re-scored three times that variability is small, a median standard
deviation of $0.03$ points and a maximum of $0.65$ (Table~\ref{tab:general-mcoe}).
Seed-to-seed variation is measured separately in \S\ref{app:seed-variance}, and the
MC$-$OE column also inherits the judge, which Table~\ref{tab:general-mcoe} shows
moving the same generations by up to $11.9$ points once binarised. No interval carries a multiplicity
correction.

\textbf{Extraction.} The multiple-choice extractor (Box~2, Appendix~\ref{app:prompts}) runs at temperature $0$ on a pinned model, \texttt{claude-sonnet-4-5-20250929}; re-scoring the NExT-QA Gaia-loop predictions, $300$ of which pass through it, reproduces the printed cell exactly. Numeric-label runs map each digit back to its letter before scoring. Scoring the digits directly gives $83.9$ rather than $85.3$ (direct QA: $87.0$ either way).

\textbf{Controls on the replay procedure.}
We additionally explored whether the recovered accuracy comes from disclosing the options or from the act of resuming (last-step replay resumes a checkpointed trajectory). 
In the first control, we resumed the same open-ended trajectory but withheld the options, matching the free-text answer onto them afterwards instead of replaying, and accuracy did not shift systematically.
In the second, we ran open-ended from scratch and revealed the options only at the \texttt{finish} call, which removes checkpointing entirely. Scores were a fraction above last-step replay.

The relevant comparison we produce is last-step replay against OE$\to$match: the two reach the options by different routes from the same open-ended trajectory, and replay is higher on $17$ of the $18$ agentic pairs of Table~\ref{tab:general-mcoe} (EgoSchema under Author-AVP being the exception). OE$\to$matching abstentions are scored as incorrect ($3.2$--$20.6\%$ of answers), and replay never abstains. 

\begin{table}[!tbp]
  \centering
  \caption{\textbf{Paired $95\%$ confidence intervals on the principal contrasts.} Each
  entry is that row's multiple-choice score minus the named condition, in accuracy
  points, with its video-clustered paired bootstrap interval. A positive number means MC
  scores higher. 
  Per-question correctness is computed using the same evaluators as Tables~\ref{tab:general-mcoe} and~\ref{tab:core}. Each interval is computed over the intersection of items completed by both conditions. This equals the full split everywhere except LVBench open-ended ($1{,}477$ of $1{,}549$ items) and the two QBench-Video Author-AVP replay conditions ($622$ of $868$ items), where differences are taken over the shared subset. The two OE columns take the
  difference against the judge binarised at $\ge 80$ (\S\ref{app:judge}), not against the
  mean rating, so both sides of every entry are per-item correctness on one scale; the
  bracketed figures in Table~\ref{tab:general-mcoe} are these binarised scores. The two
  judges still disagree, by as much as $11.9$ points on the same generations after
  binarising, which is why they are reported apart. On Video-MME\,v2 both OE columns are on
  the flat per-question scale, because the judges are defined only there, while its no-video
  and replay columns are the group score over $800$ four-question
  groups. ``--'' marks a condition that was never run: direct QA is never denied the
  options, so it has no replay variant, and the two author implementations were not run
  without video.}
  \label{tab:paired-ci}
  \scriptsize
  \setlength{\tabcolsep}{3pt}
  \begin{tabular}{llccccc}
    \toprule
    & & \multicolumn{5}{c}{$\Delta$ MC} \\
    \cmidrule(lr){3-7}
    Benchmark & Method & no video & OE$_{\ge 80}$ (Claude) & OE$_{\ge 80}$ (OpenAI) & replay (last) & replay (half) \\
    \midrule
    \multirow{4}{*}{NExT-QA}
      & Direct QA & $+46.4$ {\tiny $[45.0,47.8]$} & $+23.5$ {\tiny $[22.0,25.0]$} & $+32.7$ {\tiny $[31.1,34.2]$} & -- & -- \\
      & Gaia loop & $+43.3$ {\tiny $[41.7,44.9]$} & $+30.0$ {\tiny $[28.3,31.5]$} & $+30.3$ {\tiny $[28.7,31.9]$} & $-0.0$ {\tiny $[-0.8,0.8]$} & $+0.4$ {\tiny $[-0.5,1.2]$} \\
      & Author DVD & -- & $+22.6$ {\tiny $[21.2,24.1]$} & $+33.2$ {\tiny $[31.8,34.7]$} & $-0.1$ {\tiny $[-1.1,0.9]$} & $-1.5$ {\tiny $[-2.4,-0.6]$} \\
      & Author AVP & -- & $+33.7$ {\tiny $[32.1,35.4]$} & $+45.7$ {\tiny $[44.1,47.3]$} & $+0.9$ {\tiny $[-0.0,1.8]$} & $+0.7$ {\tiny $[-0.2,1.5]$} \\
    \midrule
    \multirow{4}{*}{EgoSchema}
      & Direct QA & $+33.0$ {\tiny $[28.2,37.8]$} & $+58.0$ {\tiny $[53.4,62.6]$} & $+50.4$ {\tiny $[45.6,55.0]$} & -- & -- \\
      & Gaia loop & $+36.6$ {\tiny $[31.6,41.6]$} & $+62.8$ {\tiny $[58.4,67.2]$} & $+56.8$ {\tiny $[52.2,61.4]$} & $+0.2$ {\tiny $[-2.8,3.2]$} & $+0.8$ {\tiny $[-1.8,3.6]$} \\
      & Author DVD & -- & $+48.4$ {\tiny $[43.6,53.2]$} & $+48.2$ {\tiny $[43.4,53.0]$} & $-1.2$ {\tiny $[-4.4,2.0]$} & $-2.0$ {\tiny $[-5.2,1.2]$} \\
      & Author AVP & -- & $+62.6$ {\tiny $[58.2,67.0]$} & $+56.4$ {\tiny $[51.8,60.8]$} & $+2.2$ {\tiny $[-0.8,5.4]$} & $+0.8$ {\tiny $[-2.2,3.8]$} \\
    \midrule
    \multirow{4}{*}{LVBench}
      & Direct QA & $+43.7$ {\tiny $[41.3,46.2]$} & $+20.6$ {\tiny $[17.7,23.6]$} & $+28.1$ {\tiny $[25.1,31.0]$} & -- & -- \\
      & Gaia loop & $+44.1$ {\tiny $[41.0,47.3]$} & $+23.6$ {\tiny $[20.5,26.7]$} & $+29.1$ {\tiny $[26.2,32.0]$} & $+3.2$ {\tiny $[1.6,4.8]$} & $+1.2$ {\tiny $[-0.4,2.7]$} \\
      & Author DVD & -- & $+22.9$ {\tiny $[20.2,25.6]$} & $+27.7$ {\tiny $[25.0,30.4]$} & $+3.4$ {\tiny $[1.4,5.4]$} & $-1.3$ {\tiny $[-3.3,0.7]$} \\
      & Author AVP & -- & $+29.1$ {\tiny $[26.3,31.9]$} & $+37.4$ {\tiny $[34.3,40.5]$} & $-1.3$ {\tiny $[-3.4,0.9]$} & $-1.4$ {\tiny $[-3.3,0.5]$} \\
    \midrule
    \multirow{4}{*}{QBench-Video}
      & Direct QA & $+25.3$ {\tiny $[21.8,28.7]$} & $+10.5$ {\tiny $[7.5,13.5]$} & $+14.4$ {\tiny $[11.3,17.6]$} & -- & -- \\
      & Gaia loop & $+25.6$ {\tiny $[21.7,29.5]$} & $+16.7$ {\tiny $[13.5,19.9]$} & $+13.1$ {\tiny $[10.0,16.3]$} & $+2.8$ {\tiny $[0.5,5.2]$} & $+0.2$ {\tiny $[-2.2,2.8]$} \\
      & Author DVD & -- & $+9.9$ {\tiny $[6.2,13.6]$} & $+12.0$ {\tiny $[8.3,15.7]$} & $+3.0$ {\tiny $[0.0,6.0]$} & $+3.0$ {\tiny $[0.1,5.8]$} \\
      & Author AVP & -- & $+21.3$ {\tiny $[18.1,24.7]$} & $+26.0$ {\tiny $[22.7,29.6]$} & $+1.8$ {\tiny $[-0.8,4.3]$} & $+2.3$ {\tiny $[-0.3,5.0]$} \\
    \midrule
    \multirow{4}{*}{Perception Test}
      & Direct QA & $+44.8$ {\tiny $[40.6,48.9]$} & $+17.0$ {\tiny $[12.3,21.8]$} & $+18.1$ {\tiny $[13.3,23.1]$} & -- & -- \\
      & Gaia loop & $+43.4$ {\tiny $[37.7,49.0]$} & $+16.5$ {\tiny $[12.3,21.1]$} & $+9.7$ {\tiny $[5.7,13.9]$} & $+1.8$ {\tiny $[-1.7,5.3]$} & $-0.4$ {\tiny $[-3.7,2.8]$} \\
      & Author DVD & -- & $+21.6$ {\tiny $[16.3,27.2]$} & $+20.9$ {\tiny $[15.4,26.7]$} & $+0.9$ {\tiny $[-2.7,4.5]$} & $-0.7$ {\tiny $[-4.5,3.2]$} \\
      & Author AVP & -- & $+21.4$ {\tiny $[16.4,26.5]$} & $+21.4$ {\tiny $[16.4,26.6]$} & $+1.1$ {\tiny $[-2.6,4.9]$} & $+0.0$ {\tiny $[-3.4,3.4]$} \\
    \midrule
    \multirow{4}{*}{Video-MME\,v2}
      & Direct QA & $+41.4$ {\tiny $[38.8,44.0]$} & $+24.7$ {\tiny $[22.5,26.8]$} & $+30.2$ {\tiny $[28.0,32.5]$} & -- & -- \\
      & Gaia loop & $+39.8$ {\tiny $[37.1,42.5]$} & $+26.2$ {\tiny $[24.0,28.3]$} & $+32.2$ {\tiny $[29.9,34.4]$} & $+1.6$ {\tiny $[-0.1,3.4]$} & $+1.6$ {\tiny $[-0.2,3.3]$} \\
      & Author DVD & -- & $+13.8$ {\tiny $[11.9,15.9]$} & $+18.2$ {\tiny $[16.1,20.2]$} & $+1.2$ {\tiny $[-0.6,3.0]$} & $-2.1$ {\tiny $[-3.8,-0.4]$} \\
      & Author AVP & -- & $+32.4$ {\tiny $[30.2,34.6]$} & $+33.2$ {\tiny $[30.9,35.4]$} & $+2.2$ {\tiny $[0.4,3.9]$} & $+1.0$ {\tiny $[-0.8,2.7]$} \\
    \bottomrule
  \end{tabular}
\end{table}

In general, black-screen video scores similarly to no video, within $4.1$ points of it on every
benchmark and within $2.8$ outside QBench-Video. Shuffling the option order has a range of at most $1.3$ points. Relabelling the options as digits has a range of at most
$1.1$. On Perception Test, direct QA answered $376$ of $454$ numeric-label questions with a letter although the options were numbered; the numeric-label evaluator credits a letter at the correct position. Only 9 of the 18 control intervals in
Table~\ref{tab:paired-ci-controls} fall inside a $\pm2$-point equivalence margin fixed in
advance, so we read those columns as descriptive rather than as evidence of
invariance.

\begin{table}[!tbp]
  \centering
  \caption{\textbf{Paired $95\%$ confidence intervals on the option-channel controls.}
  Each entry is that row's multiple-choice score minus the same run with its options
  shuffled or its labels replaced by digits, read as in Table~\ref{tab:paired-ci}. A
  $\pm2$-point equivalence margin was fixed before the intervals were computed;
  9 of the 18 fall inside it, so we read these as descriptive rather than
  as evidence of invariance. ``--'' marks a condition that was never run.}
  \label{tab:paired-ci-controls}
  \scriptsize
  \setlength{\tabcolsep}{3pt}
  \begin{tabular}{llcc}
    \toprule
    & & \multicolumn{2}{c}{$\Delta$ MC} \\
    \cmidrule(lr){3-4}
    Benchmark & Method & shuffled options & numeric labels \\
    \midrule
    \multirow{2}{*}{NExT-QA}
      & Direct QA & $-0.4$ {\tiny $[-1.1,0.3]$} & $-0.3$ {\tiny $[-0.9,0.3]$} \\
      & Gaia loop & -- & $+0.2$ {\tiny $[-0.6,0.9]$} \\
    \midrule
    \multirow{2}{*}{EgoSchema}
      & Direct QA & $+0.4$ {\tiny $[-2.2,3.0]$} & $-0.6$ {\tiny $[-2.2,1.0]$} \\
      & Gaia loop & -- & $+0.8$ {\tiny $[-1.4,3.0]$} \\
    \midrule
    \multirow{2}{*}{LVBench}
      & Direct QA & $-0.1$ {\tiny $[-1.5,1.2]$} & $+0.5$ {\tiny $[-0.9,1.8]$} \\
      & Gaia loop & -- & $+0.3$ {\tiny $[-1.3,1.9]$} \\
    \midrule
    \multirow{2}{*}{QBench-Video}
      & Direct QA & $+0.3$ {\tiny $[-1.6,2.3]$} & $-0.1$ {\tiny $[-2.0,1.8]$} \\
      & Gaia loop & -- & $-0.8$ {\tiny $[-3.1,1.5]$} \\
    \midrule
    \multirow{2}{*}{Perception Test}
      & Direct QA & $-1.3$ {\tiny $[-4.2,1.4]$} & $-0.2$ {\tiny $[-3.1,2.6]$} \\
      & Gaia loop & -- & $-1.1$ {\tiny $[-4.0,1.8]$} \\
    \midrule
    \multirow{2}{*}{Video-MME\,v2}
      & Direct QA & $+0.1$ {\tiny $[-1.6,1.7]$} & $-0.5$ {\tiny $[-2.1,1.0]$} \\
      & Gaia loop & -- & $+0.1$ {\tiny $[-1.3,1.5]$} \\
    \bottomrule
  \end{tabular}
\end{table}

\subsection{Run-to-run variance of no-video runs}
\label{app:seed-variance}
Five independent seeds of no-video direct QA with Gemini 3.6 Flash were run on each of six VideoQA benchmarks. 
On all six benchmarks, the seed standard deviation is below half the width of the narrowest interval in Table~\ref{tab:paired-ci}.
Run-to-run variance therefore stays low even in the regime where models are largely guessing among the options.

\begin{table}[!tbp]
  \centering
  \caption{\textbf{Run-to-run variance of no-video runs.} The standard deviation is
  broadly smaller on the larger evaluation sets, from $0.16$ at $n=3{,}200$ to $1.20$ at
  $n=454$, though not monotonically: QBench-Video reaches $0.98$ over $868$ questions
  against EgoSchema's $0.89$ over $500$. Every value is small enough that run differences
  remain interpretable. Video-MME\,v2 follows the paper-wide convention of
  Table~\ref{tab:core}: flat per-question accuracy first, group score in
  brackets.}
  \label{tab:seed-variance}
  \footnotesize
  \setlength{\tabcolsep}{6pt}
  \begin{tabular}{lrrrr}
    \toprule
    Benchmark & $n$ & Mean & SD & Min--max \\
    \midrule
    NExT-QA (2021-05)         & 4996 & 40.30 & 0.38 & 39.97--41.01 \\
    EgoSchema (2023-08)       &  500 & 47.64 & 0.89 & 46.40--49.00 \\
    LVBench (2024-06)         & 1549 & 40.37 & 0.70 & 39.06--41.06 \\
    QBench-Video (2024-09)    &  868 & 49.38 & 0.98 & 47.58--50.58 \\
    Perception Test (2025-07) &  454 & 37.80 & 1.20 & 35.68--38.77 \\
    Video-MME\,v2 (2026-04)   & 3200 & 26.68 {\scriptsize (10.15)} & 0.16 {\scriptsize (0.10)} & 26.41--26.88 {\scriptsize (10.01--10.31)} \\
    \bottomrule
  \end{tabular}
\end{table}

\section{Training-Free, Model-Free Heuristic Baseline}
\label{app:heuristic}
The \emph{Heuristic} column of Table~\ref{tab:core} reports the best of the five scoring
functions in Algorithm~\ref{alg:heuristic}. Each is a deterministic function of the
question stem and the option texts, with no parameters, no training data and no encoder,
and none sees a frame of video. \textsc{Overlap} is the stem-matching heuristic of
\cite{mcvqa_bias2023}; \textsc{Odd} and \textsc{Cluster} probe distractor similarity
from either direction. Scoring a tie as expected accuracy keeps a heuristic from
borrowing accuracy from a position prior. Because the best of the five is chosen per
benchmark, the column is an oracle upper bound on what \emph{these} heuristics extract,
not a score a blind system could reach without knowing in advance which regularity to
exploit, and not a ceiling on blind accuracy in general: a model given the same stem and
options scores higher on every benchmark.

\begin{algorithm}[H]
\caption{Training-free, model-free heuristics (applied per question)}
\label{alg:heuristic}
\begin{algorithmic}[1]
\Function{Heuristic}{stem, options $O$, gold $a$, rule}
  \State $G(x) \gets$ set of word $n$-grams of $x$, $n\le3$
  \If{rule $=$ \textsc{Overlap}} \State $f(o) \gets |G(o) \cap G(\text{stem})|$
  \ElsIf{rule $=$ \textsc{Longest}} \State $f(o) \gets |o|$ \Comment{characters}
  \ElsIf{rule $=$ \textsc{Shortest}} \State $f(o) \gets -|o|$
  \Else \Comment{\textsc{Odd} and \textsc{Cluster}: mean similarity to siblings}
    \State $\bar{J}(o) \gets \operatorname{mean}_{o' \in O \setminus \{o\}} \operatorname{Jaccard}(G(o), G(o'))$
    \State $f(o) \gets -\bar{J}(o)$ for \textsc{Odd}, $+\bar{J}(o)$ for \textsc{Cluster}
  \EndIf
  \State $T \gets \arg\max_{o \in O} f(o)$ \Comment{all maximisers, ties kept}
  \State \Return $\mathbf{1}[a \in T] \, / \, |T|$ \Comment{expected accuracy under a uniform tie-break}
\EndFunction
\end{algorithmic}
\end{algorithm}

\section{Evaluation Procedures and Prompts}
\label{app:procedures}

This appendix details the evaluation procedures behind Section~\ref{sec:results} alongside the exact matcher, extractor, and judge prompts. The pseudocode is transcribed directly from our evaluation harness, and prompts are reproduced verbatim up to minor typographic normalisation.

\subsection{Direct QA}
\label{app:proc-directqa}

Direct QA is a single call: frames are sampled from the video at 1\,fps, the
question (with its options embedded, when multiple choice) is placed in the user
turn, and its system prompt fixes the answer format. The no-video condition is
the same call with the frame set emptied, which is the only difference between
the \texttt{baseline} and \texttt{no\_video} configurations.

\begin{algorithm}[H]
\caption{Direct QA for one question}
\label{alg:directqa}
\begin{algorithmic}[1]
\Require video $v$, question $q$ (options embedded), gold label $a$, condition flags
\State $(q, a) \gets \Call{Ablate}{q, a}$ \Comment{Alg.~\ref{alg:ablations}; identity in the baseline}
\If{\texttt{language\_only}}
  \State $F \gets \emptyset$ \Comment{no video: the language-prior ceiling}
\Else
  \State $F \gets \Call{SampleFrames}{v, \text{fps}=1}$
\EndIf
\State $s \gets$ system prompt for the condition
\State $\hat{y} \gets \Call{VLM}{s, F, q}$
\State \Return $(\hat{y}, a)$ \Comment{scored by Alg.~\ref{alg:mcscore} (MC) or Alg.~\ref{alg:oematch} (OE)}
\end{algorithmic}
\end{algorithm}

\subsection{Agentic loop}
\label{app:proc-loop}

The agentic system replaces the single call with a tool-use loop over the same
video. The agent alternates \textsc{Think}, \textsc{Act} and \textsc{Observe}
steps, calling one tool per step, and must deliver its answer through the
\texttt{finish} tool rather than as free text. The full step sequence is
recorded, which is what makes replay (Alg.~\ref{alg:replay}) possible.

\begin{algorithm}[H]
\caption{Agentic loop for one question}
\label{alg:loop}
\begin{algorithmic}[1]
\Require video $v$, question $q$, tool set $\mathcal{T}$, step budget $N_{\max}$
\State $T \gets [\,]$ \Comment{trajectory: the recorded step sequence}
\State $c \gets (\,s_{\text{loop}},\ u(q, v)\,)$ \Comment{system + user prompt; $q$ and $v$ arrive already ablated}
\For{$n = 1$ \textbf{to} $N_{\max}$}
  \State $\tau \gets \Call{Agent}{c, T}$ \Comment{one tool call: \texttt{global\_browse}, \texttt{clip\_search}, \texttt{frame\_inspect}, \texttt{finish}}
  \If{$\tau$ is \texttt{finish}}
    \State \Return $(\text{answer}(\tau), T)$
  \EndIf
  \State $o \gets \Call{Execute}{\tau, v}$
  \State append $(\tau, o)$ to $T$
\EndFor
\State \Return $(\emptyset, T)$ \Comment{budget exhausted}
\end{algorithmic}
\end{algorithm}

\subsection{Reproducing the author systems}
\label{app:author-repro}

Two of our six benchmarks overlap with the original DVD and AVP papers, allowing direct verification of our re-runs against published results.
DVD reports $74.2$ on the full LVBench release and $76.6$ on the EgoSchema public subset, both on an OpenAI o3
orchestrator~\cite{dvd2025}. AVP reports $74.8$ on LVBench with Gemini-2.5-Pro and $63.8$ with
Gemini-2.5-Flash~\cite{avp2026}. 
We run Author-DVD with OpenAI o3 and Author-AVP with Gemini-2.5-Pro, thus keeping the models behind their published LVBench results. On the identical splits Author-DVD scores $73.7$ on LVBench and $73.8$ on
EgoSchema and Author-AVP $76.3$ on LVBench (Table~\ref{tab:general-mcoe}).  DVD therefore lands $0.5$ points below its
published LVBench value and $2.8$ below its EgoSchema one, and AVP $1.5$ above its stronger
published LVBench value, which is close enough that we read both implementations as running
as their authors intended.

\subsection{Replay and option-delayed replay}
\label{app:proc-replay}

Replay resumes a recorded open-ended trajectory from a saved intermediate state
and changes the information available from that point onward. The three replay variants differ only in the cut index $k$: the options are
injected after $k$ preserved steps, and the agent generates a new continuation
from there. $k = 0$ is MC from the start, $k = \lfloor n/2 \rfloor$ is midpoint
replay, and $k = n$ reveals the options only at the \texttt{finish} call.

\begin{algorithm}[H]
\caption{Option-delayed replay}
\label{alg:replay}
\begin{algorithmic}[1]
\Require OE trajectory $T = [t_1, \ldots, t_n]$, options $O$, cut index $k$
\State $T' \gets [t_1, \ldots, t_k]$ \Comment{preserved prefix; tools are not re-executed}
\State $q' \gets q$ with $O$ appended \Comment{options enter the context at step $k+1$}
\If{$k = n$}
  \State \Return $\Call{Finish}{c, T', q'}$ \Comment{one call: select an option from the completed trajectory}
\EndIf
\State \Return $\Call{Resume}{\text{Alg.}~\ref{alg:loop},\ T',\ q'}$ \Comment{new continuation from the cut}
\end{algorithmic}
\end{algorithm}

\subsection{Option-channel ablations}
\label{app:proc-ablations}

Each ablation is a pure function of the $(\text{question}, \text{answer})$ pair,
applied by the dataloader before inference, so no condition differs from the baseline
in any other respect. Because these transformations are applied in the dataloader prior to inference, both Direct QA (Alg.~\ref{alg:directqa}) and the agentic loop (Alg.~\ref{alg:loop}) execute unmodified across all ablations. Every ablated row in Tables~\ref{tab:core} and~\ref{tab:general-mcoe} comes
from the unmodified procedure. Multiple-choice datasets embed the options in the question
string, so each transform parses the options out, mutates them, and rewrites the
gold label to stay consistent. Option style ($\texttt{A. text}$ versus
$\texttt{(A) text}$) is preserved. Shuffling is seeded per question, so it
is reproducible.

\begin{algorithm}[H]
\caption{Option-channel ablations (applied per question)}
\label{alg:ablations}
\begin{algorithmic}[1]
\Function{Ablate}{$q$, $a$}
  \If{\texttt{hide\_question}} \Comment{runs alone: it discards the stem and options}
    \State \Return (``Based only on the video, guess the correct answer.\ Respond with a single letter (A, B, \ldots).'', $a$)
  \EndIf
  \State $(\text{stem}, O) \gets \Call{ParseOptions}{q}$; $L \gets$ labels of $O$; $i \gets$ index of $a$ in $L$
  \If{$|O| < 2$ \textbf{or} $i$ undefined} \State \Return $(q, a)$ \Comment{no-op on open-ended items} \EndIf
  \If{\texttt{shuffle\_options}}
    \State permute the option \emph{texts} with a per-question seeded RNG; labels stay in place
    \State $a \gets$ label at the correct option's new position
  \EndIf
  \If{\texttt{numeric\_labels}}
    \State relabel $\text{A}, \text{B}, \text{C}, \ldots \to 1, 2, 3, \ldots$; $a \gets$ the matching number
  \EndIf
  \State \Return $(\Call{Rebuild}{\text{stem}, O}, a)$
\EndFunction
\end{algorithmic}
\end{algorithm}

\subsection{Scoring}
\label{app:proc-scoring}

Multiple-choice runs are scored by exact match against the gold label. A
prediction that is already a clean label is compared directly. Only predictions
that are not are passed to an LLM extractor, which is asked to return one label
and nothing else. If the extractor returns anything outside the valid label set,
the raw prediction is kept and scores as wrong, so extraction failures can never
manufacture a correct answer. Labels are upper-cased, and numeric labels are mapped back to letters first.

\begin{algorithm}[H]
\caption{Multiple-choice scoring}
\label{alg:mcscore}
\begin{algorithmic}[1]
\Require predictions $\hat{Y}$, gold labels $A$, valid label set $V$
\For{each $\hat{y} \in \hat{Y}$}
  \State $p \gets \textsc{Upper}(\textsc{Strip}(\hat{y}))$
  \If{$p \notin V$}
    \State $p' \gets \Call{Extract}{\hat{y}}$ \Comment{LLM extractor, Box~2 (Appendix~\ref{app:prompts})}
    \State $p \gets p'$ \textbf{if} $p' \in V$ \textbf{else} $\hat{y}$ \Comment{a failed extraction scores wrong}
  \EndIf
  \State correct $\mathrel{+}= \mathbf{1}[p = a]$
\EndFor
\State \Return accuracy overall and per question category
\end{algorithmic}
\end{algorithm}

The OE$\to$match column re-scores an open-ended run on the multiple-choice metric by
mapping each free-text answer onto an option, so it and the MC column are on one scale.
The two OE columns are not: they are judge scores (\S\ref{app:judge}), of which only the bracketed share of items rated at least $80$ is a per-item accuracy. 
Because the model never sees the candidate options during OE generation, all responses are mapped to options using the LLM matcher (Box~1, Appendix~\ref{app:prompts}) rather than surface string matching. 
The judge is given the question as well
as the options, because datasets with fine-grained distractors build them to
differ only in the facet the question asks about, and without the question the
matcher picks whichever option the answer most resembles overall. Labels are
positional and sized to the option count, so benchmarks with more than five
options are handled. An answer that matches nothing resolves to \textsc{None}
and scores wrong. The resolved labels are then handed to
Alg.~\ref{alg:mcscore}'s scoring step, which is what preserves each benchmark's
own category breakdown.

\begin{algorithm}[H]
\caption{Open-ended to multiple-choice matching}
\label{alg:oematch}
\begin{algorithmic}[1]
\Require free-text prediction $\hat{y}$, options $O$, question $q$
\State $L \gets [\text{A}, \text{B}, \ldots]$ truncated to $|O|$ \Comment{positional labels}
\State $s \gets$ matcher prompt over $q$ and $O$ \Comment{Box~1 (Appendix~\ref{app:prompts})}
\State $\ell \gets \Call{Judge}{s, \hat{y}}$
\State \Return $\ell$ \textbf{if} $\ell \in L$ \textbf{else} \textsc{None}
\end{algorithmic}
\end{algorithm}

\subsection{The open-ended judge}
\label{app:judge}

A free-text answer has no exact match to score against, so both open-ended columns are
means of a judge's rating~\cite{autoevalvideo2024, mllmjudge2024}. The judge sees the question, the system's answer and a
reference answer, and returns one integer from $0$ to $100$; the reported score is the
mean of those integers over the benchmark's items. We run two judges,
\texttt{claude-sonnet-4-5} and \texttt{gpt-5.4-mini}, and report both rather than averaging
them. The first is sampled at temperature $0$ with extended thinking disabled and a
ten-token cap, and re-scoring NExT-QA with it reproduces the reported mean exactly. The second, a reasoning model run at its default temperature, is not: re-scoring moves a benchmark's mean by up to $0.7$ points.

Because a continuous $0$--$100$ rating is not directly comparable to binary option accuracy, we binarise the judge ratings at $\ge 80$ (Box~3's ``mostly correct'' anchor, (Appendix~\ref{app:prompts})) whenever computing MC$-$OE differences. 
For every MC$-$OE comparison the judge is binarised per item at $\ge 80$, Box~3's ``mostly
correct'' anchor, which puts both sides of the difference on per-item correctness; the mean
rating is kept in Table~\ref{tab:general-mcoe} as a descriptive figure beside it. The
ratings do not collapse to a binary on their own, so the threshold is a real choice and not
a formality: a median $79.8\%$ of the items on a variant receive a rating that is neither $0$
nor $100$, and on one variant $99.4\%$ do. 
We selected the $\ge 80$ threshold a priori based on the rubric's ``mostly correct'' anchor, and Table~\ref{tab:judge-threshold} sweeps alternative thresholds across all 48 judge-scored cells. The MC$-$OE gap increases monotonically with the threshold, and MC outperforms OE across every condition for thresholds $\ge 60$ (whereas $\ge 50$ and the raw mean rating produce a single inversion by crediting partially correct free text at parity with an exact option match).

\begin{table}[!tbp]
  \centering
  \caption{\textbf{Sensitivity of open-ended accuracy to the judge binarisation threshold.} Each row scores an open-ended
  answer correct when the judge rated it at or above the cut, over the $48$ judge-scored cells of
  Table~\ref{tab:general-mcoe}. \emph{OE} is the mean binarised score, \emph{MC$-$OE} its
  mean difference from the same row's multiple-choice accuracy, \emph{inverted} counts cells
  where open-ended scoring beats multiple choice, and \emph{recovery} is the share of the
  MC$-$OE gap that last-step replay closes on the Gaia loop. The paper reports
  $\ge 80$.}
  \label{tab:judge-threshold}
  \small
  \begin{tabular}{lccccc}
    \toprule
    Cut & OE & MC$-$OE & MC$-$OE range & inverted & recovery (\%) \\
    \midrule
    mean rating & 60.6 & $+14.1$ & $-2.7$ to $+32.3$ & 1 & 42--100 \\
    $\ge 50$ & 61.8 & $+13.0$ & $-1.3$ to $+31.5$ & 1 & 33--100 \\
    $\ge 60$ & 59.6 & $+15.2$ & $+0.0$ to $+38.2$ & 0 & 36--100 \\
    $\ge 70$ & 54.0 & $+20.7$ & $+4.0$ to $+49.8$ & 0 & 63--100 \\
    $\ge 80$ & 45.7 & $\mathbf{+29.1}$ & $+9.7$ to $+62.8$ & 0 & \textbf{79--100} \\
    $\ge 90$ & 32.5 & $+42.2$ & $+14.5$ to $+78.8$ & 0 & 87--100 \\
    \bottomrule
  \end{tabular}
\end{table}

Transient provider failures are retried rather than scored as zero, and an empty answer
scores zero without being sent.

None of the six multiple-choice benchmarks ships a free-text answer key, so the
reference is the correct option's own text with its label stripped. A system that
identifies the right event but describes it in different words than the option author
chose can therefore lose credit that multiple-choice scoring would have given it, which
inflates the MC$-$OE gap by an amount we do not separate out. Two features of the rubric
in Box~3 limit that inflation (Appendix~\ref{app:prompts}): near-synonyms and differing phrasings are to be treated as equivalent, and verbosity earns nothing.

\subsection{Prompts}
\label{app:prompts}

\begin{promptbox}{Box 1: open-ended to option matcher}
You are a strict answer matcher.

Goal: Given a free-text answer and a list of options, return the single option label whose meaning best matches the answer.

Question the answer responds to: \{question\}

Options:

\quad \{options\}

Rules:

\quad - Return ONLY a single label, written exactly as one of: \{labels\}.

\quad - Choose the one option whose meaning most closely matches the answer, even if the wording differs.

\quad - If no option is a reasonable match, return exactly: none

\quad - Do not output anything else (no punctuation, no explanations).
\end{promptbox}

\begin{promptbox}{Box 2: multiple-choice answer extractor}
You are a strict answer extractor.

Goal: Extract exactly one valid option from the given text.

Valid options: \{options\}.

Rules:

\quad - Return ONLY a single option, written exactly as one of: \{options\}.

\quad - Prefer an explicitly stated final answer (e.g., "final answer: X", "answer is (X)", "I choose X", where X is one of the valid options).

\quad - If multiple options appear, choose the one most clearly marked as the final choice.

\quad - If no valid option is present, return exactly: none

\quad - Do not output anything else (no punctuation, no explanations).
\end{promptbox}

\begin{promptbox}{Box 3: open-ended judge}
You are an impartial expert judge evaluating answers to video-based questions. Your role is to assign a score (0--100) comparing the predicted answer against the reference. Evaluate holistically using these criteria:

\quad - Factual accuracy (is the prediction true and grounded in the video?)

\quad - Relevance (does it answer the actual question directly?)

\quad - Completeness (does it cover all key parts of the reference?)

\quad - Faithfulness (does it avoid hallucinations or contradictions?)

Important fairness rules:

\quad - Do NOT reward verbosity. A short but correct answer must score the same as a long correct answer.

\quad - If both the prediction and reference indicate uncertainty, give full credit even if phrasing differs.

\quad - For numeric/count answers, allow a tolerance of $\pm1$, scoring 60--80 rather than 0 within it.

\quad - For roles, relationships and occupations, treat near-synonyms as equivalent.

\quad - If the prediction is correct but provides harmless extra context, do not penalize.

\quad - If the prediction invents details, penalize it even if other parts are correct.

\quad - Ignore differences in casing, punctuation, or formatting when judging equivalence.

\quad - If the reference specifies a final outcome and the prediction instead describes the steps leading to that same outcome, consider them equivalent.

Scoring guidelines: 100 exactly correct or semantically equivalent; 90 very accurate with tiny omissions; 80 mostly correct; 70 partially correct but noticeably incomplete; 60 major omissions or errors; 50 half correct; 40 a few accurate fragments; 30 mostly wrong; 20 a trace of overlap; 10 fragmentary alignment; 0 completely incorrect, irrelevant, or fabricated.

Respond ONLY with a single integer score from 0 to 100.

\medskip
User message: \ \ Question: \{question\} \ \textbar\ \ Predicted answer: \{prediction\} \ \textbar\ \ Reference Answer(s): \{reference\}
\end{promptbox}

\input{appendix_a}

\end{document}

%% file: preamble.tex
\usepackage{graphicx}
\usepackage{xcolor}
\usepackage{pifont}
\usepackage{tabularx}
\usepackage{textcomp}
\usepackage{xltabular}
\graphicspath{{./}{figures/}}

\definecolor{gaGoldColor}{HTML}{0B6B3A}
\definecolor{gaBadColor}{HTML}{A5231F}
\definecolor{gaMutedColor}{HTML}{6B6B6B}
\newcommand{\gaGold}[1]{\textcolor{gaGoldColor}{#1}}
\newcommand{\gaBad}[1]{\textcolor{gaBadColor}{#1}}
\newcommand{\gaMuted}[1]{\textcolor{gaMutedColor}{#1}}

%% file: appendix_a.tex
\section{Illustrating Grounding and Its Removal}
\label{sec:appendix-grounding}

\subsection{Language-prior strength}
\label{sec:appendix-a1}

\begin{figure}[H]
  \centering
  \setlength{\tabcolsep}{0.5pt}
  \begin{tabular}{cccccc}
    \includegraphics[width=\dimexpr(\linewidth-6pt)/6\relax]{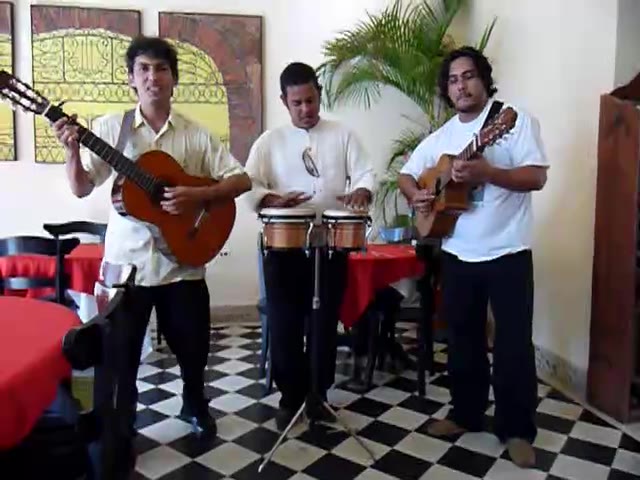} &
    \includegraphics[width=\dimexpr(\linewidth-6pt)/6\relax]{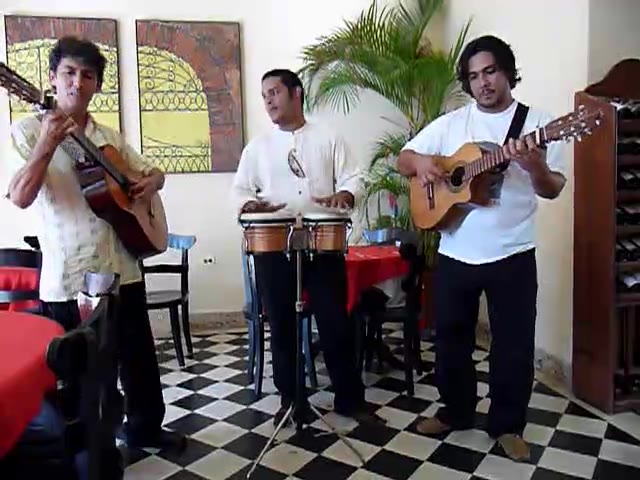} &
    \includegraphics[width=\dimexpr(\linewidth-6pt)/6\relax]{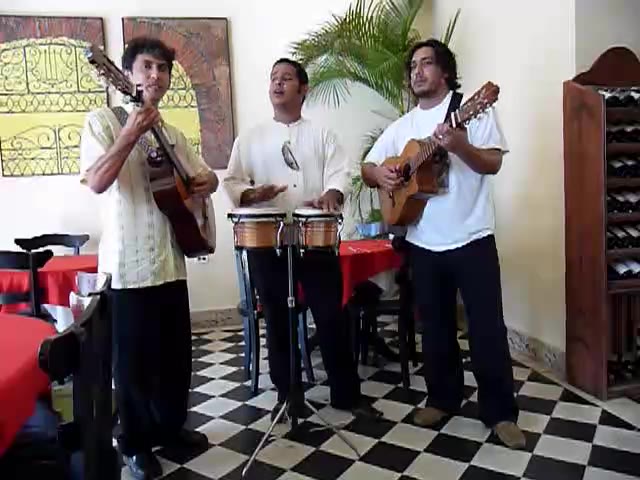} &
    \includegraphics[width=\dimexpr(\linewidth-6pt)/6\relax]{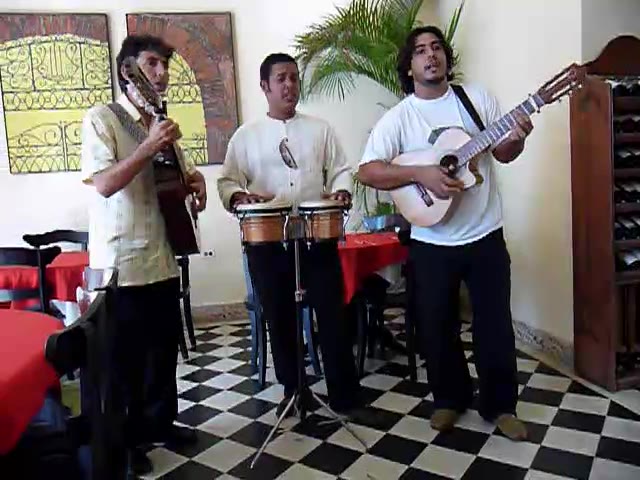} &
    \includegraphics[width=\dimexpr(\linewidth-6pt)/6\relax]{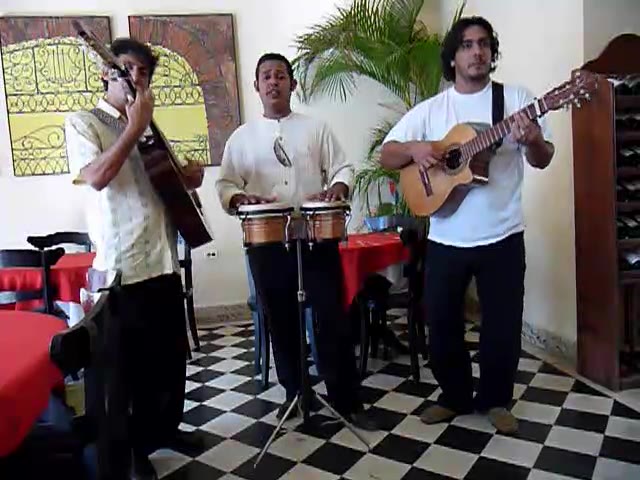} &
    \includegraphics[width=\dimexpr(\linewidth-6pt)/6\relax]{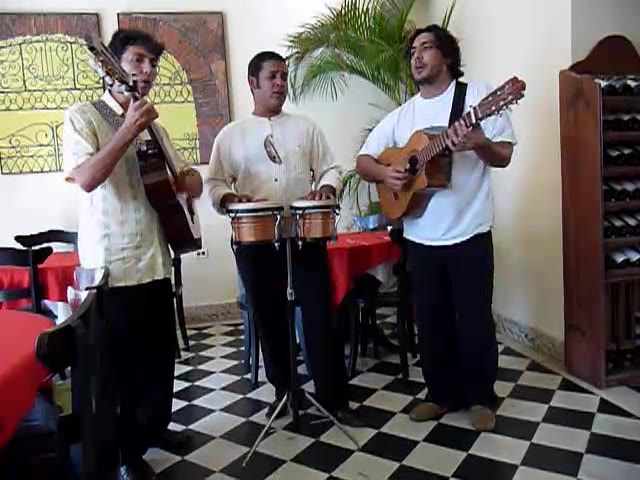} \\[1pt]
    \scriptsize\ttfamily 00:04 &
    \scriptsize\ttfamily 00:12 &
    \scriptsize\ttfamily 00:21 &
    \scriptsize\ttfamily 00:29 &
    \scriptsize\ttfamily 00:37 &
    \scriptsize\ttfamily 00:46 \\
  \end{tabular}

  \vspace{4pt}
  \begin{minipage}{\linewidth}
    \small
    \textbf{Q.} why is the man standing on the right moving his left hand up and down the guitar

    \vspace{2pt}
    \begin{tabularx}{\linewidth}{@{}l X@{}}
      (A) & play the keys \\
      (B) & enjoy the music \\
      \gaGold{(C)} & \gaGold{strum guitar~~$\leftarrow$ gold} \\
      (D) & singing \\
      (E) & part of performance \\
    \end{tabularx}

    \vspace{3pt}
    {\footnotesize \gaGold{\ding{51}}~video + question\quad \gaGold{\ding{51}}~question only\quad \gaGold{\ding{51}}~question only (DVD)\quad \gaBad{\ding{55}}~video only}
  \end{minipage}

  \caption{
    \textbf{NExT-QA}, \texttt{3270490566.mp4},
    question 5 (frames sampled evenly across the whole video).
    The stem already names the instrument and the hand motion. Only \emph{strum guitar} is compatible with both; the rest are ruled out before a frame is seen.
  }
  \label{fig:a1-1}
\end{figure}

\begin{figure}[H]
  \centering
  \setlength{\tabcolsep}{0.5pt}
  \begin{tabular}{cccccc}
    \includegraphics[width=\dimexpr(\linewidth-6pt)/6\relax]{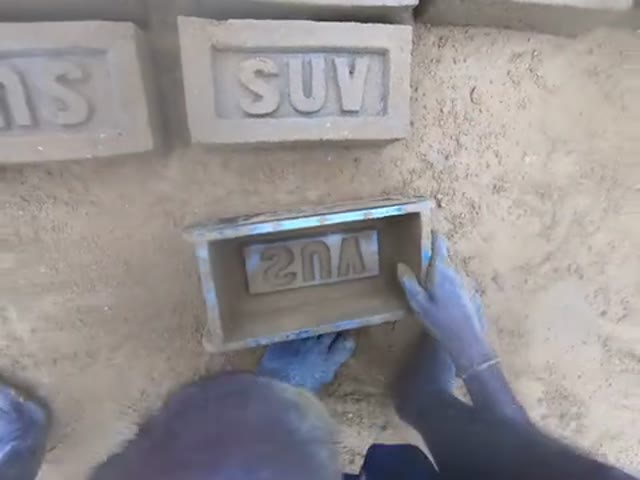} &
    \includegraphics[width=\dimexpr(\linewidth-6pt)/6\relax]{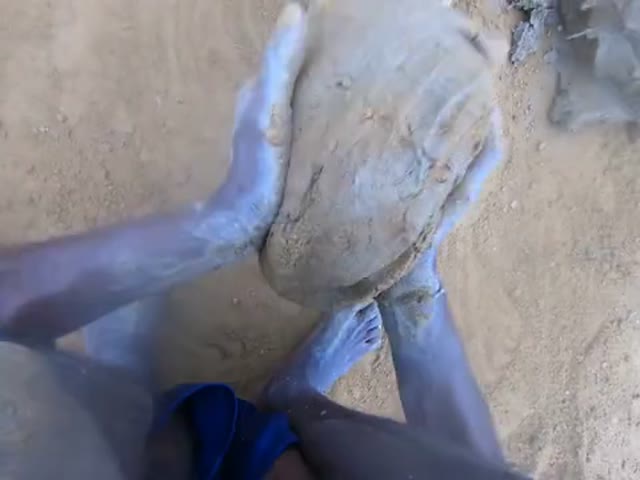} &
    \includegraphics[width=\dimexpr(\linewidth-6pt)/6\relax]{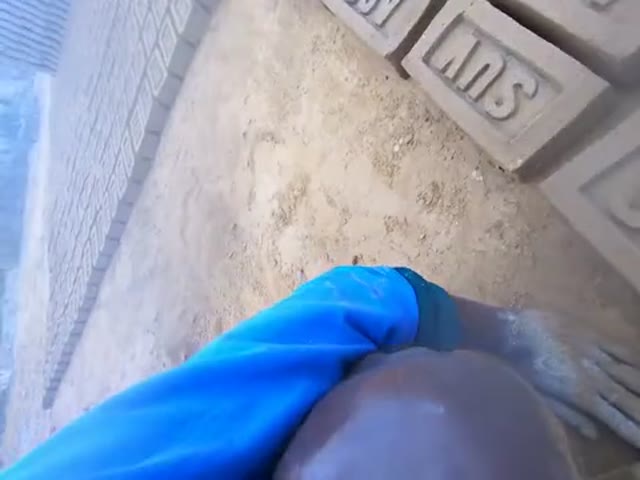} &
    \includegraphics[width=\dimexpr(\linewidth-6pt)/6\relax]{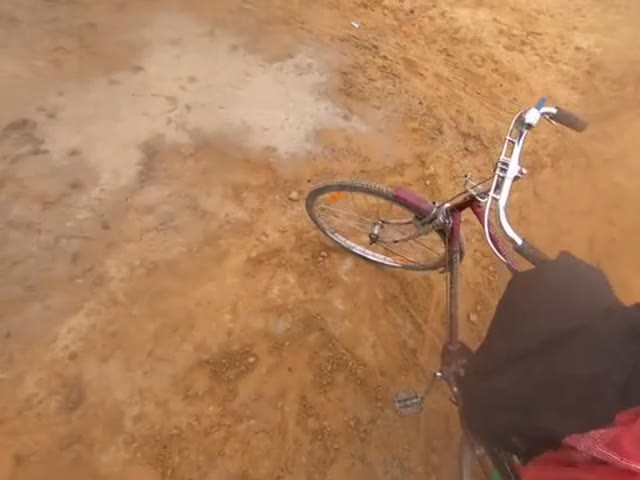} &
    \includegraphics[width=\dimexpr(\linewidth-6pt)/6\relax]{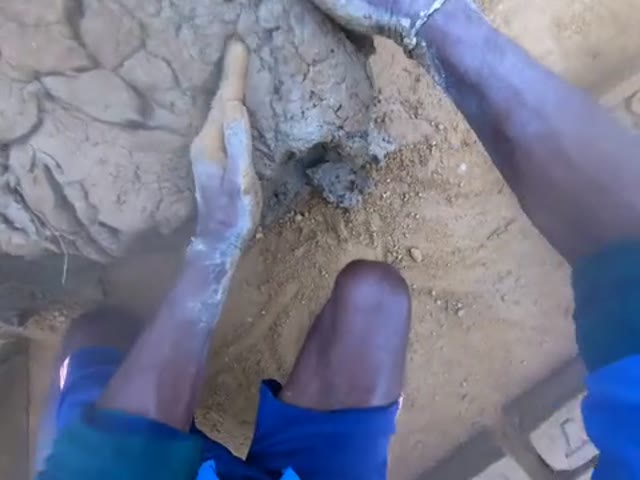} &
    \includegraphics[width=\dimexpr(\linewidth-6pt)/6\relax]{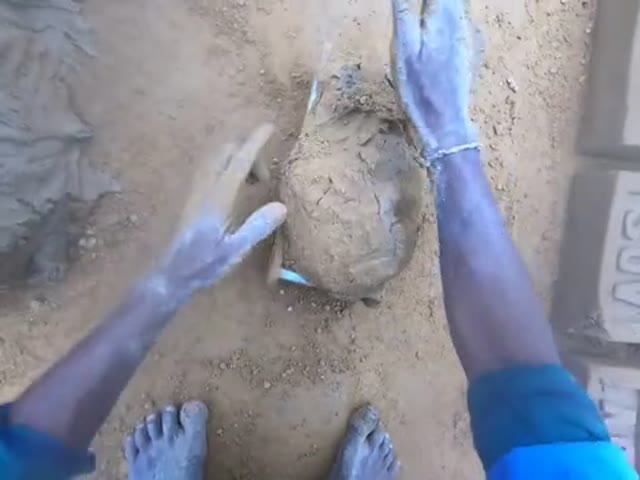} \\[1pt]
    \scriptsize\ttfamily 00:15 &
    \scriptsize\ttfamily 00:45 &
    \scriptsize\ttfamily 01:15 &
    \scriptsize\ttfamily 01:45 &
    \scriptsize\ttfamily 02:15 &
    \scriptsize\ttfamily 02:45 \\
  \end{tabular}

  \vspace{4pt}
  \begin{minipage}{\linewidth}
    \small
    \textbf{Q.} What are the key techniques and steps c utilized to work with the clay and clay mould throughout the video? please summarize them succinctly.

    \vspace{2pt}
    \begin{tabularx}{\linewidth}{@{}l X@{}}
      (A) & C uses a hoe to shape the clay into bricks. \\
      \gaGold{(B)} & \gaGold{C uses a clay mould to shape the clay into bricks.~~$\leftarrow$ gold} \\
      (C) & Carefully, c employs his skillful hands to progressively mold the clay into sturdy bricks. \\
      (D) & Carefully, c employs a wheel for shaping the clay effectively into well-formed bricks. \\
      (E) & C skillfully uses a high-temperature kiln to carefully shape the malleable clay into strong, durable bricks. \\
    \end{tabularx}

    \vspace{3pt}
    {\footnotesize \gaGold{\ding{51}}~video + question\quad \gaGold{\ding{51}}~question only\quad \gaGold{\ding{51}}~question only (DVD)\quad \gaBad{\ding{55}}~video only}
  \end{minipage}

  \caption{
    \textbf{EgoSchema}, \texttt{111b7189-cc45-438c-8144-ded1eed0f6c2.mp4},
    question 0 (frames sampled evenly across the whole video).
    The stem names the clay mould, and exactly one option mentions a clay mould. The question leaks its own answer.
  }
  \label{fig:a1-2}
\end{figure}

\begin{figure}[H]
  \centering
  \setlength{\tabcolsep}{0.5pt}
  \begin{tabular}{cccccc}
    \includegraphics[width=\dimexpr(\linewidth-6pt)/6\relax]{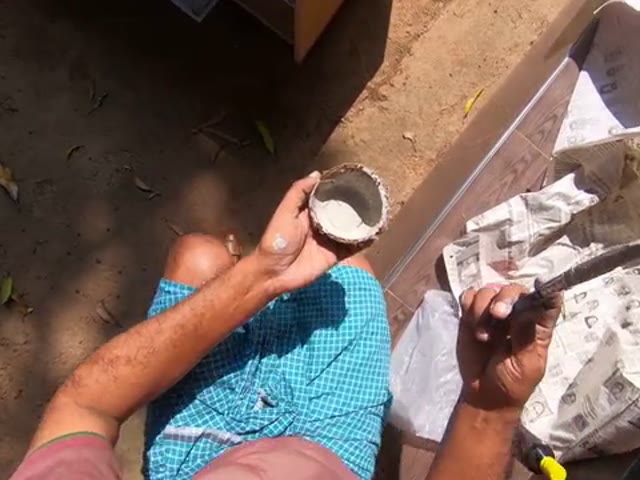} &
    \includegraphics[width=\dimexpr(\linewidth-6pt)/6\relax]{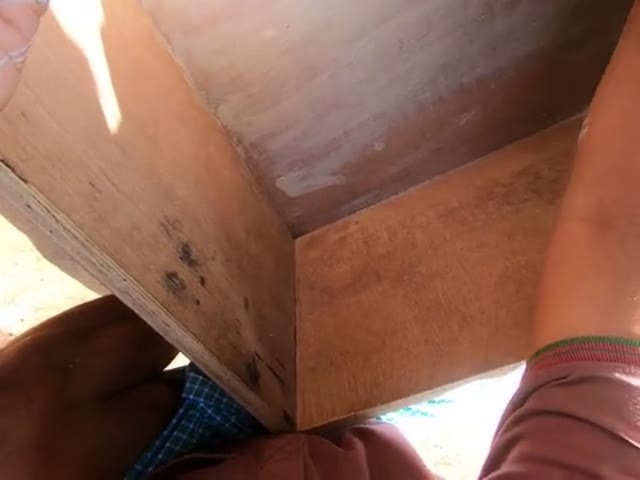} &
    \includegraphics[width=\dimexpr(\linewidth-6pt)/6\relax]{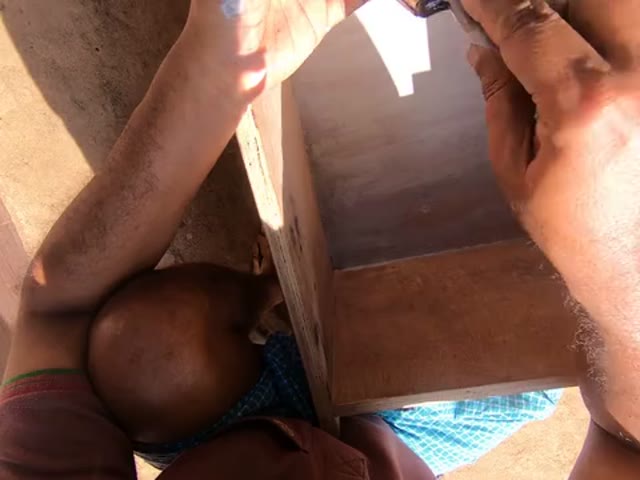} &
    \includegraphics[width=\dimexpr(\linewidth-6pt)/6\relax]{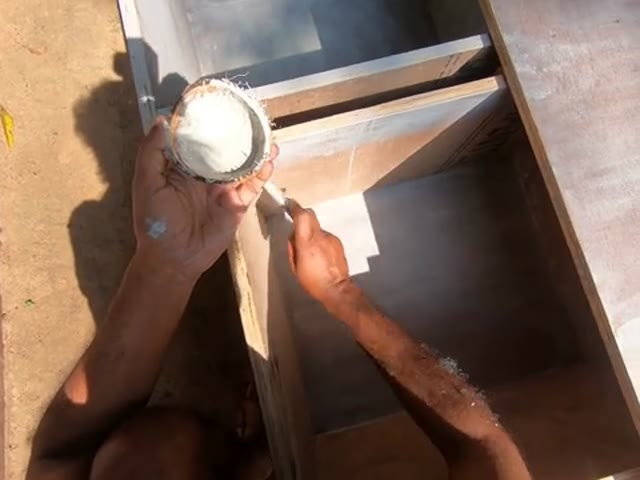} &
    \includegraphics[width=\dimexpr(\linewidth-6pt)/6\relax]{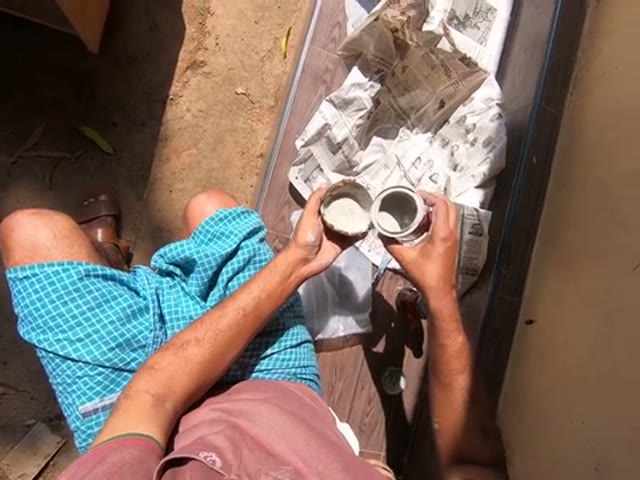} &
    \includegraphics[width=\dimexpr(\linewidth-6pt)/6\relax]{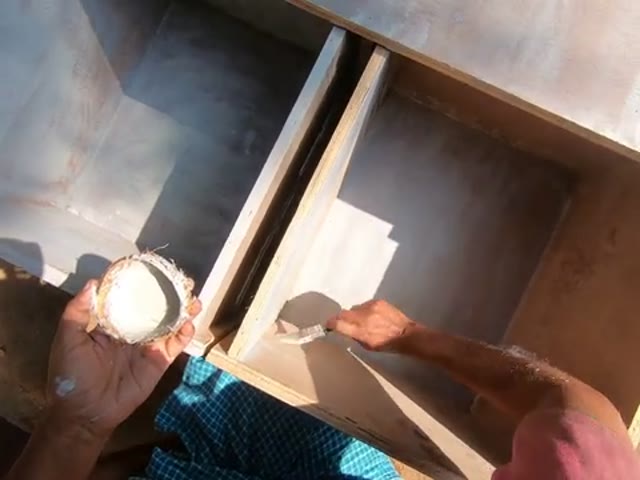} \\[1pt]
    \scriptsize\ttfamily 00:15 &
    \scriptsize\ttfamily 00:45 &
    \scriptsize\ttfamily 01:15 &
    \scriptsize\ttfamily 01:45 &
    \scriptsize\ttfamily 02:15 &
    \scriptsize\ttfamily 02:45 \\
  \end{tabular}

  \vspace{4pt}
  \begin{minipage}{\linewidth}
    \small
    \textbf{Q.} Summarize the overall process demonstrated in the video and identify two materials used by c to stir the paint in the coconut shell.

    \vspace{2pt}
    \begin{tabularx}{\linewidth}{@{}l X@{}}
      \gaGold{(A)} & \gaGold{C stirs the paint in the coconut shell with a paint brush and a piece of wood.~~$\leftarrow$ gold} \\
      (B) & Casually, c stirs the paint thoroughly inside the coconut shell, utilizing a repurposed bottle cover. \\
      (C) & Carefully, c stirs the paint inside the coconut shell using his hands, mixing it well. \\
      (D) & Enthusiastically, c stirs the paint carefully inside the coconut shell with a thin wooden stick. \\
      (E) & C stirs the paint in the coconut shell with a spoon. \\
    \end{tabularx}

    \vspace{3pt}
    {\footnotesize \gaGold{\ding{51}}~video + question\quad \gaGold{\ding{51}}~question only\quad \gaGold{\ding{51}}~question only (DVD)\quad \gaBad{\ding{55}}~video only}
  \end{minipage}

  \caption{
    \textbf{EgoSchema}, \texttt{59dcde1d-f210-48fc-b888-5574d8962a2c.mp4},
    question 0 (frames sampled evenly across the whole video).
    The stem asks for two materials. Only one option names two; the other four name one. The answer is fixed by arity, before any content is considered.
  }
  \label{fig:a1-3}
\end{figure}

\subsection{World knowledge}
\label{sec:appendix-a2}

\begin{figure}[H]
  \centering
  \setlength{\tabcolsep}{0.5pt}
  \begin{tabular}{cccccc}
    \includegraphics[width=\dimexpr(\linewidth-6pt)/6\relax]{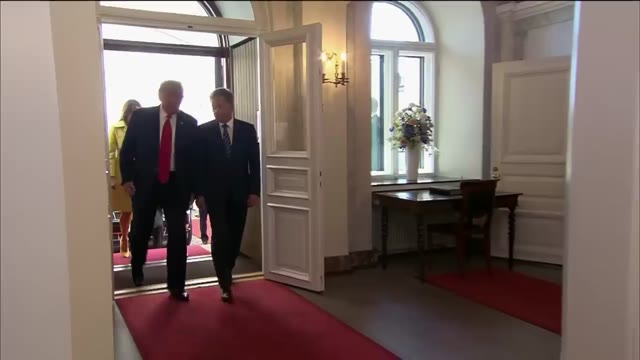} &
    \includegraphics[width=\dimexpr(\linewidth-6pt)/6\relax]{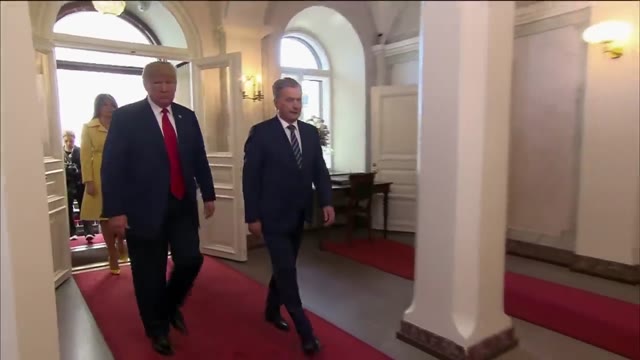} &
    \includegraphics[width=\dimexpr(\linewidth-6pt)/6\relax]{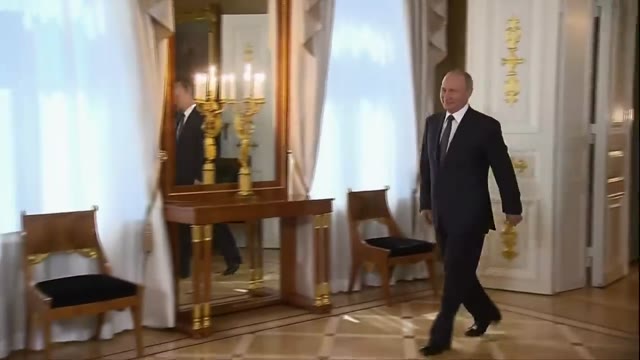} &
    \includegraphics[width=\dimexpr(\linewidth-6pt)/6\relax]{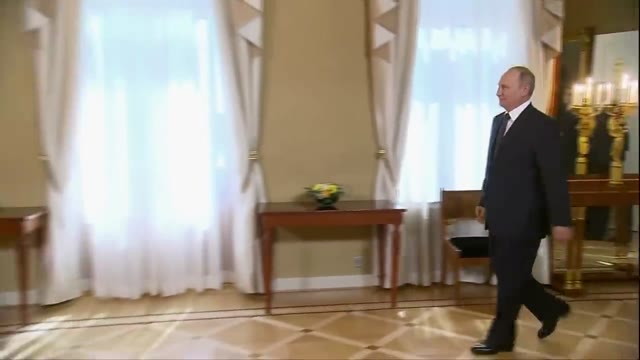} &
    \includegraphics[width=\dimexpr(\linewidth-6pt)/6\relax]{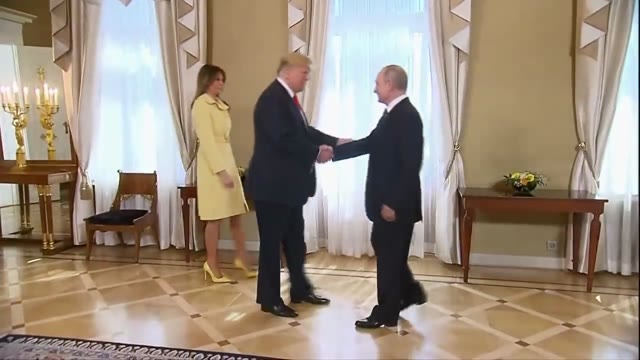} &
    \includegraphics[width=\dimexpr(\linewidth-6pt)/6\relax]{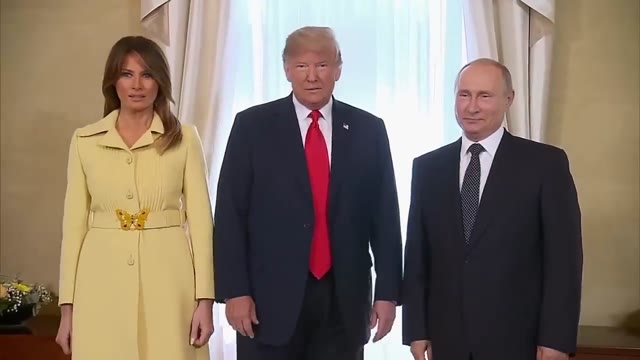} \\[1pt]
    \scriptsize\ttfamily 35:36 &
    \scriptsize\ttfamily 35:38 &
    \scriptsize\ttfamily 35:41 &
    \scriptsize\ttfamily 35:43 &
    \scriptsize\ttfamily 35:46 &
    \scriptsize\ttfamily 35:48 \\
  \end{tabular}

  \vspace{4pt}
  \begin{minipage}{\linewidth}
    \small
    \textbf{Q.} What color of coat does Melania Trump wear for the meeting that is hosted at Helsinki, Finland?

    \vspace{2pt}
    \begin{tabularx}{\linewidth}{@{}l X@{}}
      (A) & Green \\
      (B) & Dark blue \\
      (C) & Bright red \\
      \gaGold{(D)} & \gaGold{Yellow~~$\leftarrow$ gold} \\
    \end{tabularx}

    \vspace{3pt}
    {\footnotesize \gaGold{\ding{51}}~video + question\quad \gaGold{\ding{51}}~question only\quad \gaGold{\ding{51}}~question only (DVD)\quad \gaBad{\ding{55}}~video only}
  \end{minipage}

  \caption{
    \textbf{LVBench}, \texttt{aJI8XTa\_DII.mp4},
    question 10 (frames sampled inside the annotated \texttt{time\_reference} window).
    The 2018 Helsinki summit is a heavily reported public event. The coat colour is recoverable from pretraining text, so the video is redundant.
  }
  \label{fig:a2-1}
\end{figure}

\begin{figure}[H]
  \centering
  \setlength{\tabcolsep}{0.5pt}
  \begin{tabular}{cccccc}
    \includegraphics[width=\dimexpr(\linewidth-6pt)/6\relax]{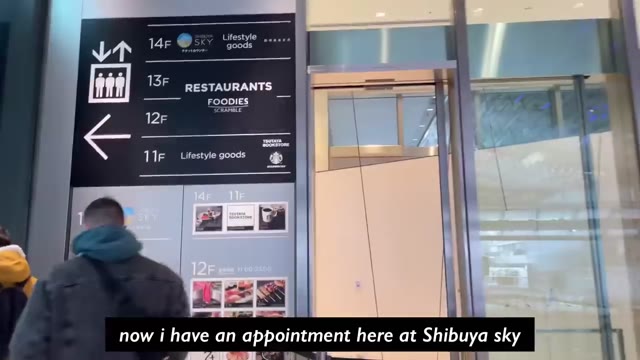} &
    \includegraphics[width=\dimexpr(\linewidth-6pt)/6\relax]{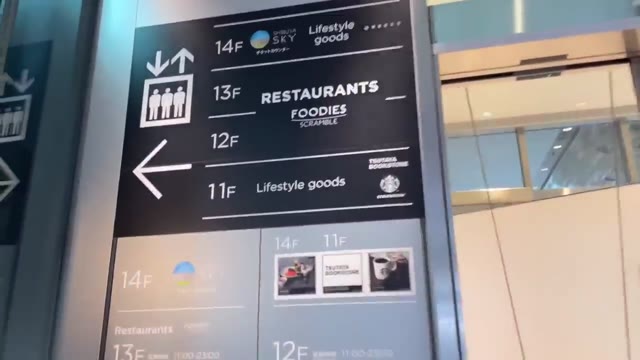} &
    \includegraphics[width=\dimexpr(\linewidth-6pt)/6\relax]{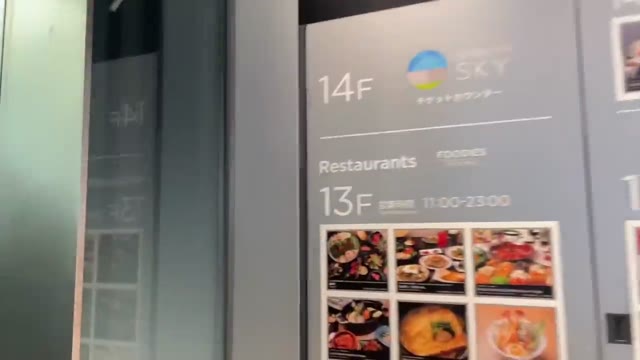} &
    \includegraphics[width=\dimexpr(\linewidth-6pt)/6\relax]{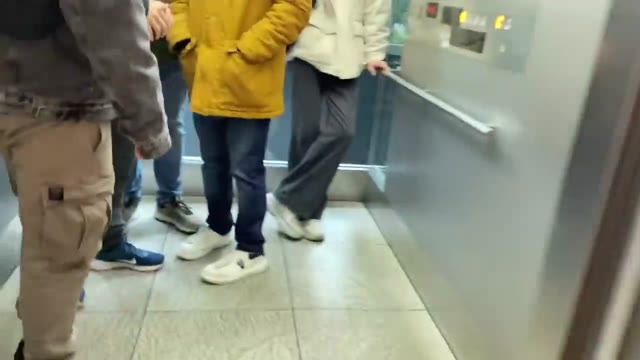} &
    \includegraphics[width=\dimexpr(\linewidth-6pt)/6\relax]{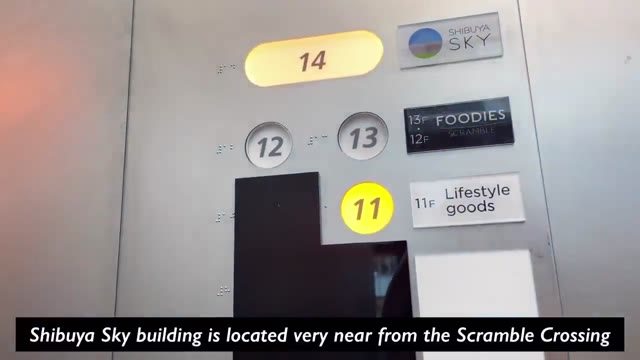} &
    \includegraphics[width=\dimexpr(\linewidth-6pt)/6\relax]{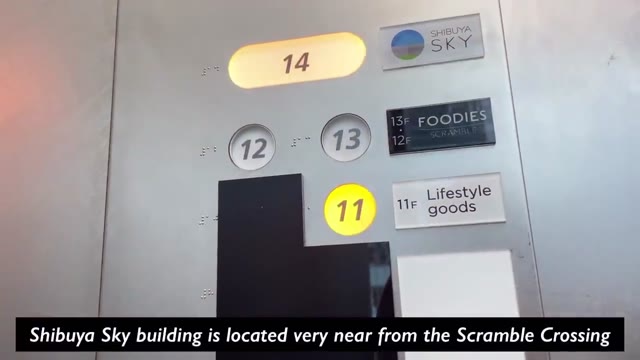} \\[1pt]
    \scriptsize\ttfamily 10:45 &
    \scriptsize\ttfamily 10:47 &
    \scriptsize\ttfamily 10:49 &
    \scriptsize\ttfamily 10:50 &
    \scriptsize\ttfamily 10:52 &
    \scriptsize\ttfamily 10:54 \\
  \end{tabular}

  \vspace{4pt}
  \begin{minipage}{\linewidth}
    \small
    \textbf{Q.} On which floor is Shibuya Sky?

    \vspace{2pt}
    \begin{tabularx}{\linewidth}{@{}l X@{}}
      \gaGold{(A)} & \gaGold{14~~$\leftarrow$ gold} \\
      (B) & 11 \\
      (C) & 12 \\
      (D) & 13 \\
    \end{tabularx}

    \vspace{3pt}
    {\footnotesize \gaGold{\ding{51}}~video + question\quad \gaGold{\ding{51}}~question only\quad \gaGold{\ding{51}}~question only (DVD)\quad \gaBad{\ding{55}}~video only}
  \end{minipage}

  \caption{
    \textbf{LVBench}, \texttt{RCAqKnvu\_P0.mp4},
    question 8 (frames sampled inside the annotated \texttt{time\_reference} window).
    A guidebook fact about a named landmark. The question is trivia wearing a video-QA costume.
  }
  \label{fig:a2-2}
\end{figure}

\begin{figure}[H]
  \centering
  \setlength{\tabcolsep}{0.5pt}
  \begin{tabular}{cccccc}
    \includegraphics[width=\dimexpr(\linewidth-6pt)/6\relax]{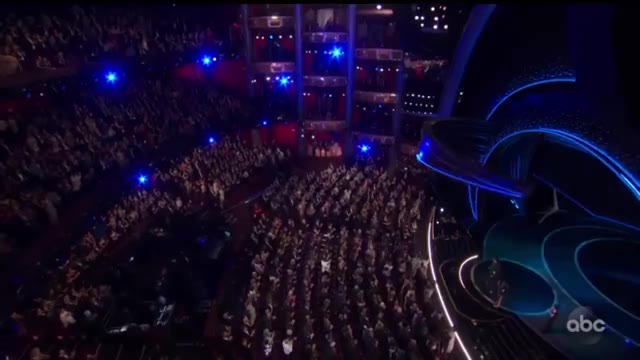} &
    \includegraphics[width=\dimexpr(\linewidth-6pt)/6\relax]{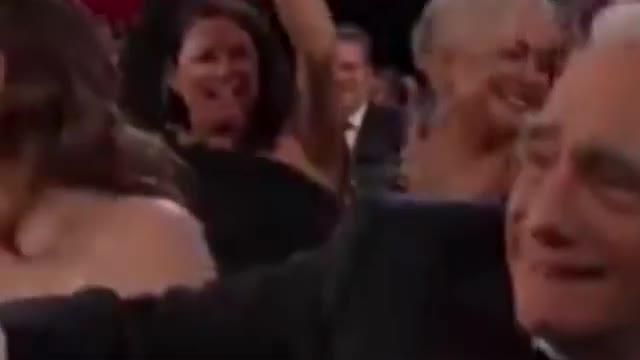} &
    \includegraphics[width=\dimexpr(\linewidth-6pt)/6\relax]{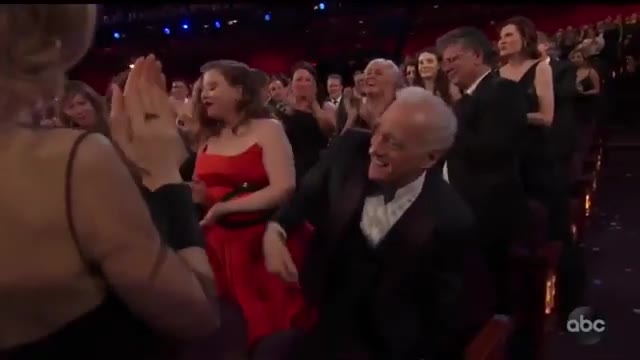} &
    \includegraphics[width=\dimexpr(\linewidth-6pt)/6\relax]{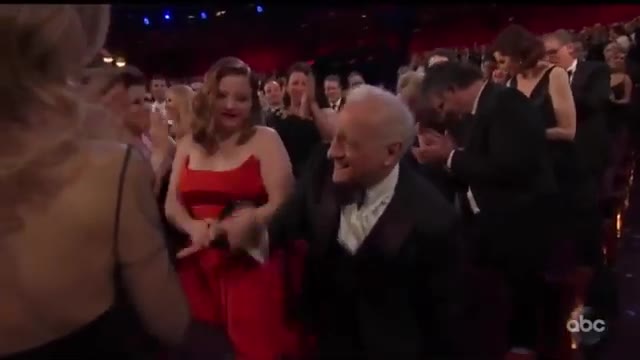} &
    \includegraphics[width=\dimexpr(\linewidth-6pt)/6\relax]{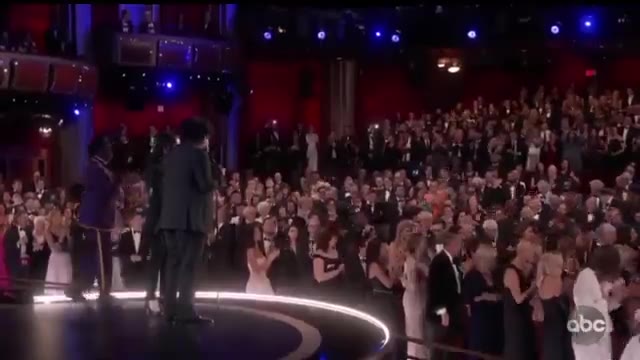} &
    \includegraphics[width=\dimexpr(\linewidth-6pt)/6\relax]{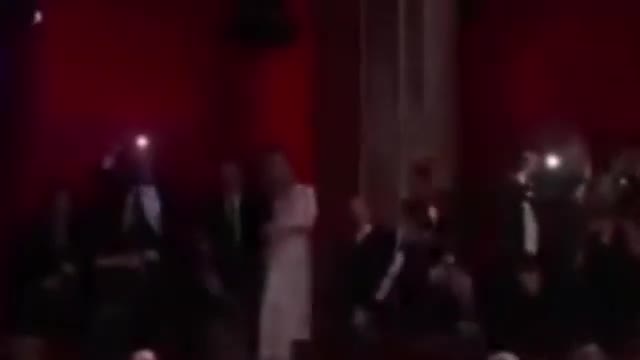} \\[1pt]
    \scriptsize\ttfamily 70:44 &
    \scriptsize\ttfamily 70:46 &
    \scriptsize\ttfamily 70:47 &
    \scriptsize\ttfamily 70:48 &
    \scriptsize\ttfamily 70:50 &
    \scriptsize\ttfamily 70:51 \\
  \end{tabular}

  \vspace{4pt}
  \begin{minipage}{\linewidth}
    \small
    \textbf{Q.} What is the relationship between Bong Joon Ho and the elderly man whom people are standing up and applauding for during his speech?

    \vspace{2pt}
    \begin{tabularx}{\linewidth}{@{}l X@{}}
      (A) & Actor and producer \\
      (B) & Employer and employee \\
      (C) & Father and son \\
      \gaGold{(D)} & \gaGold{Senior producer and young producer~~$\leftarrow$ gold} \\
    \end{tabularx}

    \vspace{3pt}
    {\footnotesize \gaGold{\ding{51}}~video + question\quad \gaGold{\ding{51}}~question only\quad \gaGold{\ding{51}}~question only (DVD)\quad \gaBad{\ding{55}}~video only}
  \end{minipage}

  \caption{
    \textbf{LVBench}, \texttt{-WnyRMZqV1U.mp4},
    question 7 (frames sampled inside the annotated \texttt{time\_reference} window).
    Names a public figure and an award ceremony; the relationship is documented outside the video.
  }
  \label{fig:a2-3}
\end{figure}

\subsection{Distractor and construction defects}
\label{sec:appendix-a3}

\begin{figure}[H]
  \centering
  \setlength{\tabcolsep}{0.5pt}
  \begin{tabular}{cccccc}
    \includegraphics[width=\dimexpr(\linewidth-6pt)/6\relax]{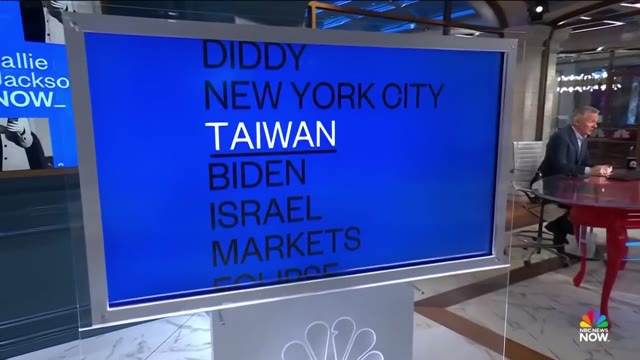} &
    \includegraphics[width=\dimexpr(\linewidth-6pt)/6\relax]{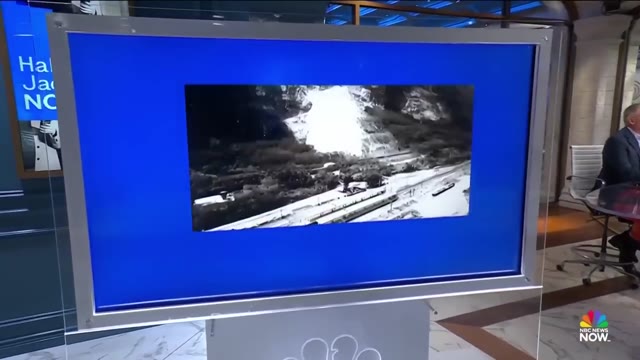} &
    \includegraphics[width=\dimexpr(\linewidth-6pt)/6\relax]{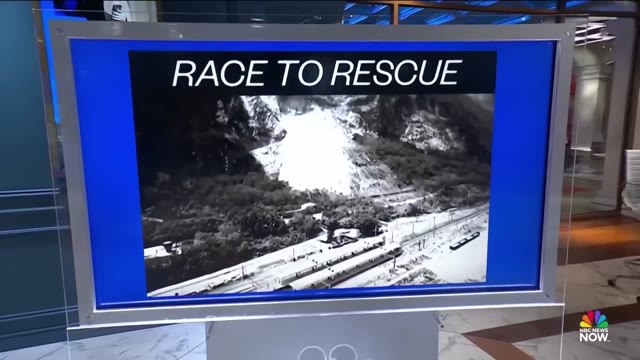} &
    \includegraphics[width=\dimexpr(\linewidth-6pt)/6\relax]{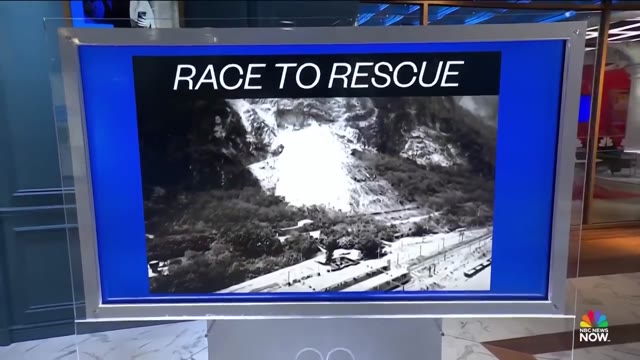} &
    \includegraphics[width=\dimexpr(\linewidth-6pt)/6\relax]{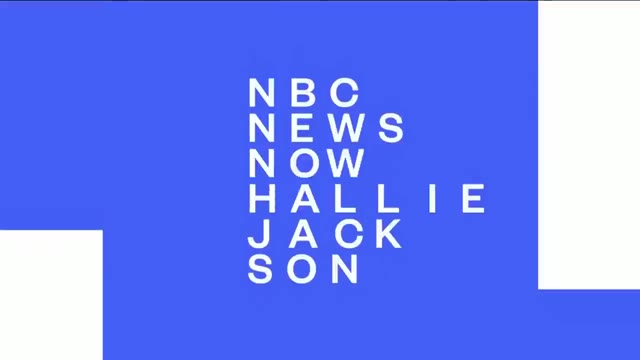} &
    \includegraphics[width=\dimexpr(\linewidth-6pt)/6\relax]{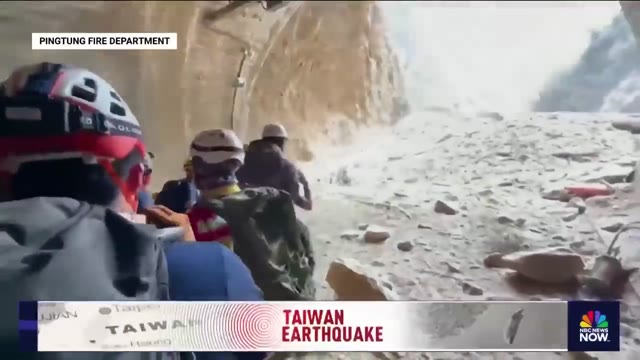} \\[1pt]
    \scriptsize\ttfamily 08:35 &
    \scriptsize\ttfamily 08:37 &
    \scriptsize\ttfamily 08:38 &
    \scriptsize\ttfamily 08:39 &
    \scriptsize\ttfamily 08:41 &
    \scriptsize\ttfamily 08:42 \\
  \end{tabular}

  \vspace{4pt}
  \begin{minipage}{\linewidth}
    \small
    \textbf{Q.} What words appear on the screen after the blue scene with white word 'TAIWAN' shifted to picture?

    \vspace{2pt}
    \begin{tabularx}{\linewidth}{@{}l X@{}}
      \gaBad{(A)} & \gaBad{RACE TO RESCUE~~$\leftarrow$ identical to gold, scored wrong} \\
      (B) & TAIWAN TO RESCUE \\
      (C) & RACE RESOUE \\
      \gaGold{(D)} & \gaGold{RACE TO RESCUE~~$\leftarrow$ gold} \\
    \end{tabularx}

    \vspace{3pt}
    {\footnotesize \gaMuted{--~video + question}\quad \gaGold{\ding{51}}~question only\quad \gaGold{\ding{51}}~question only (DVD)\quad \gaGold{\ding{51}}~video only}
  \end{minipage}

  \caption{
    \textbf{LVBench}, \texttt{moALQl818ZY.mp4},
    question 3 (frames sampled inside the annotated \texttt{time\_reference} window).
    Options (A) and (D) are the identical string `RACE TO RESCUE', and only (D) is gold. A model that answers (A) gives a textually identical answer and is scored wrong.
  }
  \label{fig:a3-1}
\end{figure}

\begin{figure}[H]
  \centering
  \setlength{\tabcolsep}{0.5pt}
  \begin{tabular}{cccccc}
    \includegraphics[width=\dimexpr(\linewidth-6pt)/6\relax]{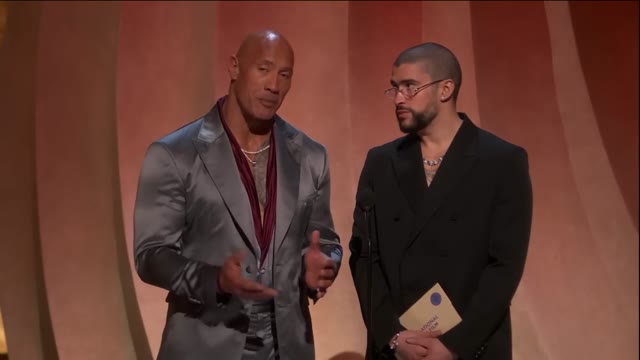} &
    \includegraphics[width=\dimexpr(\linewidth-6pt)/6\relax]{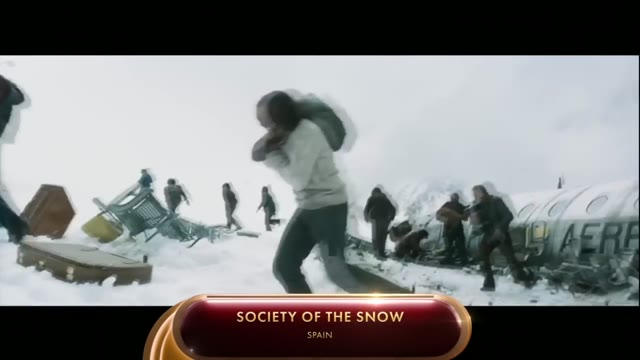} &
    \includegraphics[width=\dimexpr(\linewidth-6pt)/6\relax]{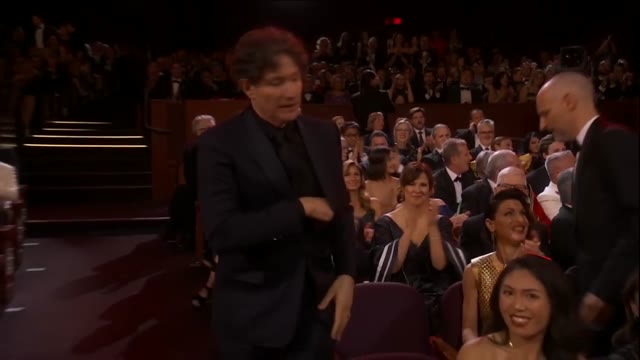} &
    \includegraphics[width=\dimexpr(\linewidth-6pt)/6\relax]{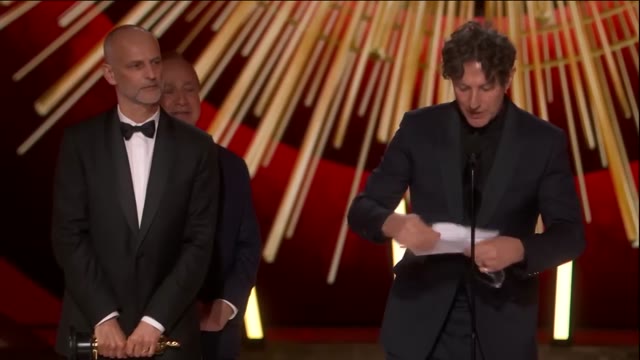} &
    \includegraphics[width=\dimexpr(\linewidth-6pt)/6\relax]{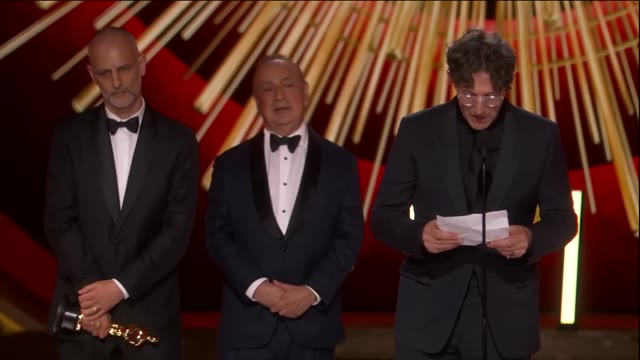} &
    \includegraphics[width=\dimexpr(\linewidth-6pt)/6\relax]{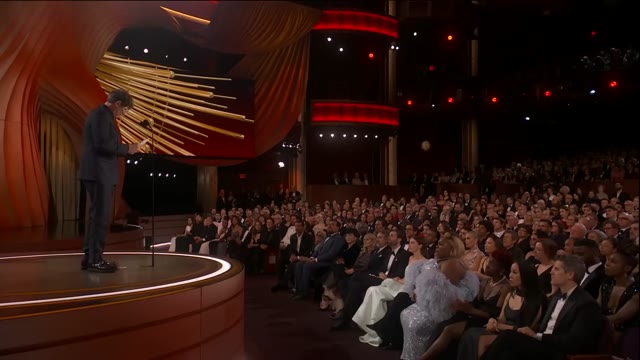} \\[1pt]
    \scriptsize\ttfamily 35:30 &
    \scriptsize\ttfamily 36:01 &
    \scriptsize\ttfamily 36:32 &
    \scriptsize\ttfamily 37:02 &
    \scriptsize\ttfamily 37:33 &
    \scriptsize\ttfamily 38:04 \\
  \end{tabular}

  \vspace{4pt}
  \begin{minipage}{\linewidth}
    \small
    \textbf{Q.} There is a performance of respectful tribal musicians playing a huge drum together. All of the drummers \_\_\_.

    \vspace{2pt}
    \begin{tabularx}{\linewidth}{@{}l X@{}}
      \gaGold{(A)} & \gaGold{Play synchronously most of the time~~$\leftarrow$ gold} \\
      \gaBad{(B)} & \gaBad{Play synchronously most of the time~~$\leftarrow$ identical to gold, scored wrong} \\
      (C) & Play asynchronously all the time \\
      (D) & Play synchronously all the time \\
    \end{tabularx}

    \vspace{3pt}
    {\footnotesize \gaBad{\ding{55}}~video + question\quad \gaGold{\ding{51}}~question only\quad \gaGold{\ding{51}}~question only (DVD)\quad \gaGold{\ding{51}}~video only}
  \end{minipage}

  \caption{
    \textbf{LVBench}, \texttt{t-RtDI2RWQs.mp4},
    question 16 (frames sampled inside the annotated \texttt{time\_reference} window).
    Options (A) and (B) are identical and are the gold answer. The item has two correct labels but accepts one.
  }
  \label{fig:a3-2}
\end{figure}

\begin{figure}[H]
  \centering
  \setlength{\tabcolsep}{0.5pt}
  \begin{tabular}{cccccc}
    \includegraphics[width=\dimexpr(\linewidth-6pt)/6\relax]{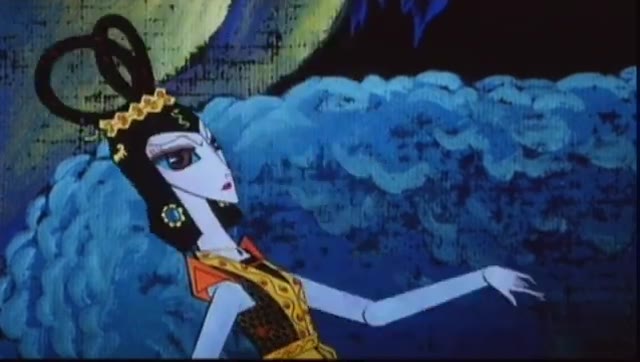} &
    \includegraphics[width=\dimexpr(\linewidth-6pt)/6\relax]{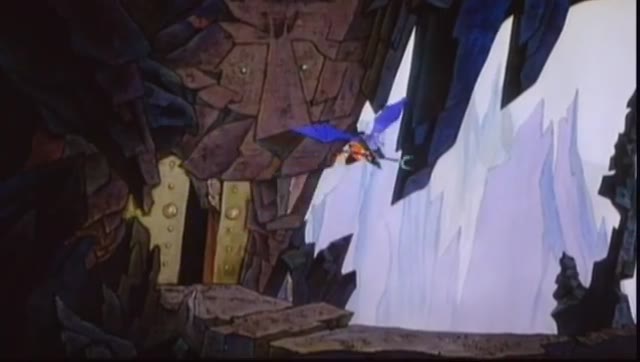} &
    \includegraphics[width=\dimexpr(\linewidth-6pt)/6\relax]{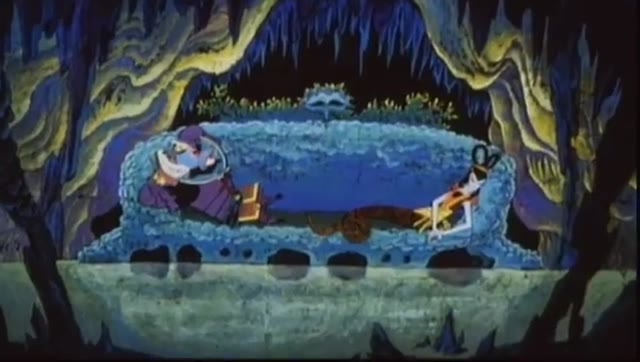} &
    \includegraphics[width=\dimexpr(\linewidth-6pt)/6\relax]{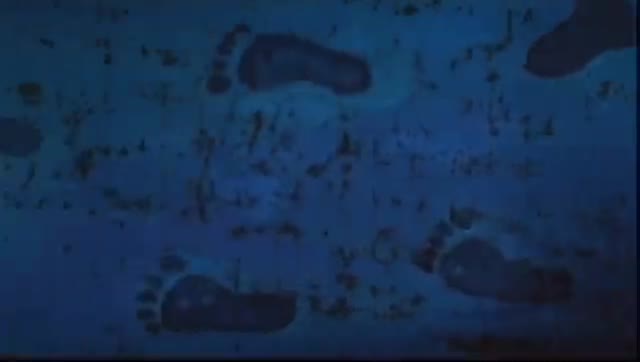} &
    \includegraphics[width=\dimexpr(\linewidth-6pt)/6\relax]{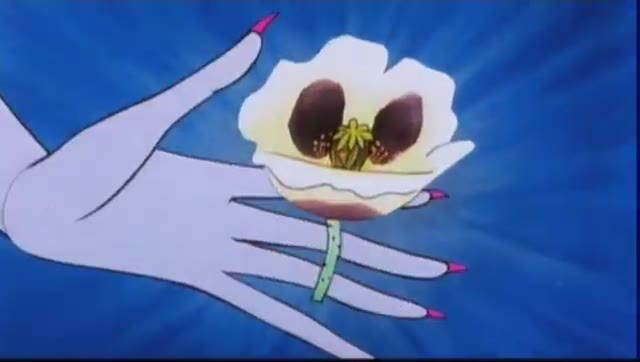} &
    \includegraphics[width=\dimexpr(\linewidth-6pt)/6\relax]{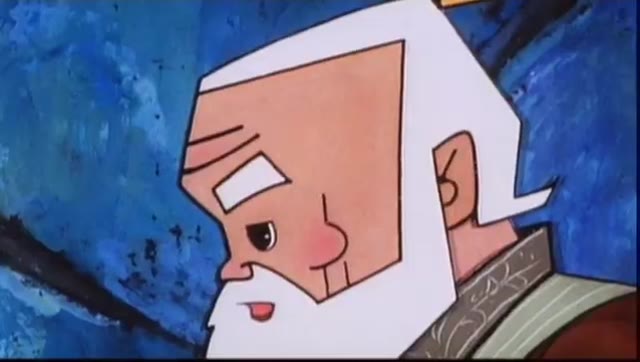} \\[1pt]
    \scriptsize\ttfamily 21:16 &
    \scriptsize\ttfamily 22:23 &
    \scriptsize\ttfamily 23:30 &
    \scriptsize\ttfamily 24:37 &
    \scriptsize\ttfamily 25:44 &
    \scriptsize\ttfamily 26:51 \\
  \end{tabular}

  \vspace{4pt}
  \begin{minipage}{\linewidth}
    \small
    \textbf{Q.} What are the skills of the orange calabash boy?

    \vspace{2pt}
    \begin{tabularx}{\linewidth}{@{}l X@{}}
      (A) & He is indestructible \\
      \gaBad{(B)} & \gaBad{He has infinitely power and can get bigger~~$\leftarrow$ identical to gold, scored wrong} \\
      (C) & He has invisibility \\
      \gaGold{(D)} & \gaGold{He has infinitely power and can get bigger~~$\leftarrow$ gold} \\
    \end{tabularx}

    \vspace{3pt}
    {\footnotesize \gaGold{\ding{51}}~video + question\quad \gaGold{\ding{51}}~question only\quad \gaGold{\ding{51}}~question only (DVD)\quad \gaBad{\ding{55}}~video only}
  \end{minipage}

  \caption{
    \textbf{LVBench}, \texttt{XNtNNplAwiI.mp4},
    question 6 (frames sampled inside the annotated \texttt{time\_reference} window).
    Options (B) and (D) are identical and (D) is gold, so answering (B) is scored wrong for an identical string. The item is really a three-way choice.
  }
  \label{fig:a3-3}
\end{figure}

\subsection{Properly grounded items}
\label{sec:appendix-a4}

\begin{figure}[H]
  \centering
  \setlength{\tabcolsep}{0.5pt}
  \begin{tabular}{cccccc}
    \includegraphics[width=\dimexpr(\linewidth-6pt)/6\relax]{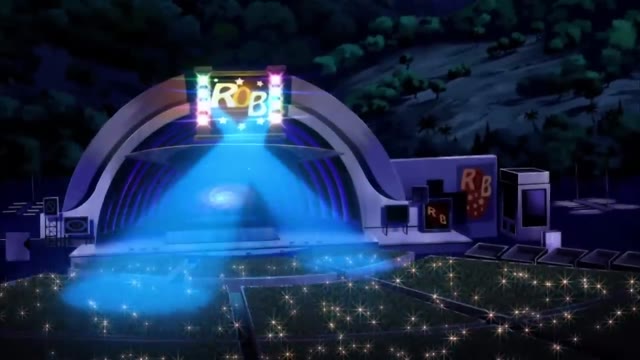} &
    \includegraphics[width=\dimexpr(\linewidth-6pt)/6\relax]{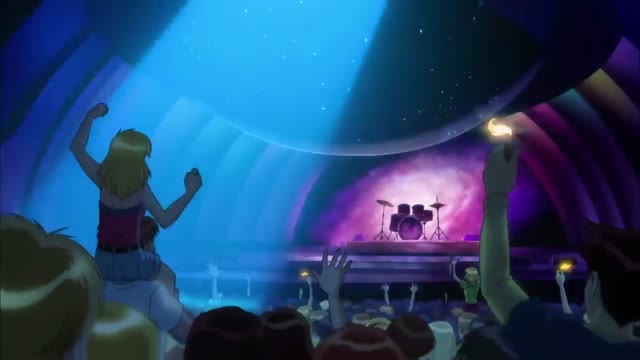} &
    \includegraphics[width=\dimexpr(\linewidth-6pt)/6\relax]{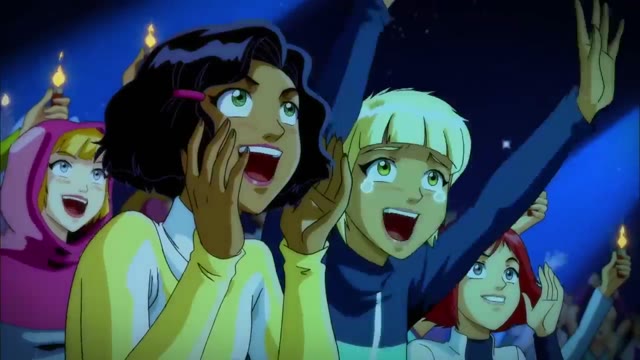} &
    \includegraphics[width=\dimexpr(\linewidth-6pt)/6\relax]{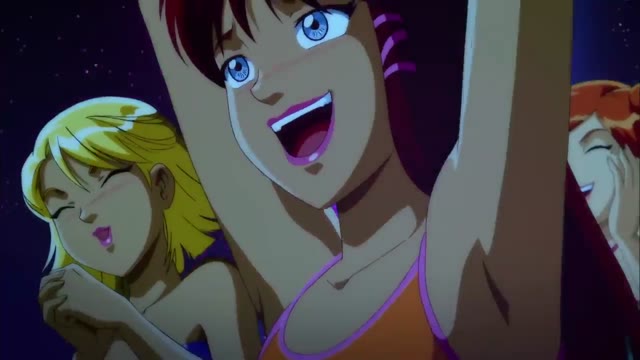} &
    \includegraphics[width=\dimexpr(\linewidth-6pt)/6\relax]{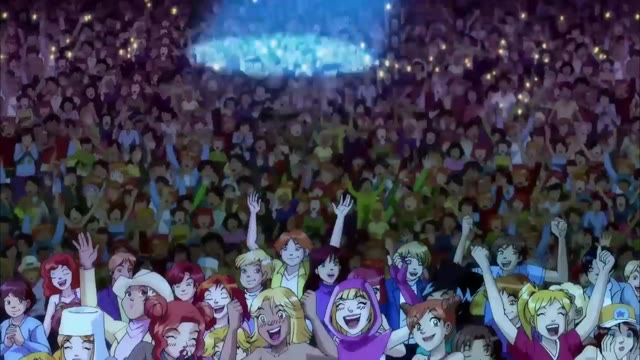} &
    \includegraphics[width=\dimexpr(\linewidth-6pt)/6\relax]{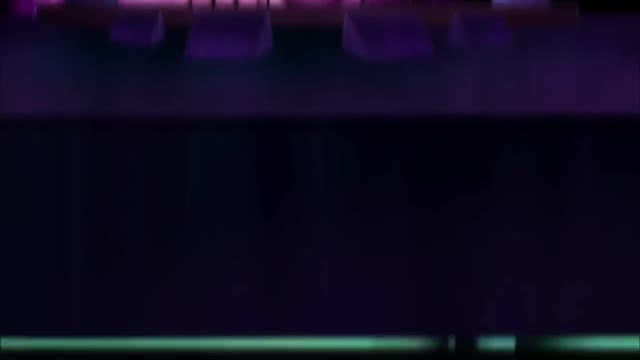} \\[1pt]
    \scriptsize\ttfamily 00:26 &
    \scriptsize\ttfamily 00:28 &
    \scriptsize\ttfamily 00:29 &
    \scriptsize\ttfamily 00:30 &
    \scriptsize\ttfamily 00:32 &
    \scriptsize\ttfamily 00:33 \\
  \end{tabular}

  \vspace{4pt}
  \begin{minipage}{\linewidth}
    \small
    \textbf{Q.} What instruments are already on the stage before the actors come on?

    \vspace{2pt}
    \begin{tabularx}{\linewidth}{@{}l X@{}}
      (A) & Harp \\
      (B) & Bass \\
      \gaGold{(C)} & \gaGold{Drum kit~~$\leftarrow$ gold} \\
      (D) & Piano \\
    \end{tabularx}

    \vspace{3pt}
    {\footnotesize \gaGold{\ding{51}}~video + question\quad \gaBad{\ding{55}}~question only\quad \gaBad{\ding{55}}~question only (DVD)\quad \gaBad{\ding{55}}~video only}
  \end{minipage}

  \caption{
    \textbf{LVBench}, \texttt{\_zrgbA3FMVE.mp4},
    question 0 (frames sampled inside the annotated \texttt{time\_reference} window).
    Harp, bass, drum kit and piano are all plausible on a stage. Nothing in the wording favours one, and both blind systems miss it.
  }
  \label{fig:a4-1}
\end{figure}

\begin{figure}[H]
  \centering
  \setlength{\tabcolsep}{0.5pt}
  \begin{tabular}{cccccc}
    \includegraphics[width=\dimexpr(\linewidth-6pt)/6\relax]{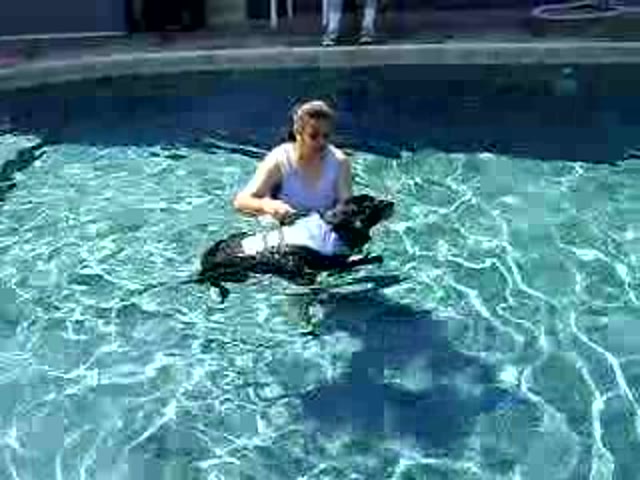} &
    \includegraphics[width=\dimexpr(\linewidth-6pt)/6\relax]{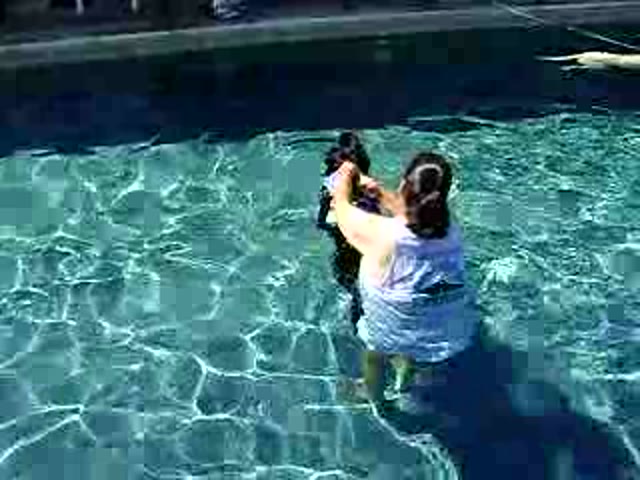} &
    \includegraphics[width=\dimexpr(\linewidth-6pt)/6\relax]{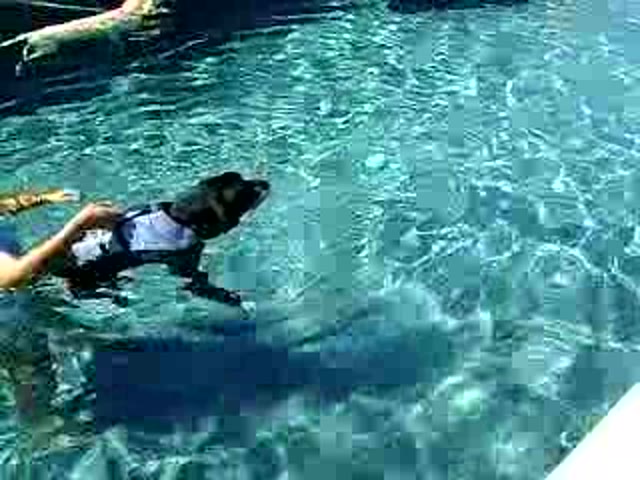} &
    \includegraphics[width=\dimexpr(\linewidth-6pt)/6\relax]{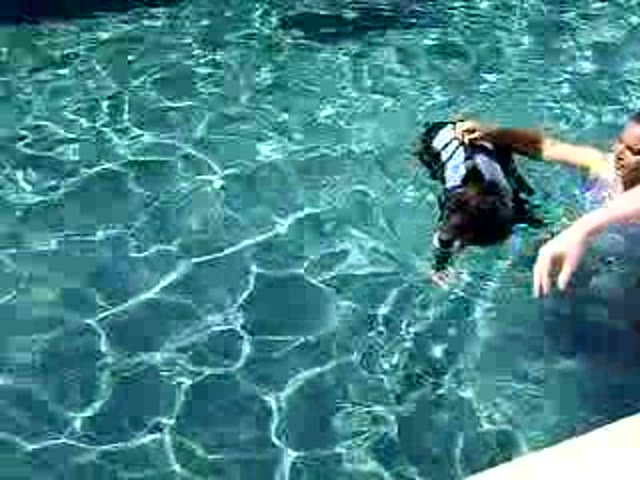} &
    \includegraphics[width=\dimexpr(\linewidth-6pt)/6\relax]{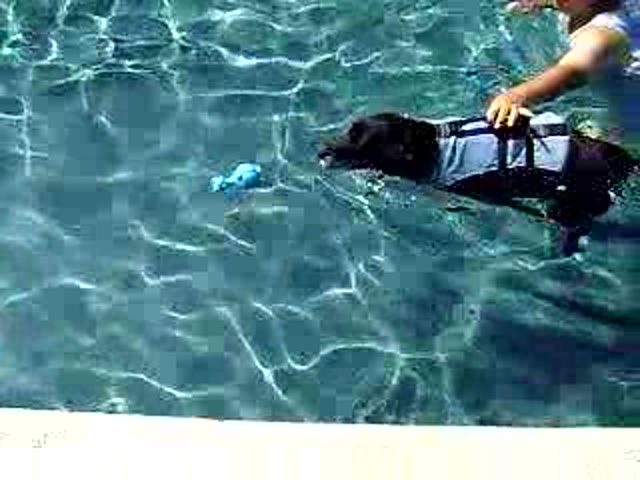} &
    \includegraphics[width=\dimexpr(\linewidth-6pt)/6\relax]{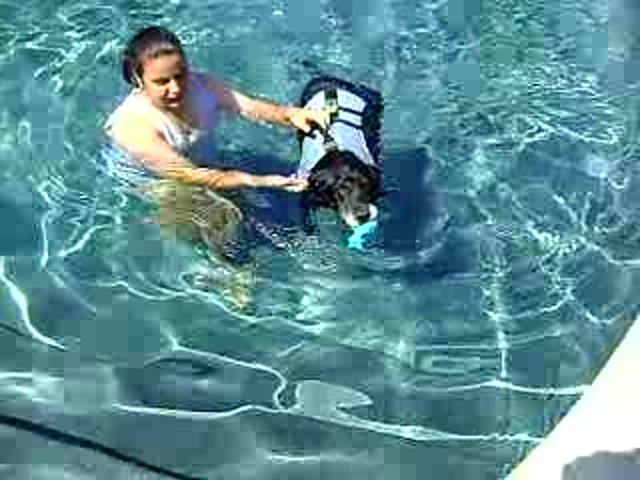} \\[1pt]
    \scriptsize\ttfamily 00:05 &
    \scriptsize\ttfamily 00:16 &
    \scriptsize\ttfamily 00:26 &
    \scriptsize\ttfamily 00:37 &
    \scriptsize\ttfamily 00:48 &
    \scriptsize\ttfamily 00:59 \\
  \end{tabular}

  \vspace{4pt}
  \begin{minipage}{\linewidth}
    \small
    \textbf{Q.} what is the dog doing

    \vspace{2pt}
    \begin{tabularx}{\linewidth}{@{}l X@{}}
      (A) & on the right \\
      \gaGold{(B)} & \gaGold{swimming~~$\leftarrow$ gold} \\
      (C) & lick baby s hand \\
      (D) & playing with sticks \\
      (E) & play with ball \\
    \end{tabularx}

    \vspace{3pt}
    {\footnotesize \gaGold{\ding{51}}~video + question\quad \gaBad{\ding{55}}~question only\quad \gaBad{\ding{55}}~question only (DVD)\quad \gaBad{\ding{55}}~video only}
  \end{minipage}

  \caption{
    \textbf{NExT-QA}, \texttt{2793806282.mp4},
    question 1 (frames sampled evenly across the whole video).
    A four-word stem carrying no constraint at all, against five ordinary dog activities. The answer lives only in the frames.
  }
  \label{fig:a4-2}
\end{figure}

\begin{figure}[H]
  \centering
  \setlength{\tabcolsep}{0.5pt}
  \begin{tabular}{cccccc}
    \includegraphics[width=\dimexpr(\linewidth-6pt)/6\relax]{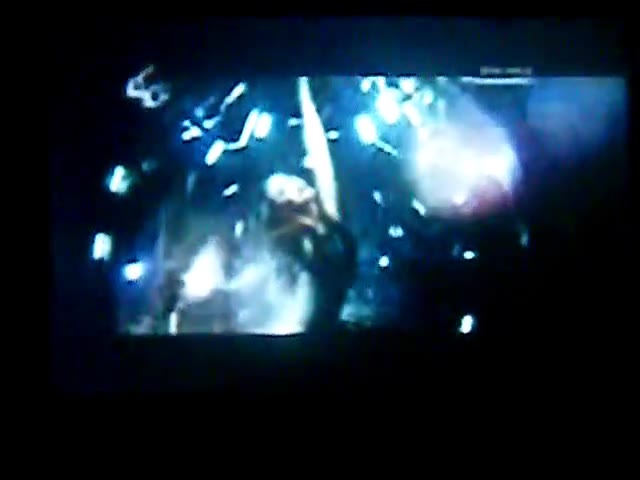} &
    \includegraphics[width=\dimexpr(\linewidth-6pt)/6\relax]{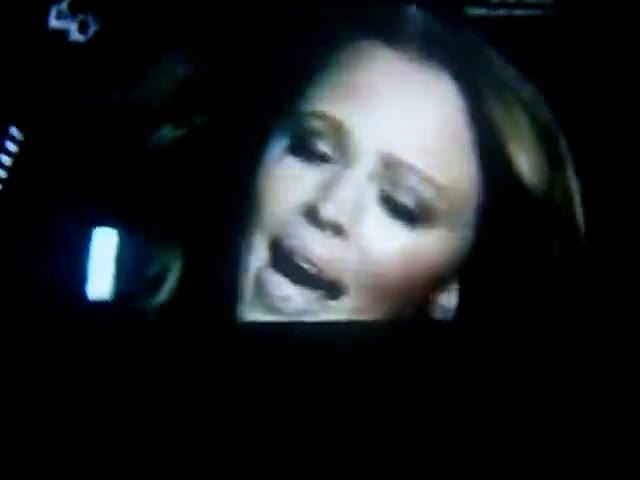} &
    \includegraphics[width=\dimexpr(\linewidth-6pt)/6\relax]{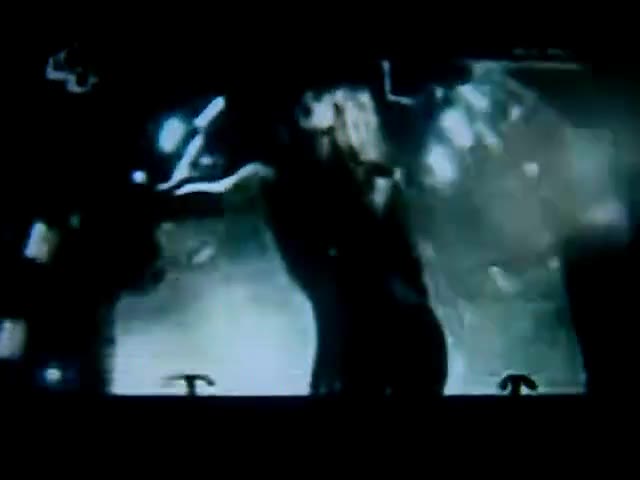} &
    \includegraphics[width=\dimexpr(\linewidth-6pt)/6\relax]{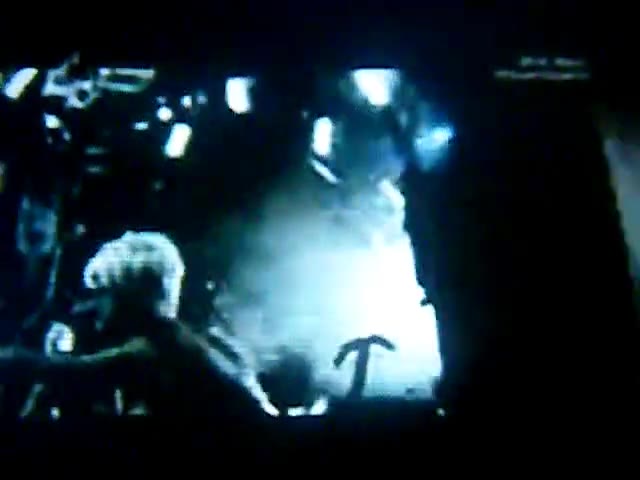} &
    \includegraphics[width=\dimexpr(\linewidth-6pt)/6\relax]{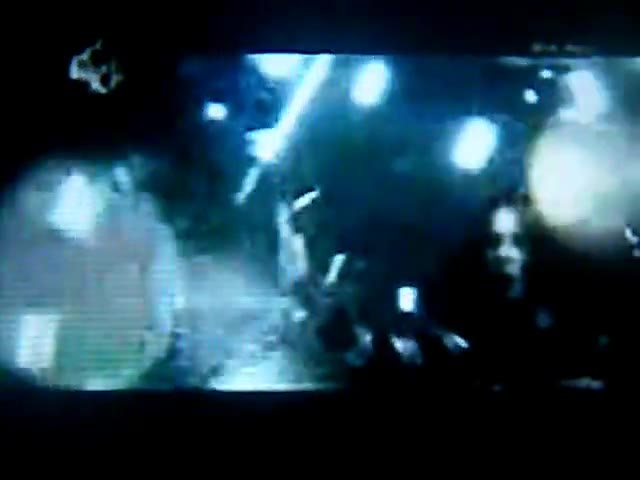} &
    \includegraphics[width=\dimexpr(\linewidth-6pt)/6\relax]{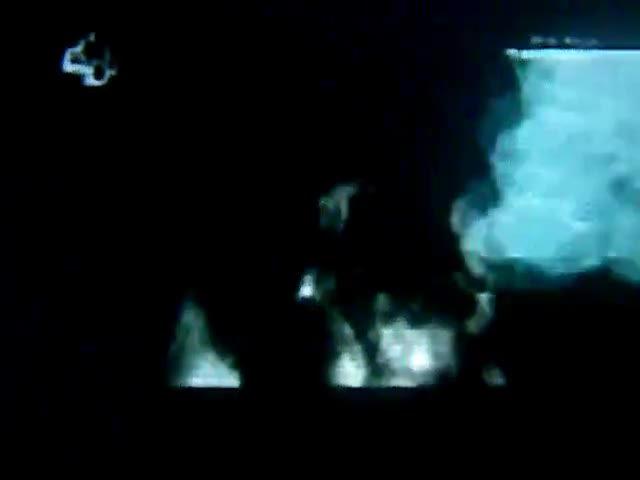} \\[1pt]
    \scriptsize\ttfamily 00:01 &
    \scriptsize\ttfamily 00:03 &
    \scriptsize\ttfamily 00:05 &
    \scriptsize\ttfamily 00:07 &
    \scriptsize\ttfamily 00:09 &
    \scriptsize\ttfamily 00:11 \\
  \end{tabular}

  \vspace{4pt}
  \begin{minipage}{\linewidth}
    \small
    \textbf{Q.} How is the clarity of the characters in this video?

    \vspace{2pt}
    \begin{tabularx}{\linewidth}{@{}l X@{}}
      (A) & Average \\
      (B) & Below average \\
      (C) & Good \\
      \gaGold{(D)} & \gaGold{Very poor~~$\leftarrow$ gold} \\
    \end{tabularx}

    \vspace{3pt}
    {\footnotesize \gaGold{\ding{51}}~video + question\quad \gaBad{\ding{55}}~question only\quad \gaBad{\ding{55}}~question only (DVD)\quad \gaBad{\ding{55}}~video only}
  \end{minipage}

  \caption{
    \textbf{QBench-Video}, \texttt{0019.mp4},
    question 0 (frames sampled evenly across the whole video).
    Four genuinely distinct points on an ordinal quality scale, none blank. The judgement cannot be made without watching.
  }
  \label{fig:a4-3}
\end{figure}

\subsection{When the video does not help}
\label{sec:appendix-a5}

\begin{figure}[H]
  \centering
  \setlength{\tabcolsep}{0.5pt}
  \begin{tabular}{cccccc}
    \includegraphics[width=\dimexpr(\linewidth-6pt)/6\relax]{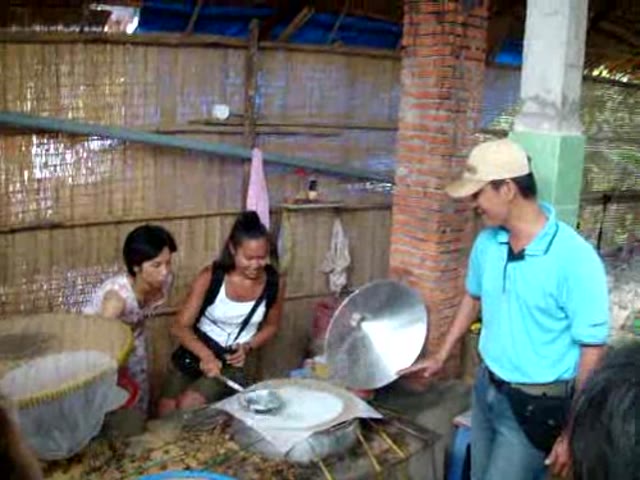} &
    \includegraphics[width=\dimexpr(\linewidth-6pt)/6\relax]{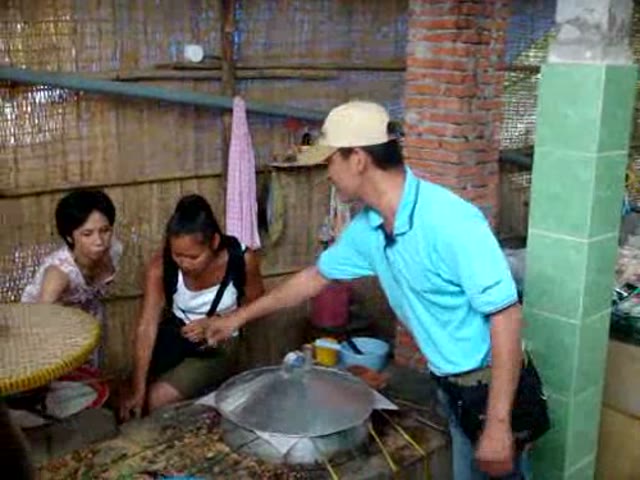} &
    \includegraphics[width=\dimexpr(\linewidth-6pt)/6\relax]{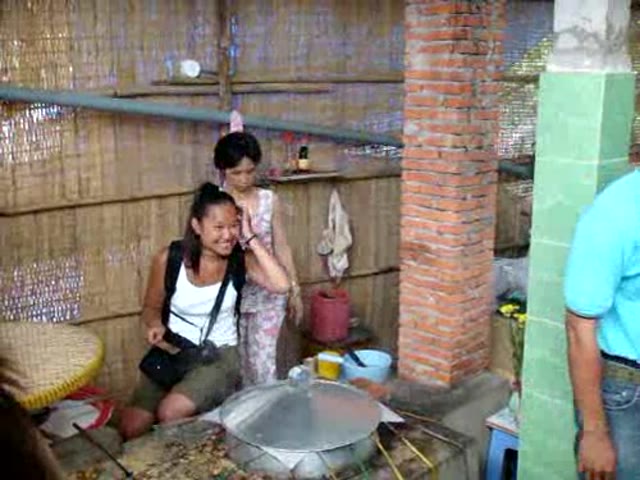} &
    \includegraphics[width=\dimexpr(\linewidth-6pt)/6\relax]{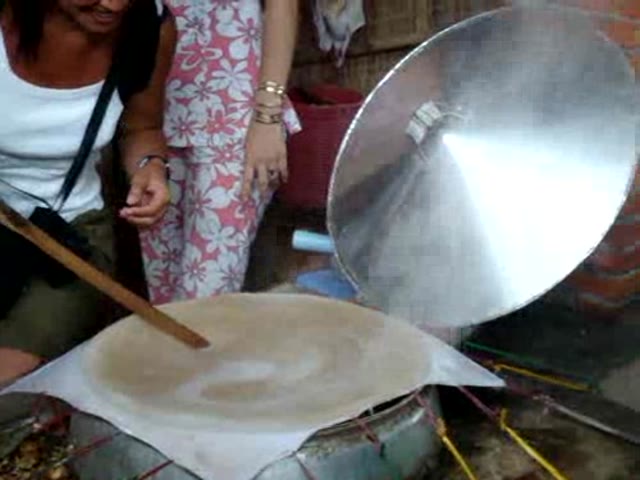} &
    \includegraphics[width=\dimexpr(\linewidth-6pt)/6\relax]{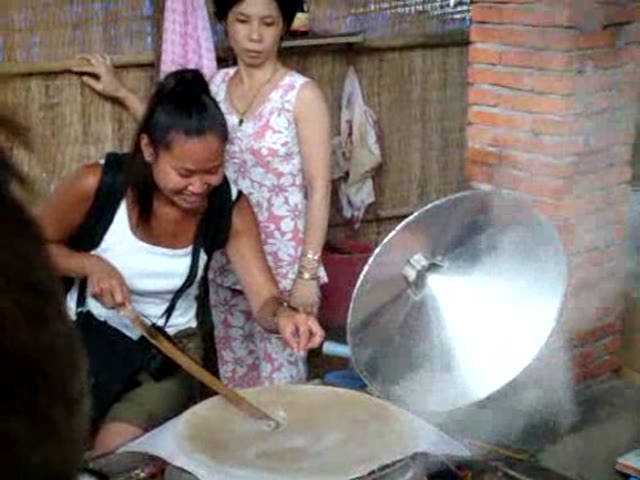} &
    \includegraphics[width=\dimexpr(\linewidth-6pt)/6\relax]{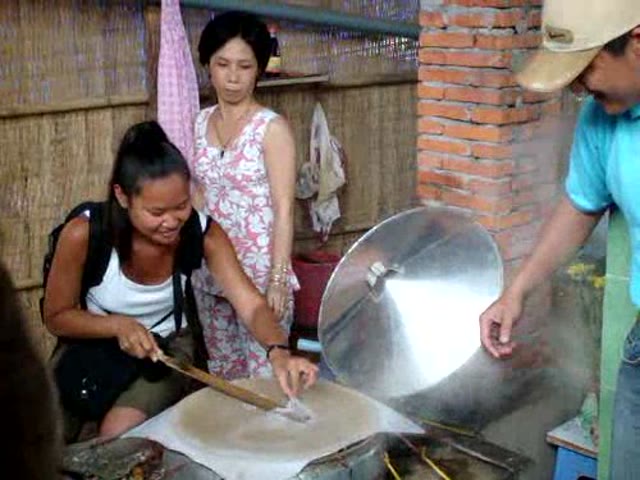} \\[1pt]
    \scriptsize\ttfamily 00:04 &
    \scriptsize\ttfamily 00:12 &
    \scriptsize\ttfamily 00:21 &
    \scriptsize\ttfamily 00:30 &
    \scriptsize\ttfamily 00:38 &
    \scriptsize\ttfamily 00:47 \\
  \end{tabular}

  \vspace{4pt}
  \begin{minipage}{\linewidth}
    \small
    \textbf{Q.} how does the girl pick up the food after it is cooked

    \vspace{2pt}
    \begin{tabularx}{\linewidth}{@{}l X@{}}
      \gaGold{(A)} & \gaGold{use tongs~~$\leftarrow$ gold} \\
      (B) & hold it tightly \\
      (C) & with her hands \\
      (D) & use hand gestures \\
      (E) & with the spoon \\
    \end{tabularx}

    \vspace{3pt}
    {\footnotesize \gaBad{\ding{55}}~video + question\quad \gaGold{\ding{51}}~question only\quad \gaGold{\ding{51}}~question only (DVD)\quad \gaBad{\ding{55}}~video only}
  \end{minipage}

  \caption{
    \textbf{NExT-QA}, \texttt{2868183184.mp4},
    question 4 (frames sampled evenly across the whole video).
    The baseline answers \emph{with her hands}; the blind systems answer \emph{use tongs} and are right. The frames evidently supported the wrong reading while the prior held.
  }
  \label{fig:a5-1}
\end{figure}

\begin{figure}[H]
  \centering
  \setlength{\tabcolsep}{0.5pt}
  \begin{tabular}{cccccc}
    \includegraphics[width=\dimexpr(\linewidth-6pt)/6\relax]{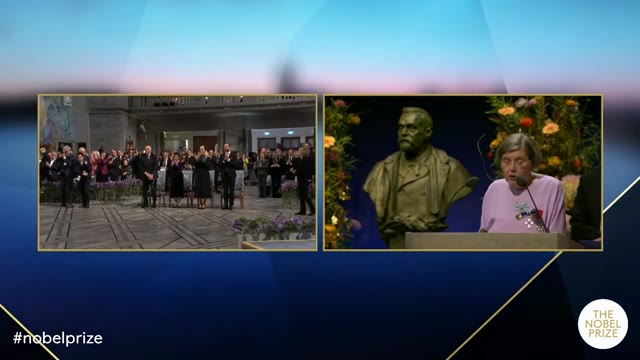} &
    \includegraphics[width=\dimexpr(\linewidth-6pt)/6\relax]{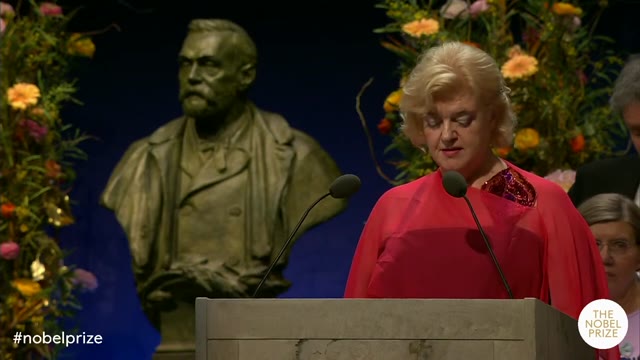} &
    \includegraphics[width=\dimexpr(\linewidth-6pt)/6\relax]{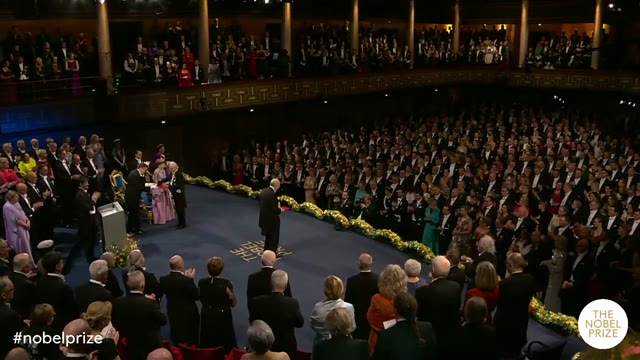} &
    \includegraphics[width=\dimexpr(\linewidth-6pt)/6\relax]{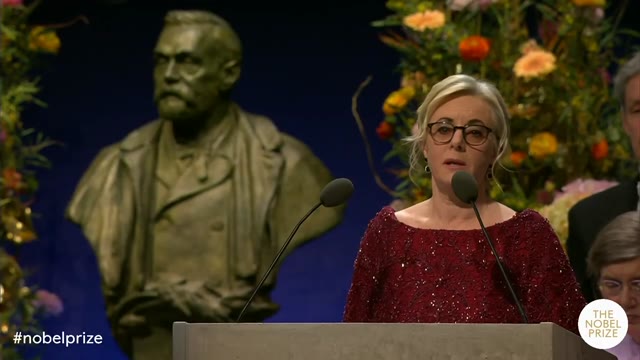} &
    \includegraphics[width=\dimexpr(\linewidth-6pt)/6\relax]{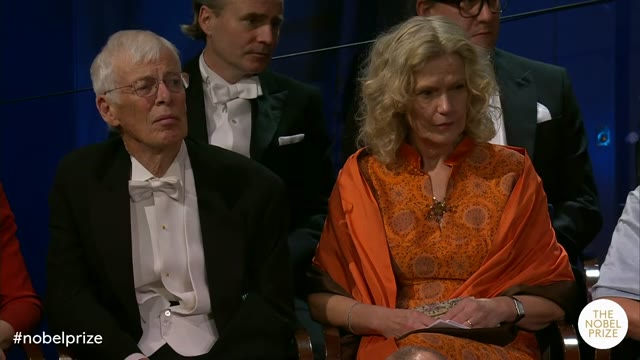} &
    \includegraphics[width=\dimexpr(\linewidth-6pt)/6\relax]{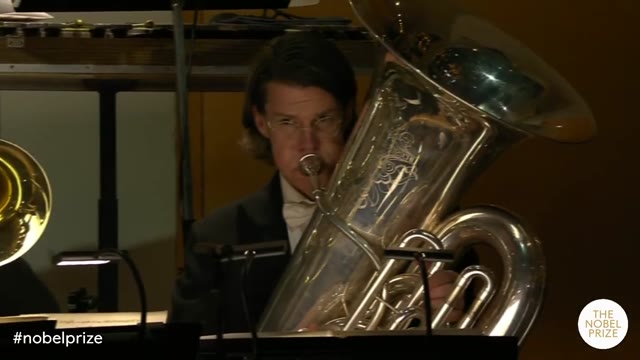} \\[1pt]
    \scriptsize\ttfamily 05:50 &
    \scriptsize\ttfamily 17:30 &
    \scriptsize\ttfamily 29:10 &
    \scriptsize\ttfamily 40:50 &
    \scriptsize\ttfamily 52:30 &
    \scriptsize\ttfamily 64:10 \\
  \end{tabular}

  \vspace{4pt}
  \begin{minipage}{\linewidth}
    \small
    \textbf{Q.} How many hostesses are there?

    \vspace{2pt}
    \begin{tabularx}{\linewidth}{@{}l X@{}}
      (A) & 5 \\
      (B) & 1 \\
      \gaGold{(C)} & \gaGold{3~~$\leftarrow$ gold} \\
      (D) & 2 \\
    \end{tabularx}

    \vspace{3pt}
    {\footnotesize \gaBad{\ding{55}}~video + question\quad \gaGold{\ding{51}}~question only\quad \gaGold{\ding{51}}~question only (DVD)\quad \gaBad{\ding{55}}~video only}
  \end{minipage}

  \caption{
    \textbf{LVBench}, \texttt{lDlA7cfNk8A.mp4},
    question 3 (frames sampled inside the annotated \texttt{time\_reference} window).
    Counting from an hour of video: the grounded run undercounts, the blind guess lands. Counting is the most common way the video condition loses to its own prior.
  }
  \label{fig:a5-2}
\end{figure}

\begin{figure}[H]
  \centering
  \setlength{\tabcolsep}{0.5pt}
  \begin{tabular}{cccccc}
    \includegraphics[width=\dimexpr(\linewidth-6pt)/6\relax]{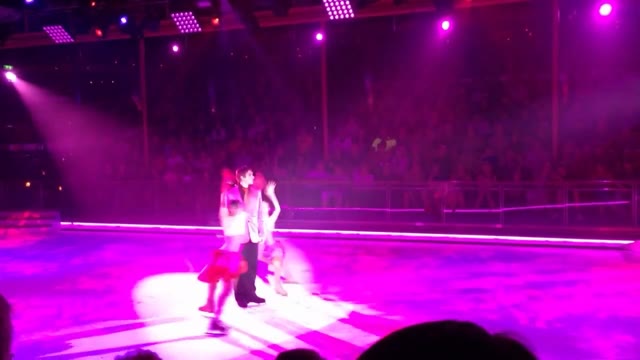} &
    \includegraphics[width=\dimexpr(\linewidth-6pt)/6\relax]{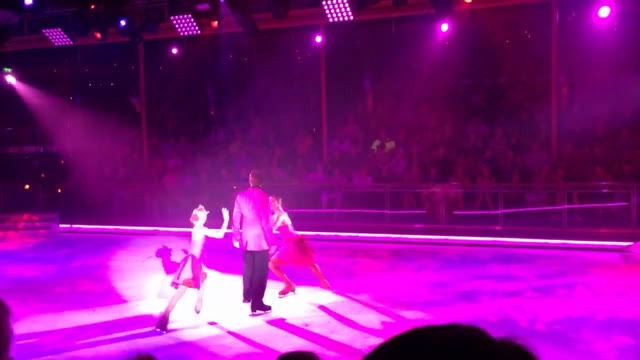} &
    \includegraphics[width=\dimexpr(\linewidth-6pt)/6\relax]{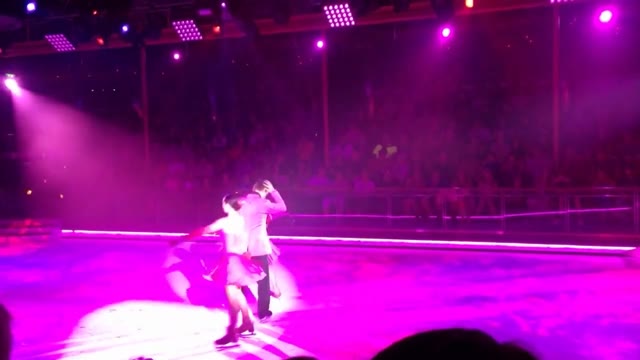} &
    \includegraphics[width=\dimexpr(\linewidth-6pt)/6\relax]{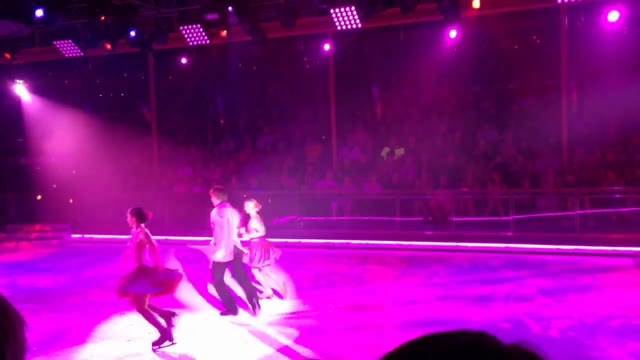} &
    \includegraphics[width=\dimexpr(\linewidth-6pt)/6\relax]{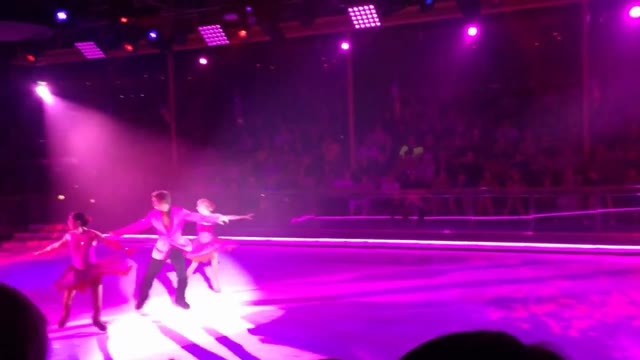} &
    \includegraphics[width=\dimexpr(\linewidth-6pt)/6\relax]{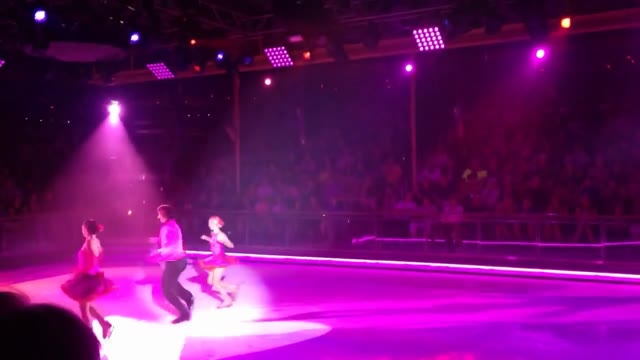} \\[1pt]
    \scriptsize\ttfamily 00:00 &
    \scriptsize\ttfamily 00:02 &
    \scriptsize\ttfamily 00:04 &
    \scriptsize\ttfamily 00:05 &
    \scriptsize\ttfamily 00:07 &
    \scriptsize\ttfamily 00:09 \\
  \end{tabular}

  \vspace{4pt}
  \begin{minipage}{\linewidth}
    \small
    \textbf{Q.} What are the main factors affecting the quality of dancers in the video?

    \vspace{2pt}
    \begin{tabularx}{\linewidth}{@{}l X@{}}
      \gaGold{(A)} & \gaGold{Uneven lighting on the dancers' bodies~~$\leftarrow$ gold} \\
      (B) & Blurry content \\
      (C) & Low brightness \\
      (D) & Overexposure \\
    \end{tabularx}

    \vspace{3pt}
    {\footnotesize \gaBad{\ding{55}}~video + question\quad \gaGold{\ding{51}}~question only\quad \gaGold{\ding{51}}~question only (DVD)\quad \gaGold{\ding{51}}~video only}
  \end{minipage}

  \caption{
    \textbf{QBench-Video}, \texttt{1376.mp4},
    question 0 (frames sampled evenly across the whole video).
    Uneven lighting and overexposure are easily confused from frames, and the grounded run picks the wrong one while the blind prior picks the more commonly annotated cause.
  }
  \label{fig:a5-3}
\end{figure}

\subsection{When neither condition helps}
\label{sec:appendix-a6}

\begin{figure}[H]
  \centering
  \setlength{\tabcolsep}{0.5pt}
  \begin{tabular}{cccccc}
    \includegraphics[width=\dimexpr(\linewidth-6pt)/6\relax]{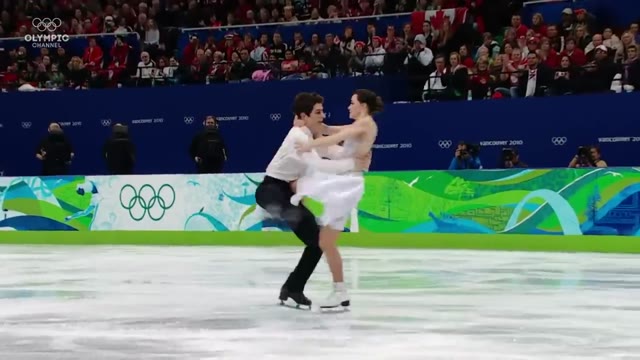} &
    \includegraphics[width=\dimexpr(\linewidth-6pt)/6\relax]{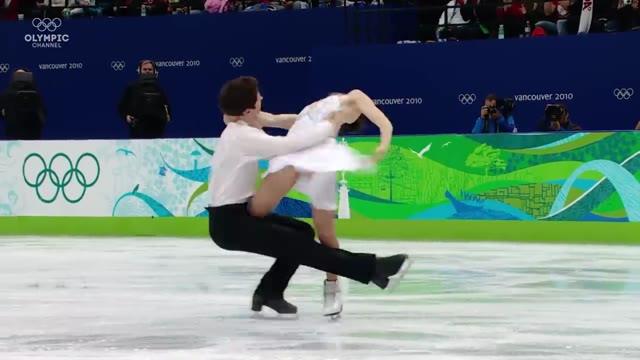} &
    \includegraphics[width=\dimexpr(\linewidth-6pt)/6\relax]{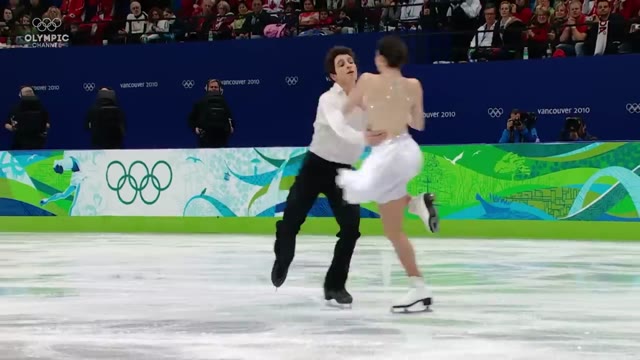} &
    \includegraphics[width=\dimexpr(\linewidth-6pt)/6\relax]{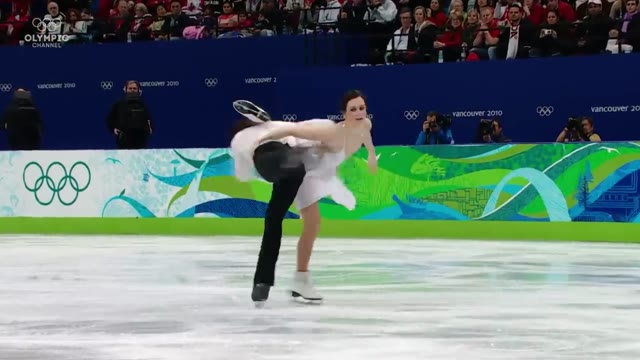} &
    \includegraphics[width=\dimexpr(\linewidth-6pt)/6\relax]{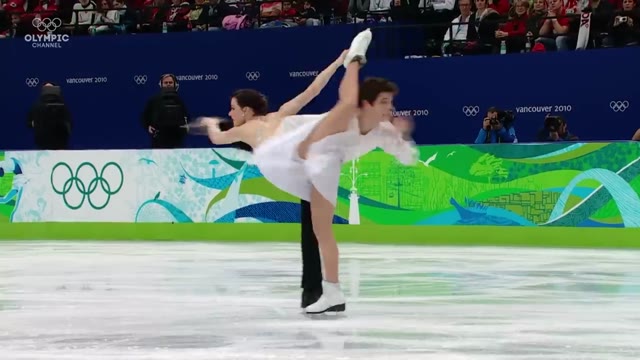} &
    \includegraphics[width=\dimexpr(\linewidth-6pt)/6\relax]{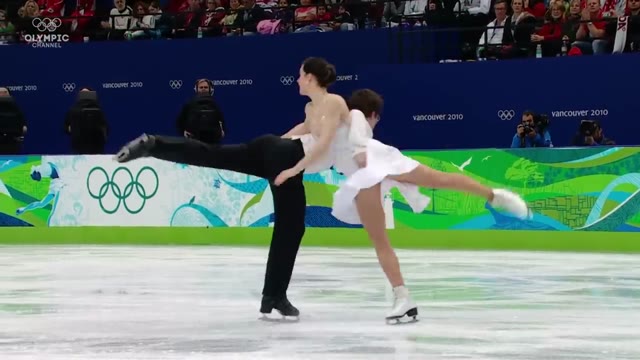} \\[1pt]
    \scriptsize\ttfamily 10:21 &
    \scriptsize\ttfamily 10:23 &
    \scriptsize\ttfamily 10:25 &
    \scriptsize\ttfamily 10:28 &
    \scriptsize\ttfamily 10:30 &
    \scriptsize\ttfamily 10:32 \\
  \end{tabular}

  \vspace{4pt}
  \begin{minipage}{\linewidth}
    \small
    \textbf{Q.} How many times do two contestants circle together between 10:20-10:34?

    \vspace{2pt}
    \begin{tabularx}{\linewidth}{@{}l X@{}}
      (A) & 20 \\
      (B) & 8 \\
      (C) & 18 \\
      \gaGold{(D)} & \gaGold{13~~$\leftarrow$ gold} \\
    \end{tabularx}

    \vspace{3pt}
    {\footnotesize \gaBad{\ding{55}}~video + question\quad \gaBad{\ding{55}}~question only\quad \gaBad{\ding{55}}~question only (DVD)\quad \gaBad{\ding{55}}~video only}
  \end{minipage}

  \caption{
    \textbf{LVBench}, \texttt{2LH3JCGkEBU.mp4},
    question 4 (frames sampled inside the annotated \texttt{time\_reference} window).
    Fine-grained counting over a 14-second window, with 8/13/18/20 as options. Both conditions answer 18. Whether any model could resolve this from 1 fps frames is doubtful.
  }
  \label{fig:a6-1}
\end{figure}

\begin{figure}[H]
  \centering
  \setlength{\tabcolsep}{0.5pt}
  \begin{tabular}{cccccc}
    \includegraphics[width=\dimexpr(\linewidth-6pt)/6\relax]{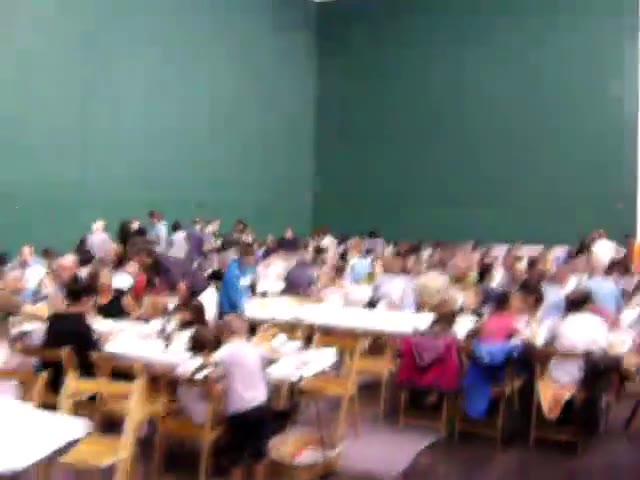} &
    \includegraphics[width=\dimexpr(\linewidth-6pt)/6\relax]{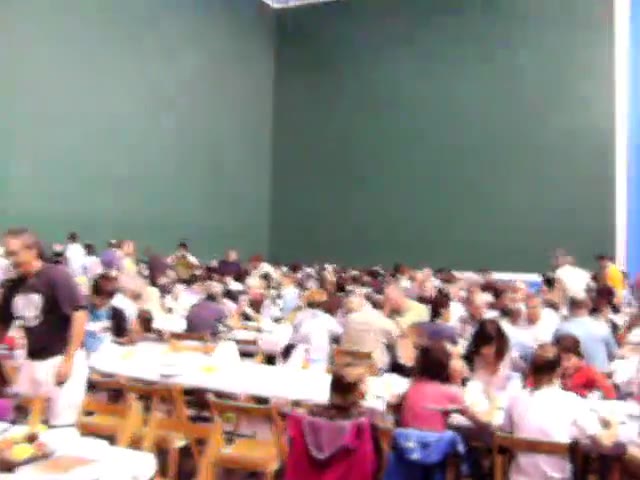} &
    \includegraphics[width=\dimexpr(\linewidth-6pt)/6\relax]{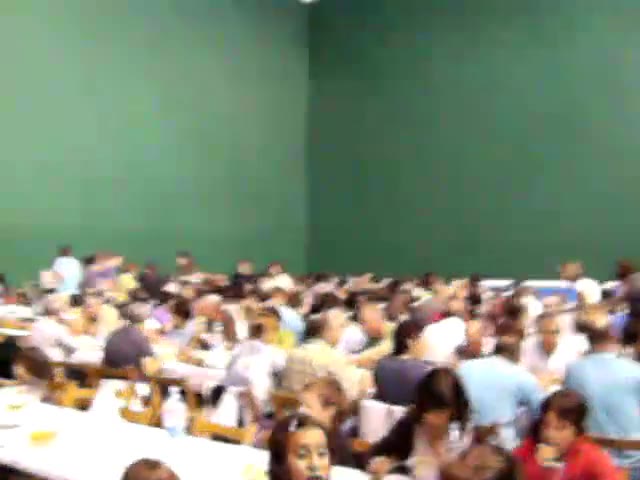} &
    \includegraphics[width=\dimexpr(\linewidth-6pt)/6\relax]{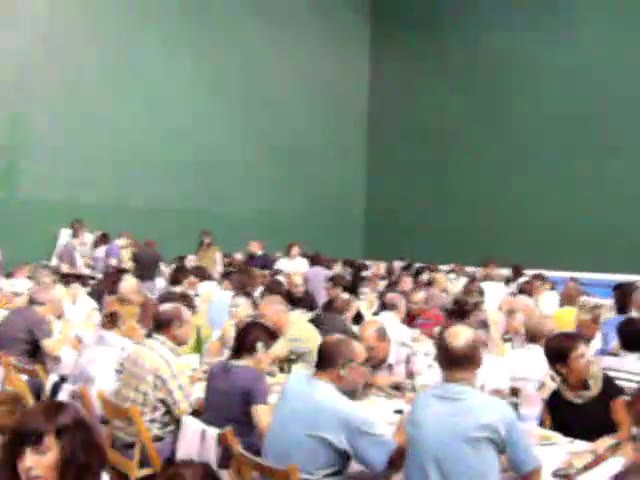} &
    \includegraphics[width=\dimexpr(\linewidth-6pt)/6\relax]{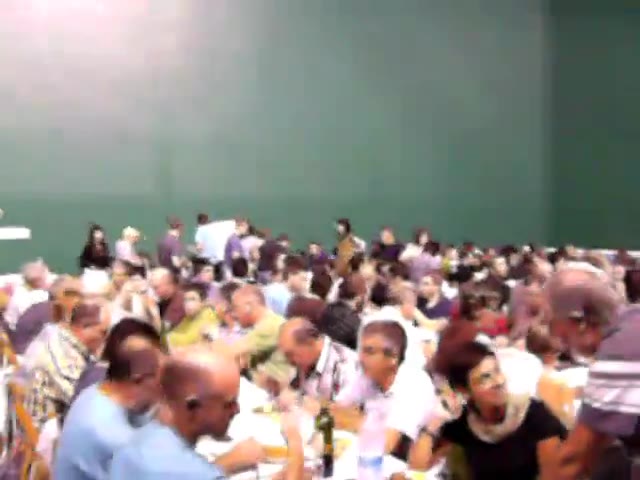} &
    \includegraphics[width=\dimexpr(\linewidth-6pt)/6\relax]{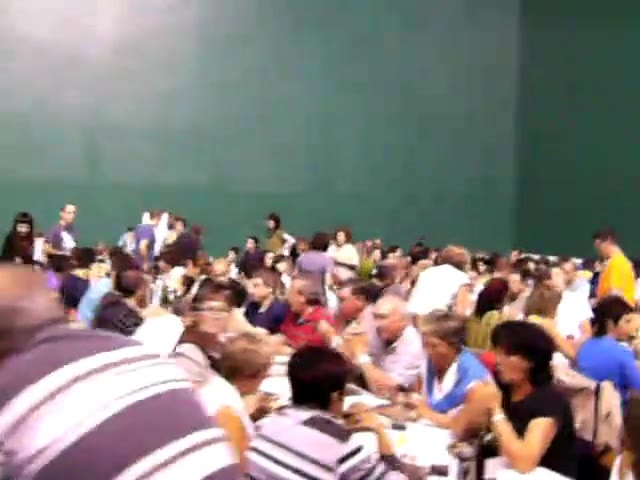} \\[1pt]
    \scriptsize\ttfamily 00:01 &
    \scriptsize\ttfamily 00:03 &
    \scriptsize\ttfamily 00:05 &
    \scriptsize\ttfamily 00:07 &
    \scriptsize\ttfamily 00:09 &
    \scriptsize\ttfamily 00:11 \\
  \end{tabular}

  \vspace{4pt}
  \begin{minipage}{\linewidth}
    \small
    \textbf{Q.} How clear are the details of the crowd in the video?

    \vspace{2pt}
    \begin{tabularx}{\linewidth}{@{}l X@{}}
      (A) & Poor \\
      (B) & Average \\
      (C) & Good \\
      \gaGold{(D)} & \gaGold{Very poor~~$\leftarrow$ gold} \\
    \end{tabularx}

    \vspace{3pt}
    {\footnotesize \gaBad{\ding{55}}~video + question\quad \gaBad{\ding{55}}~question only\quad \gaBad{\ding{55}}~question only (DVD)\quad \gaBad{\ding{55}}~video only}
  \end{minipage}

  \caption{
    \textbf{QBench-Video}, \texttt{0103.mp4},
    question 0 (frames sampled evenly across the whole video).
    An ordinal scale whose adjacent points (Poor / Very poor) are not separated by any stated criterion. The baseline answers one step off gold. The item measures agreement with an annotator's cut point.
  }
  \label{fig:a6-2}
\end{figure}

\begin{figure}[H]
  \centering
  \setlength{\tabcolsep}{0.5pt}
  \begin{tabular}{cccccc}
    \includegraphics[width=\dimexpr(\linewidth-6pt)/6\relax]{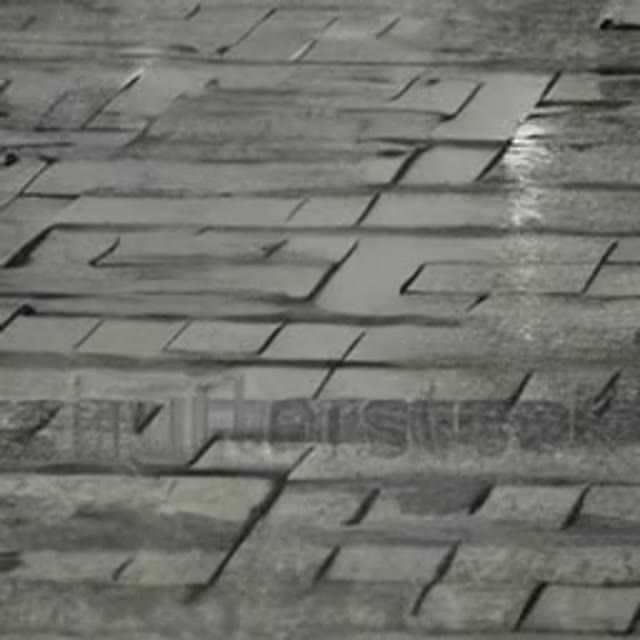} &
    \includegraphics[width=\dimexpr(\linewidth-6pt)/6\relax]{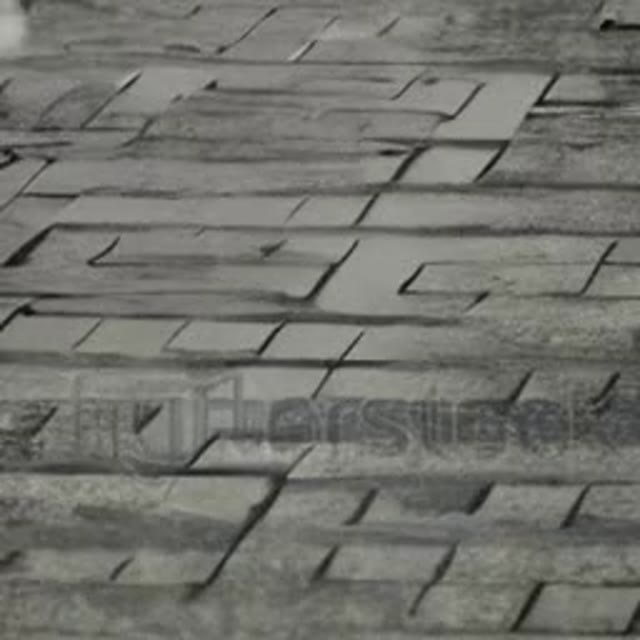} &
    \includegraphics[width=\dimexpr(\linewidth-6pt)/6\relax]{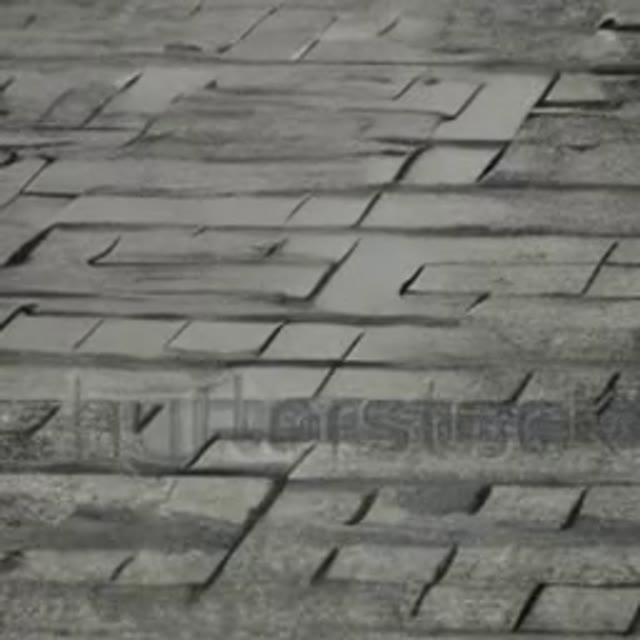} &
    \includegraphics[width=\dimexpr(\linewidth-6pt)/6\relax]{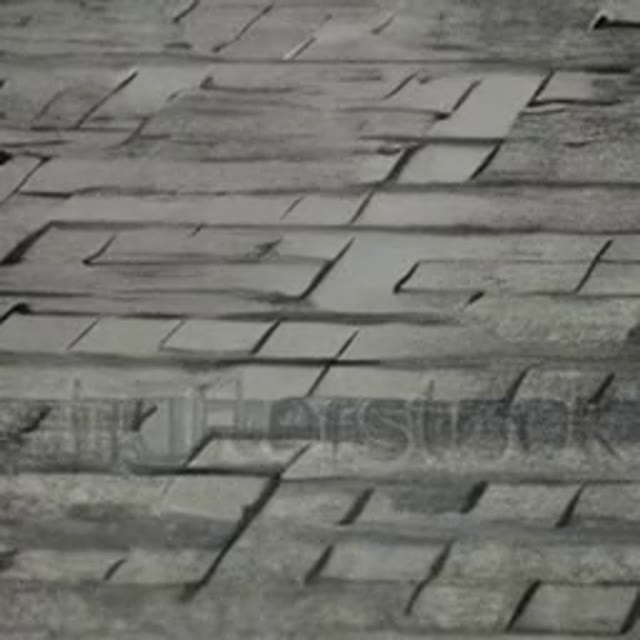} &
    \includegraphics[width=\dimexpr(\linewidth-6pt)/6\relax]{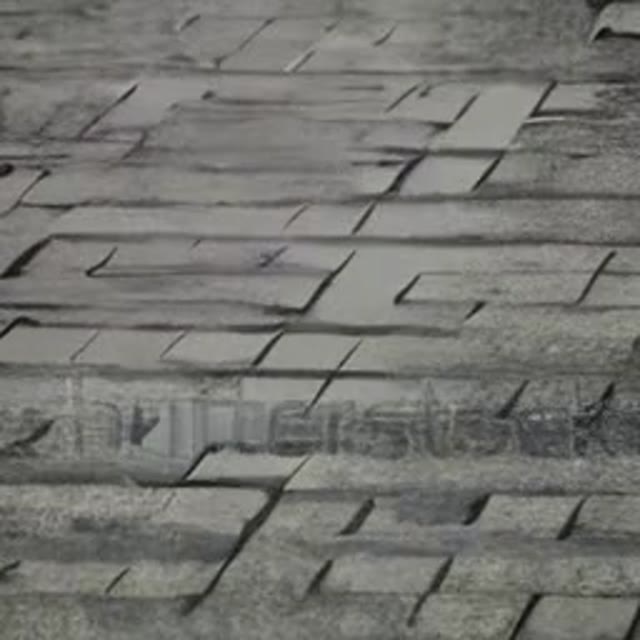} &
    \includegraphics[width=\dimexpr(\linewidth-6pt)/6\relax]{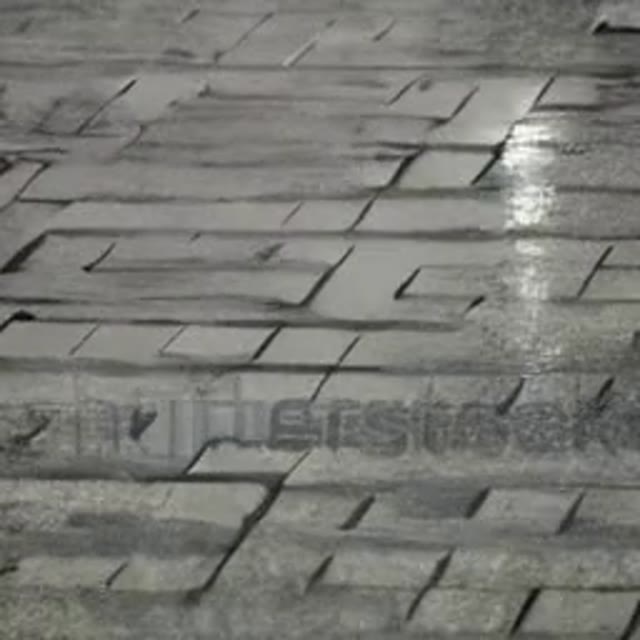} \\[1pt]
    \scriptsize\ttfamily 00:00 &
    \scriptsize\ttfamily 00:00 &
    \scriptsize\ttfamily 00:00 &
    \scriptsize\ttfamily 00:01 &
    \scriptsize\ttfamily 00:01 &
    \scriptsize\ttfamily 00:01 \\
  \end{tabular}

  \vspace{4pt}
  \begin{minipage}{\linewidth}
    \small
    \textbf{Q.} What kind of distortion does not exist in this video?

    \vspace{2pt}
    \begin{tabularx}{\linewidth}{@{}l X@{}}
      (A) & Blurriness \\
      (B) & Overexposure \\
      (C) & Noise \\
      \gaGold{(D)} & \gaGold{Underexposure~~$\leftarrow$ gold} \\
    \end{tabularx}

    \vspace{3pt}
    {\footnotesize \gaBad{\ding{55}}~video + question\quad \gaBad{\ding{55}}~question only\quad \gaBad{\ding{55}}~question only (DVD)\quad \gaBad{\ding{55}}~video only}
  \end{minipage}

  \caption{
    \textbf{QBench-Video}, \texttt{3001002.mp4},
    question 0 (frames sampled evenly across the whole video).
    A negated stem over four fine perceptual judgements. Both conditions answer Overexposure; gold is Underexposure.
  }
  \label{fig:a6-3}
\end{figure}